\documentclass{article}
\usepackage{graphicx}
\usepackage{booktabs} %

\usepackage{mathtools}
\usepackage{algorithm}
\usepackage{multirow} 
\usepackage{xspace}

\usepackage{comment}
\usepackage{amssymb} %
\usepackage{bbold}

\usepackage{tabulary}
\usepackage{tabularx}
\usepackage{makecell}
\usepackage{tabulary}
\usepackage{makecell}
\usepackage{wrapfig}
\usepackage{subcaption}

\usepackage{amssymb}
\usepackage{mathtools}
\usepackage{amsthm}
\usepackage{array}

\usepackage{listings}
\usepackage{xcolor}

\usepackage{titlesec}

\titlespacing*{\section}{0pt}{0.6\baselineskip}{0.4\baselineskip}
\titlespacing*{\subsection}{0pt}{0.5\baselineskip}{0.3\baselineskip}
\titlespacing*{\paragraph}{0pt}{0.3\baselineskip}{1em}

\usepackage{enumitem}
\setlist{nosep} 

\definecolor{codebg}{rgb}{0.97,0.97,0.97}
\definecolor{keywordcolor}{rgb}{0.0,0.2,0.6}
\definecolor{funcname}{rgb}{0.4,0.1,0.7}
\definecolor{classname}{rgb}{0.1,0.4,0.7}
\definecolor{commentcolor}{rgb}{0.4,0.4,0.4}
\definecolor{stringcolor}{rgb}{0.137, 0.667, 1.0}
\usepackage[preprint]{corl_2025} 

\title{Lucid-XR: An Extended-Reality Data Engine for Robotic Manipulation}

\author{
\textbf{Yajvan Ravan\textsuperscript{1,2}*}, 
\textbf{Adam Rashid\textsuperscript{1}*},
\textbf{Alan Yu\textsuperscript{1,2}},
\textbf{Kai McClennen\textsuperscript{1,2}},
\textbf{Gio Huh\textsuperscript{2,3}},
\textbf{Kevin Yang\textsuperscript{2,4}},
\\[0.2em]
\textbf{Zhutian Yang\textsuperscript{1}},
\textbf{Qinxi Yu\textsuperscript{5}},
\textbf{Xiaolong Wang\textsuperscript{5}},
\textbf{Phillip Isola\textsuperscript{1$\dagger$}},
\textbf{Ge Yang\textsuperscript{1,2$\dagger$}*}
\\[0.6em]
\textsuperscript{1}MIT CSAIL,\;
\textsuperscript{2}FortyFive Labs,\;
\textsuperscript{3}Caltech,\;
\textsuperscript{4}Harvard University,\;
\textsuperscript{5}UC San Diego
\\[0.6em]
\textsuperscript{*}\textit{Equal contribution} \quad
\textsuperscript{$\dagger$}\textit{Equal advising}
}

\begin{document}

\makeatletter
\let\@oldmaketitle\@maketitle%
\renewcommand{\@maketitle}{\@oldmaketitle%
\includegraphics[width=1.0\linewidth]{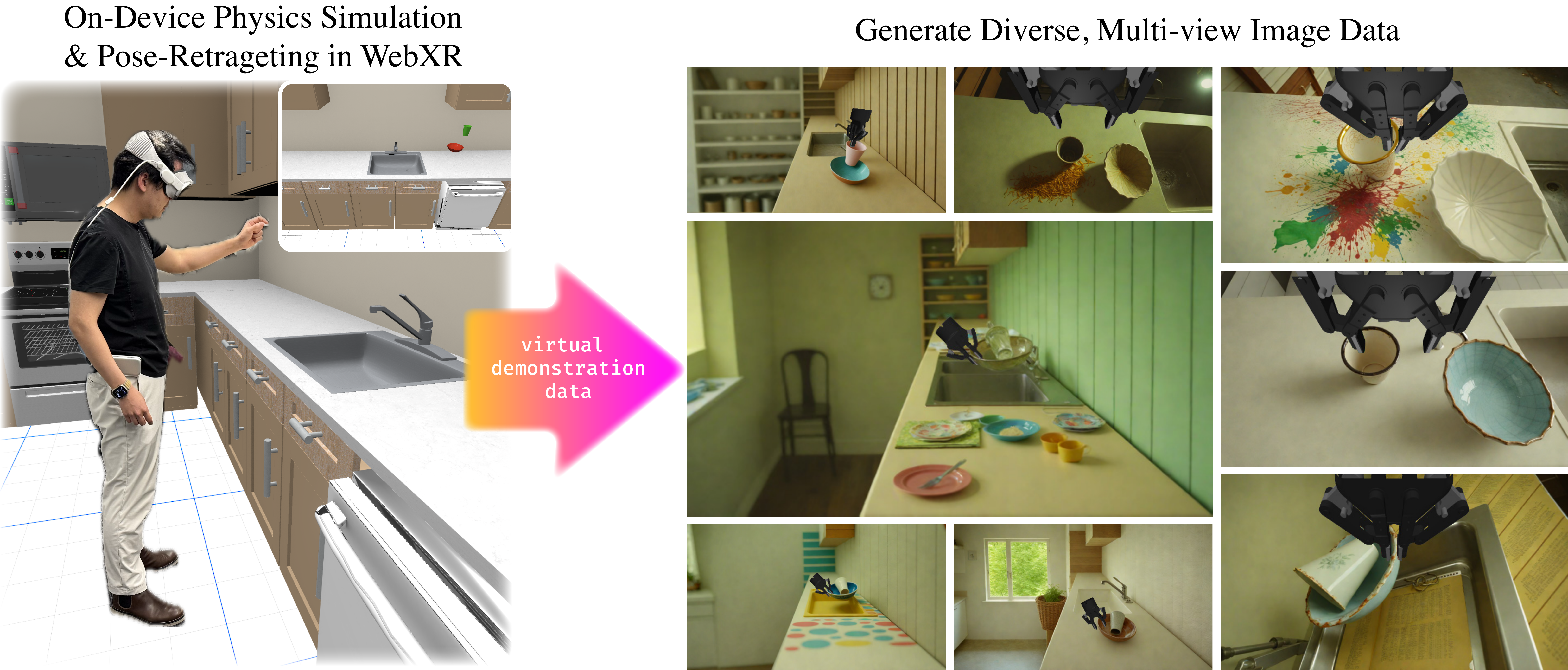}%
\captionof{figure}{
\textbf{An Extended Reality Data Engine for Robotic Manipulation. }
\textit{Left}:~we deliver physics simulation to run directly on XR devices via the web browser, to enable internet-scale crowdsourcing of demonstration data collection.
\textit{Right}:~Our GenAI-powered synthetic data engine creates steerable, diverse, and realistic multi-view visual data to train real-world robots.
}\label{fig:teaser}
\vspace{1.5em}%
}%
\maketitle


\begin{abstract}
We introduce Lucid-XR, a generative data engine for creating diverse and realistic-looking multi-modal data to train real-world robotic systems. At the core of Lucid-XR is vuer, a web-based physics simulation environment that runs directly on the XR headset, enabling internet-scale access to immersive, latency-free virtual interactions without requiring specialized equipment. The complete system integrates on-device physics simulation with human-to-robot pose retargeting. Data collected is further amplified by a physics-guided video generation pipeline steerable via natural language specifications. We demonstrate zero-shot transfer of robot visual policies to unseen, cluttered, and badly lit evaluation environments, after training entirely on Lucid-XR's synthetic data. We include examples across dexterous manipulation tasks that involve soft materials, loosely bound particles, and rigid body contact. Project website: \href{https://lucidxr.github.io}{https://lucidxr.github.io}
\end{abstract}



\section{Introduction: From Atoms to Bits}
\label{sec:intro}

When viewed from afar, training a robot controller is not too different from making a movie, in that both involve carefully curating content. Feature films over the years have transitioned steadily from practical special effects in the physical world to digitally created virtual special effects, driven primarily by creators' demand for even greater freedom in storytelling~\cite{Ryu2007vfx-reality,Turnock2009vfx-ILM,Das2023vfx-evolution,Murodillayev2024vfx-impact}. Hardening robotic systems for real-world deployment requires a similar level of control over the training environment to cover rare but mission-critical events that, by definition, are scarce in real-world datasets. What makes robotics more difficult is that digitally creating millions of realistic-looking virtual worlds at the scale required for our robots to generalize to the real world is infeasibly expensive.

In this work, we introduce Lucid-XR,  an extended-reality (XR) data engine for robot manipulation. Lucid-XR uses a generative image pipeline to convert human demonstration data collected in low-fidelity virtual environments into diverse and visually realistic data to train the robot. Key to our vision is to enable internet scale deployment of real-time multi-physics simulations through the web browser to enable the crowd-sourcing of unlimited human demonstration data. Furthermore, virtual demonstrations in sparsely populated 3D environments alone are insufficient for training real-world computer vision systems. We leverage language and text-to-image generative models to construct a physics-guided video generation pipeline that amplifies a small number of simple designs into millions of diverse and realistic-looking multi-view images for the robot.

One challenge that makes crowd-sourcing teleoperation difficult is that retargeting human poses to robot form factors that are kinematically different involves writing custom computer code for each robot and setting up a server. We solve this problem by leveraging an inverse kinematics solver, built-in to MuJoCo, and allowing users to define bindings between motion capture (MoCap) sites and robot parts in the markup schema.

Our contributions are three-fold: first, moving physics simulation onto the XR device, to deliver a latency-free multi-physics simulation in an immersive environment via the open internet; second, a way to retarget human pose data to virtual robots, without requiring a custom computer program; and third, a demonstration of a policy trained on Lucid-XR's synthetic data deployed on scanned digital twins of real environments. 


\begin{figure}%
\centering%
\includegraphics[width=0.85\linewidth]{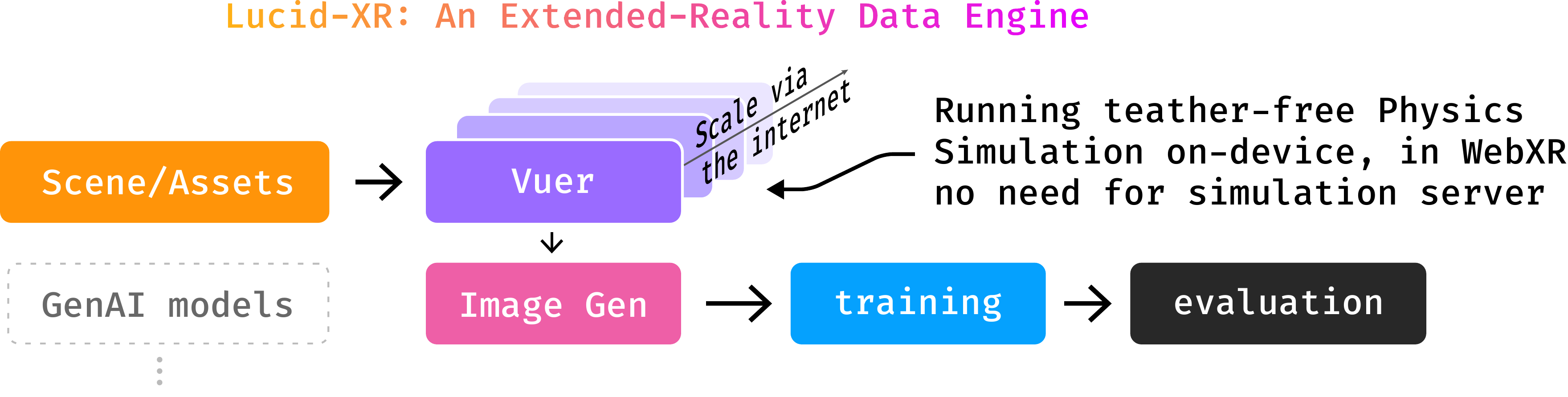}%
\caption{
\textbf{System Schematic of the Lucid-XR Pipeline.} 
The results in this paper require hand-crafted, but basic 3D scenes. The data collection is done collectively by the authors. The resulting simulated datasets are augmented by a generative pipeline powered by language and text-to-image models. 
}\label{fig:enter-label}
\end{figure}

\section{A Touch of Physics in Extended Reality (XR)}

Modern internet browsers are de facto operating systems built according to a commonly agreed-upon programming interface. The key limitation is the constraint on compute and memory, as the device's operating system needs to throttle the browser process to maintain a smooth user experience while handling a multitude of system and application tasks. In the past decade, however, the competition between vendors has significantly improved browser performance; on newer extended-reality devices, the browser receives first-class support in feature sets, system resource limits, and performance.

Lucid-XR builds on three web standards. First, to ensure simulations are \textit{fast}, we compile MuJoCo into WebAssembly, bypassing the single-threaded V8 engine and achieving near-native speed directly on XR hardware. Second, we built upon the web-XR standard, a user-interaction programming interface that is shared across device vendors. This enabled us to support \textit{hands}, \textit{motion controllers}, and XR devices from as low as \$300 to \$4000. The third web standard is webGL. We wrote a performant rendering, interaction, and programming interface from scratch using react-three/fiber, a modern lightweight and performant 3D framework with strong community backing. This has enabled real-time visualization of complex scenes and deformable objects. The current technological landscape is fast-evolving --- in the near future, when \href{https://www.w3.org/TR/webgpu/}{webGPU} becomes fully supported, we will be able to parallelize the physics simulation using compute shaders with hardware acceleration. This will enable more complex physics that involve a large quantity of interacting particles, deformable
vertices, or liquid.

\begin{figure}%
\centering%
\raisebox{15pt}[0pt][0pt]{%
\includegraphics[width=0.47\linewidth]{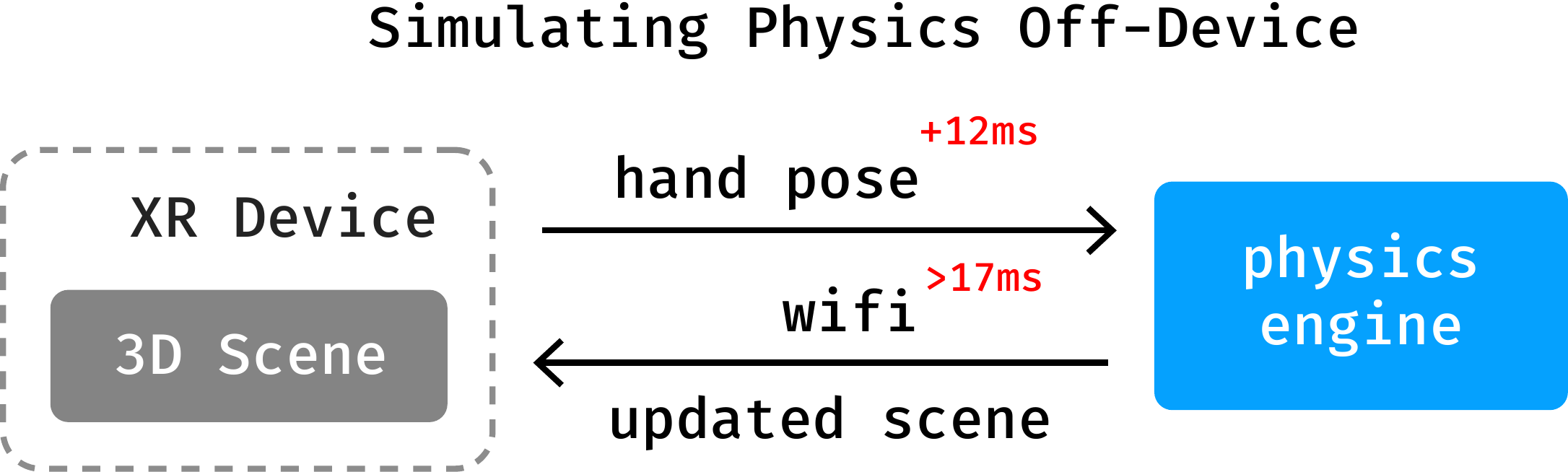}}%
\hfill%
\includegraphics[width=0.47\linewidth]{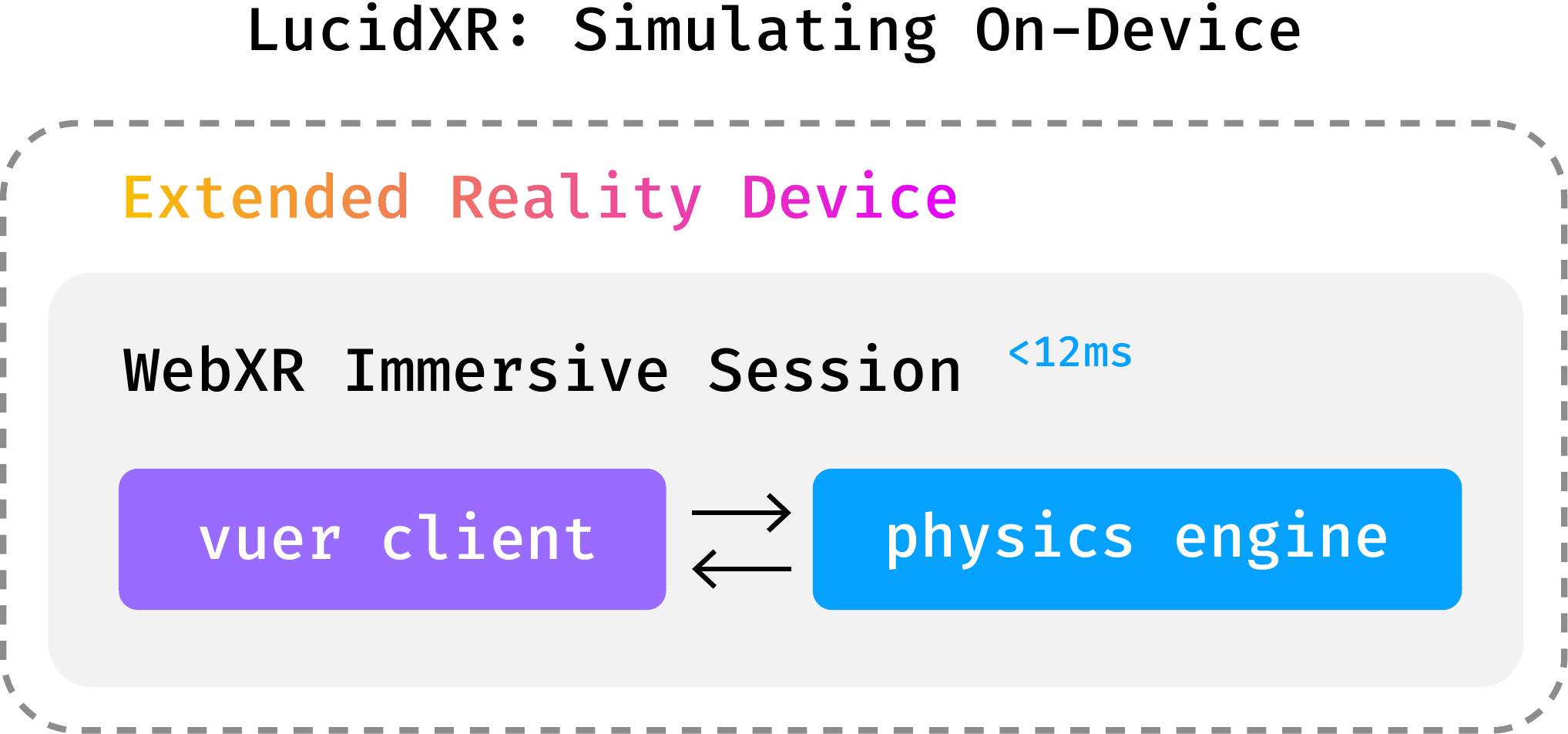}
\caption{
\textbf{Moving physics simulation on-device enables untethered access to immersive simulations.} 
The key benefits are twofold: first, it enables the simulation of deformable objects that involve modifying a large amount of mesh data that are too slow to send over WiFi. Second, it \textit{eliminates} delays due to network latency, allowing the simulation to run at the device's native frame rate.}
\label{fig:on-device}
\end{figure}

\subsection{Multi-Physics Simulation in Vuer}

At the core of Lucid-XR is \textit{vuer}, a simulator-agnostic XR framework running in the browser. We use MuJoCo~\cite{todorov2012mujoco} compiled to WebAssembly with a custom react-three/fiber front-end, achieving real-time simulation at 90\,fps on Apple Vision Pro. Data is collected at 25\,fps; each step takes under 12\,ms to simulate, so latency is negligible unless overloaded. The MuJoCo \textit{decimation} parameter (5–20 steps per frame) controls simulation fidelity. Rendering runs natively on the XR device at 90\,fps for user comfort. Like Blender, \textit{vuer} is simulator-agnostic and benefits as physics engines improve.

\begin{figure}[t]
\centering
\includegraphics[width=\linewidth]{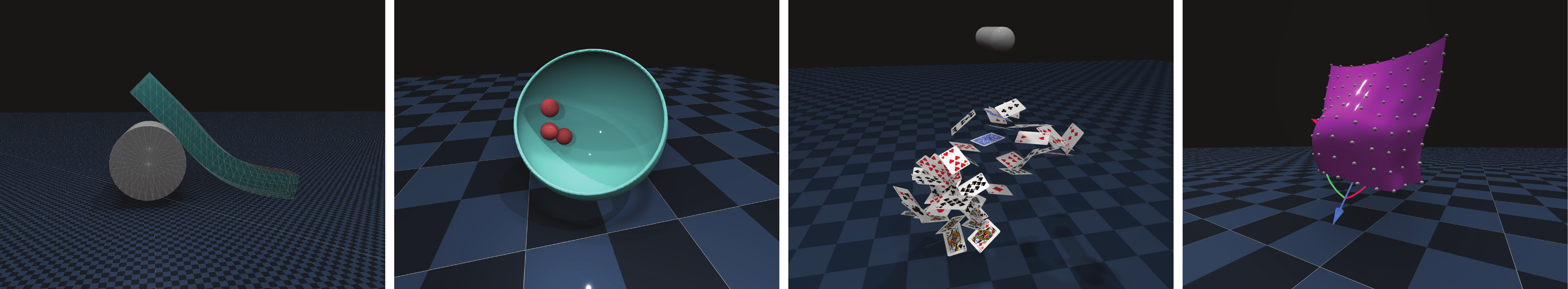}\\
\caption{\textbf{Multiple types of physics running natively inside the browser in real-time.} From left to right: flexible material interacting with a solid; Signed-distance function (SDF) based collision solver for non-convex shapes; A deck of cards interacting with air/wind; Soft skin material interacting with a solid. Rendering and simulation are both native in a web browser. }
\label{fig:multi-physics}
\end{figure}

\textbf{Flexible materials.} Prior VR setups run physics on an external computer~\cite{Park2024DexHubAD}, which introduces latency when simulating flexible or particle-rich scenes. Running physics on-device eliminates this bottleneck and enables interactive deformable-object simulation (Fig.~\ref{fig:multi-physics}).

\textbf{Fluid forces.} MuJoCo’s integrated fluid model~\cite{mujoco_fluid_forces_2025} supports wind and fluid interactions with rigid and deformable objects. For example, Fig.~\ref{fig:multi-physics}c shows poker cards falling against a capsule under air resistance.  

\textbf{Collision without convex decomposition.} We include the signed-distance function (SDF) collision solver in Mujoco by default, removing the need for convex decomposition during modeling. This simplifies setup, at the cost of higher runtime compute and memory.

\subsection{Precise Interactions at A Distance: Hitchhiking Controllers}

The virtual world presents an opportunity to define new ways of interacting with robots beyond the constraint of real-world physicality. When the robot is situated far from the user's home location in VR, na\"ively controlling those grippers by ``grasping'' them from afar yields poor user experience because the hand tracking error gets amplified as the distance increases. We solve this problem by applying SE(3) transformation to the target gripper in the local frame of the mocap site. This design took inspiration from the \textit{hitchhiking hand}~\cite{ban2023hitchhiking}, and let the user activate MoCap sites at a distance by looking at the object and clicking on it. Once activated, the user can engage with the site via a natural grasping gesture, where they close the lower three fingers.


\subsection{On-Device Retargeting for Dexterous Hand Control}


The problem of retargeting kinematic poses between humans and robots across different body kinematics is a common problem shared between dexterous manipulation, locomotion and humanoid whole-body control~\cite{cheng2024exbody}. Existing solutions rely on running the kinematics solver separately on a server~\cite{cheng2024tv}, making it hard to scale. Our solution involves binding mocap sites to the tip of each finger and utilizing the relative pose to the wrist joint, as well as the action space.

\begin{figure}
\centering
\includegraphics[width=0.6\linewidth]{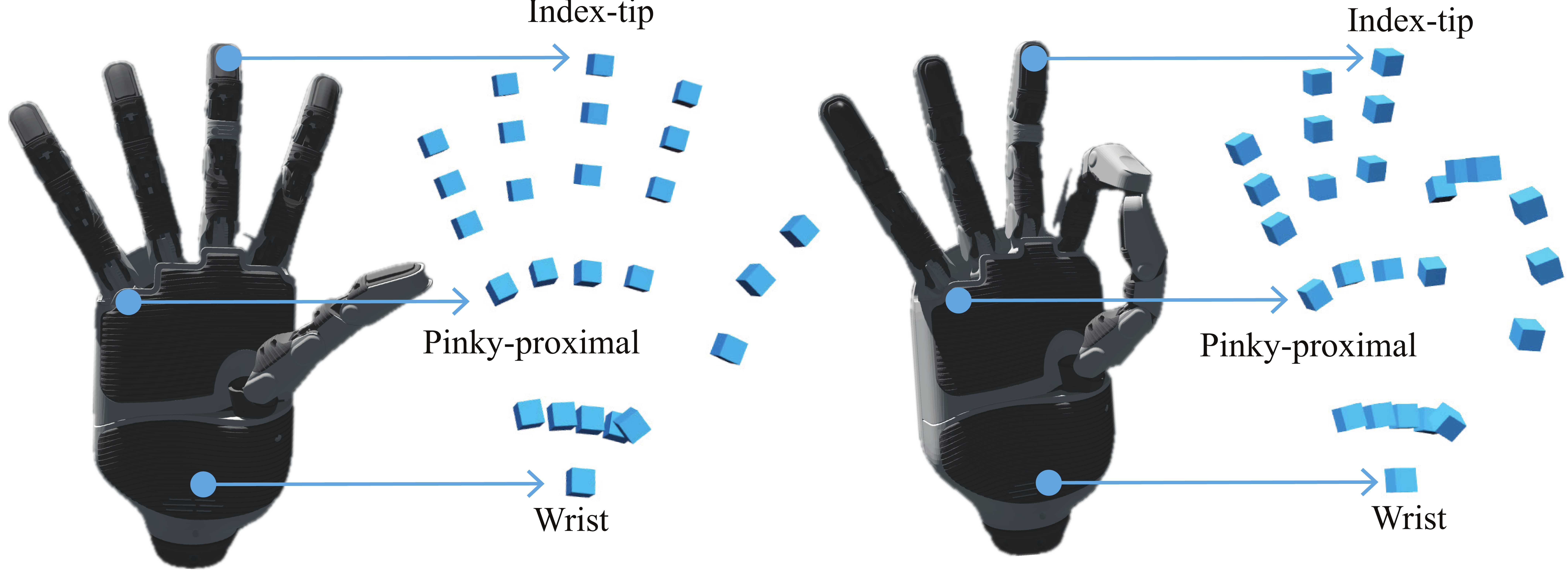}
\caption{
\textbf{Controlling Hand Pose via Motion Caption Sites.}
We specify mocap sites by first aligning the proximal joints, and scale the hand so that fingers are similar in size as the robot hand. We then weld the SE(3) pose of the fingertip to similar sites on the robot hand and the wrist. We adjust the torque scale to balance tracking of the position and the rotation portion of the pose.}
\label{fig:hand-mocap-control}
\vspace{-1em}
\end{figure}

\begin{figure}%
\centering%
\includegraphics[width=\linewidth]{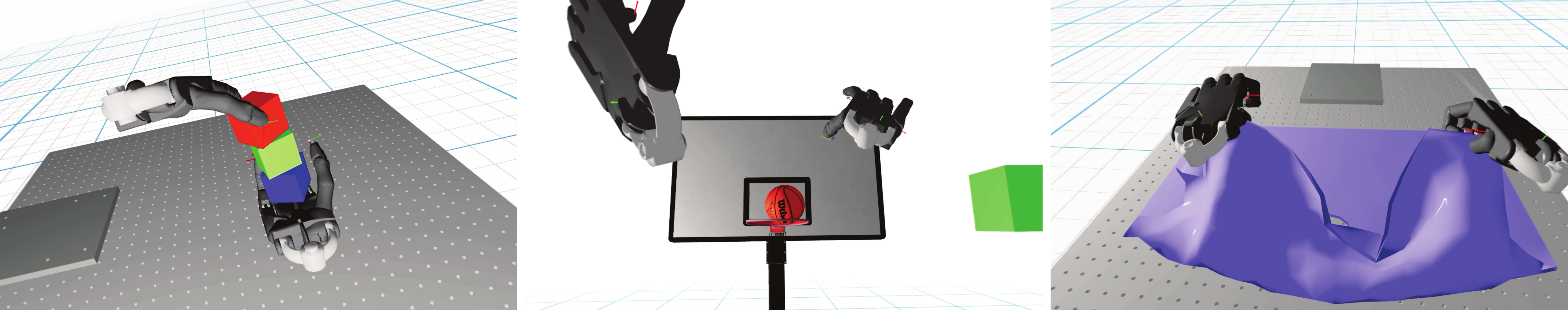}
\caption{
\textbf{Handling dynamic tasks and deformable objects.}
The on-device retargeting is accurate enough to balance three blocks on one hand; is fast enough to handle dynamic tasks such as throwing a basketball, and handle deformable objects for cloth folding.}
\label{fig:hand-retargetting}
\end{figure}

This hand control scheme is very general. We found that it worked well for all of the robot hands that we experimented with. We designed a schema-based programming interface for users to specify custom bindings between mocap sites or geometric bodies with landmarks and gestures of the hand in python.

\subsection{Porting Existing Environments} 
\begin{figure}[h]
\vspace{-0.5em}
\includegraphics[width=\linewidth]{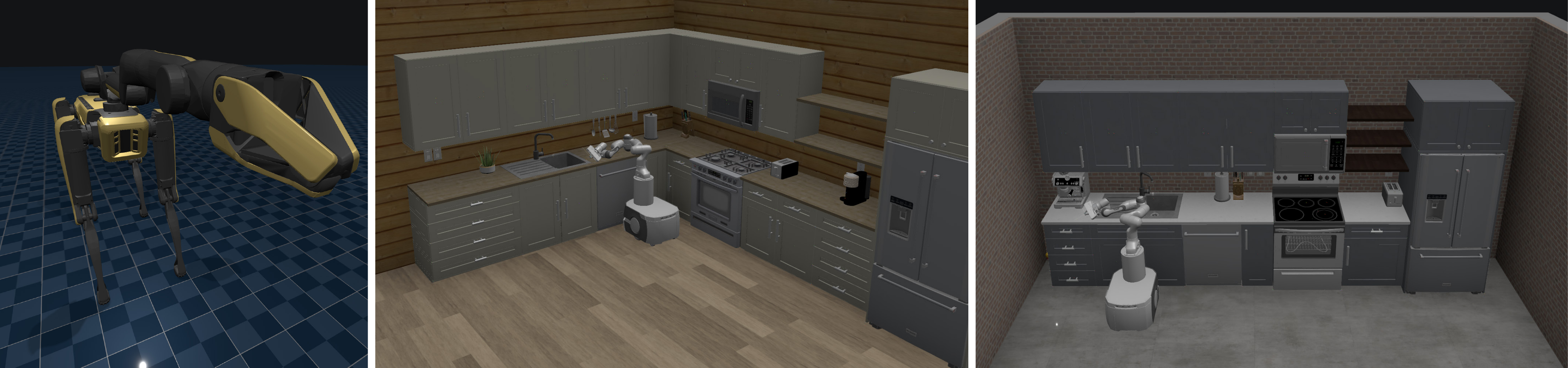}
\caption{(a) Spot robot from Menagerie (b - c) RobotCasa Scenes}
\label{fig:export_envs}%
\end{figure}

Take Robocasa as an example -- you can extract the XML from MuJoCo by calling \texttt{env.get\_xml}, and then iterate through the \texttt{`file=`} attributes to collect all of the assets. You can then load this file bundle by simply dragging and dropping it into Vuer. We were able to extract scenes from RoboHive, RoboCasa, RoboSuite, and MuJoCo Menagerie (see \ref{fig:export_envs}) . 

\section{Synthesizing Diverse Manipulation Data from Virtual Demonstrations}

The third component of Lucid-XR is a generative engine that converts the virtual human demonstrations collected in the sim into diverse and realistic-looking multiview image datasets for the robot. 

\subsection{Generating Realistic Images from Virtual Demonstrations.}
\label{sec:augmentation}
\begin{figure}[h]
    \centering
    \includegraphics[width=0.85\linewidth]{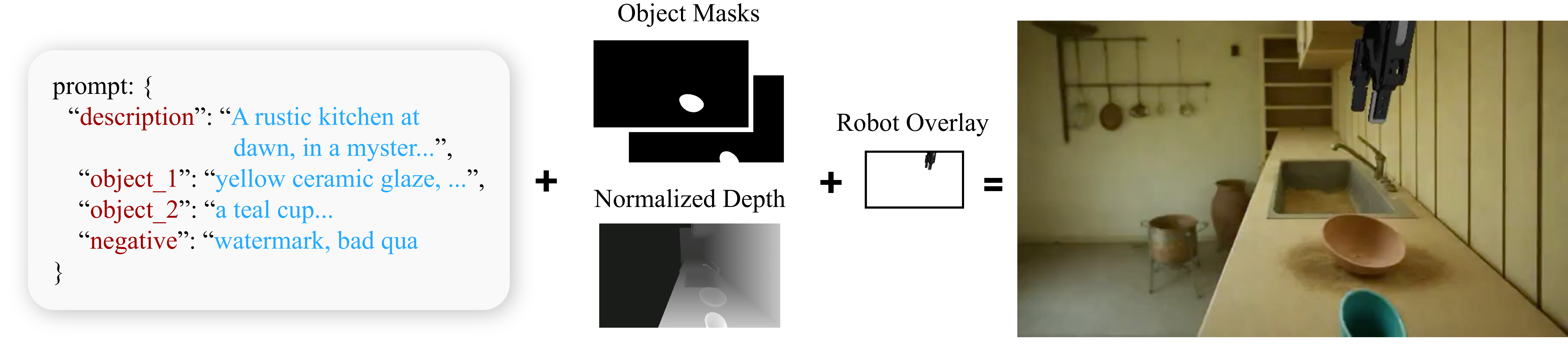}
    \caption{
    \textbf{Image generation pipeline.}
    We apply semantic masking and depth conditioning to precisely control the scene.
    }
    \label{fig:image-generation}
\end{figure}

Figure~\ref{fig:image-generation} shows our setup for generating realistic-looking images. We follow the LucidSim~\cite{lucidsim} recipe that starts with a collection of diverse text prompts collected from chatGPT, and use the semantic mask labels from the physics simulation to control the image generation process. Prompts for generation are sourced en masse from ChatGPT via a meta-prompt (see appendix). Prompts for the background tend to be more complex. In alignment with observations made by prior works~\cite{lucidsim}, we found it is key to generate these images from a diverse set of text prompts.

\subsection{Demonstration Augmentation} 
We designed an interface to generate MuJoCo XMLs procedurally in Python, allowing for a user to quickly compose scenes in LucidXR and multiply data. This allows robust control of the distribution of initial scene configurations during data collection. 

\textbf{Repositioning cameras post-demonstrations} Camera poses can be modified very quickly in the Python script without needing data to be recollected. Collected trajectories can simply be played back and rendered from the new camera. This allows for easy visualization and quick iteration of image views for training. The simulator also allows for rendering the ground truth depth, which we use to warp images post-rendering. Given the exact camera extrinsics, intrinsics, and depth we can compute the optical flow for nearby poses. This is a crucial part of the sim-to-real pipeline to prevent sensitivity to the camera pose.

\textbf{Trajectory warping for repositioning objects and robots}
Similar to MimicGen \cite{mandlekar2023mimicgendatagenerationscalable} we select keypoints in the trajectories,  and move them around within a specified distribution. With linear interpolation for pose + spherical interpolation for rotation, this creates an entirely new demonstration. This allows us to reposition objects, robots, and change the distribution of initial scene setup post-demonstration. We use this augmentation to make trained policies more robust to variations in object position.

\section{Results}



\begin{figure}[h]%
\centering%
\begin{subfigure}[b]{0.198\textwidth}
\includegraphics[width=\textwidth]{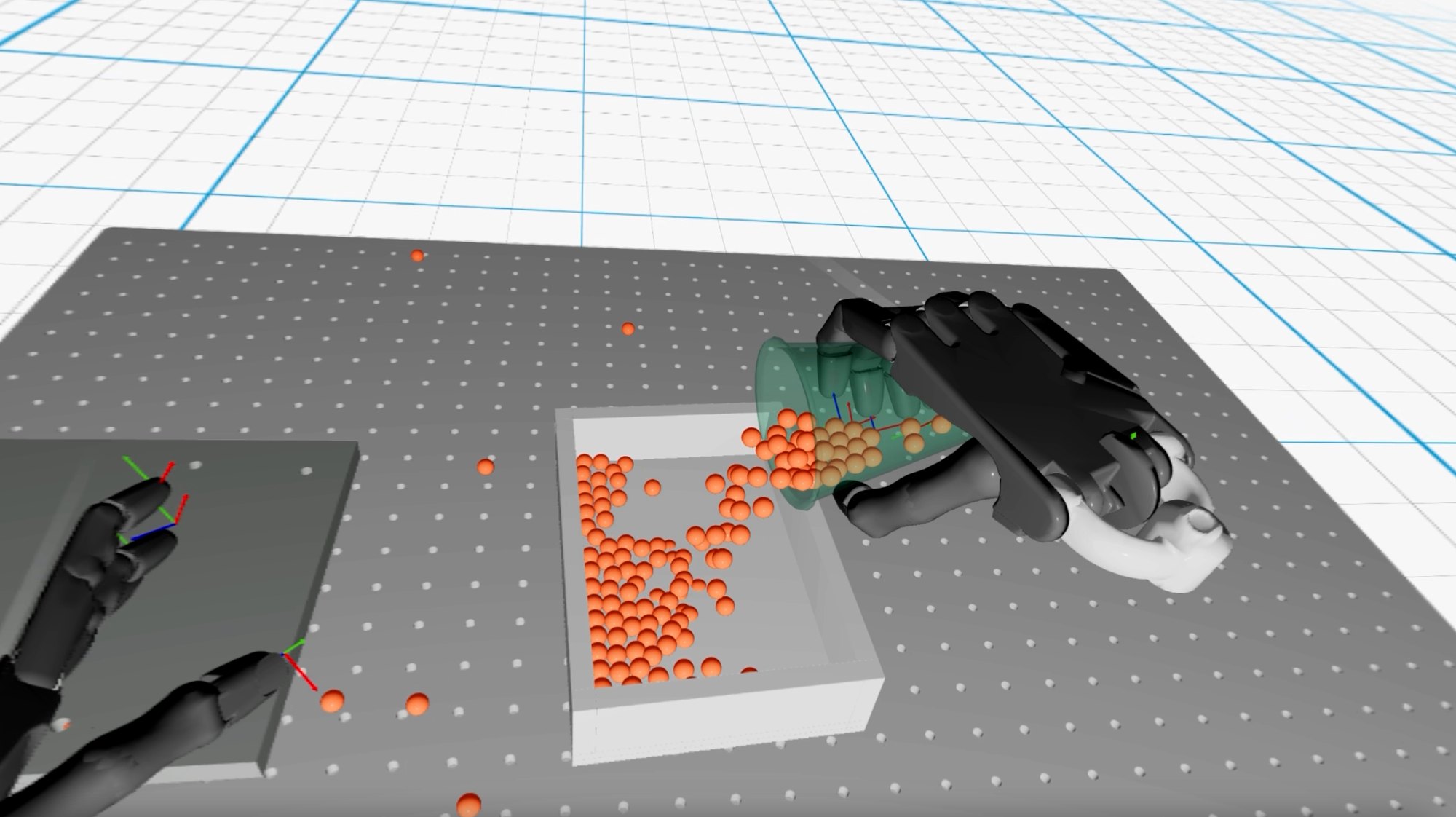}
\caption{Pouring cup.}
\label{fig:pouring}
\end{subfigure}%
\hfill%
\begin{subfigure}[b]{0.198\textwidth}
\includegraphics[width=\textwidth]{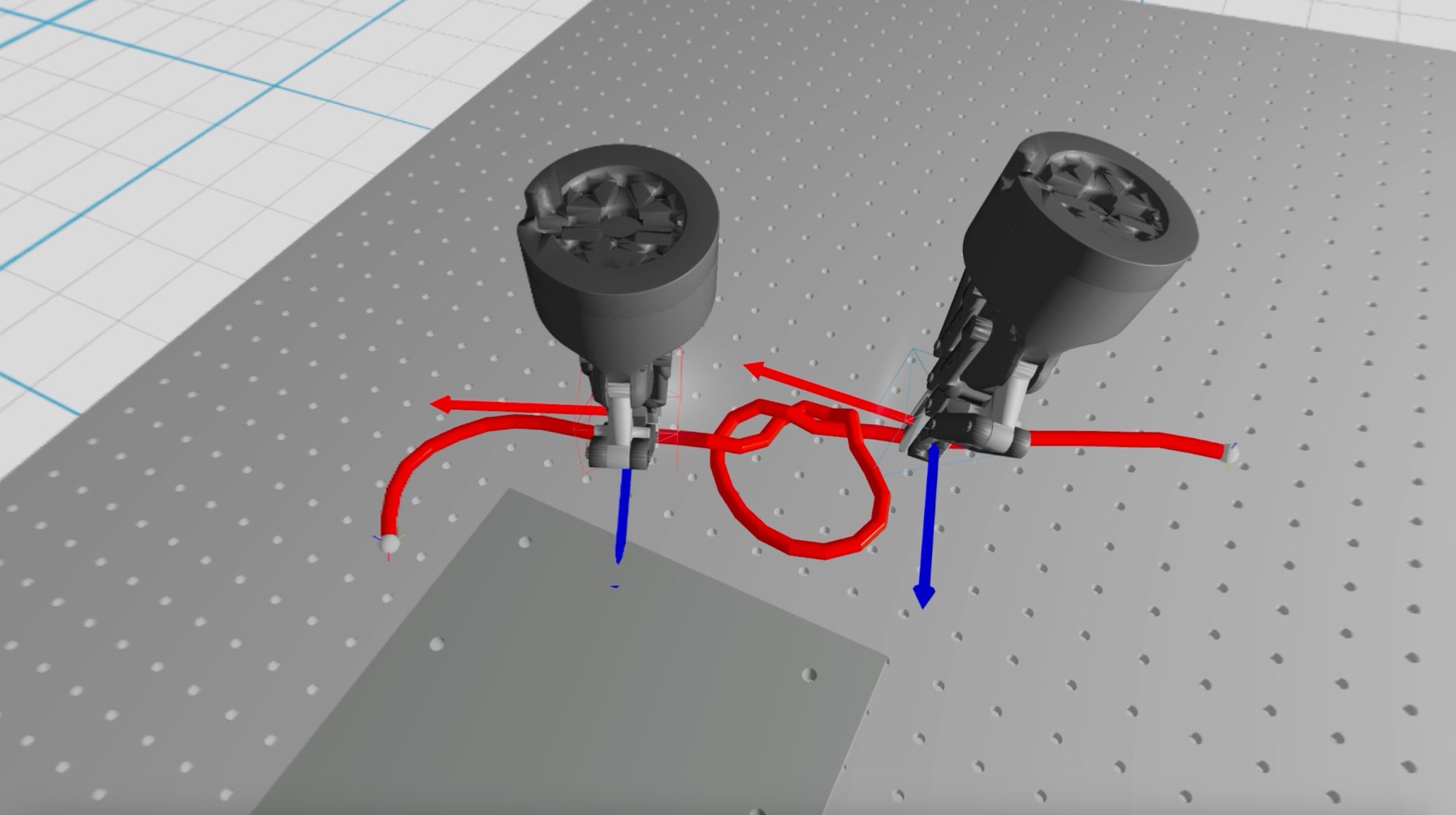}
\caption{Tying a knot.}
\label{fig:knot}
\end{subfigure}%
\hfill%
\begin{subfigure}[b]{0.198\textwidth}
\includegraphics[width=\textwidth]{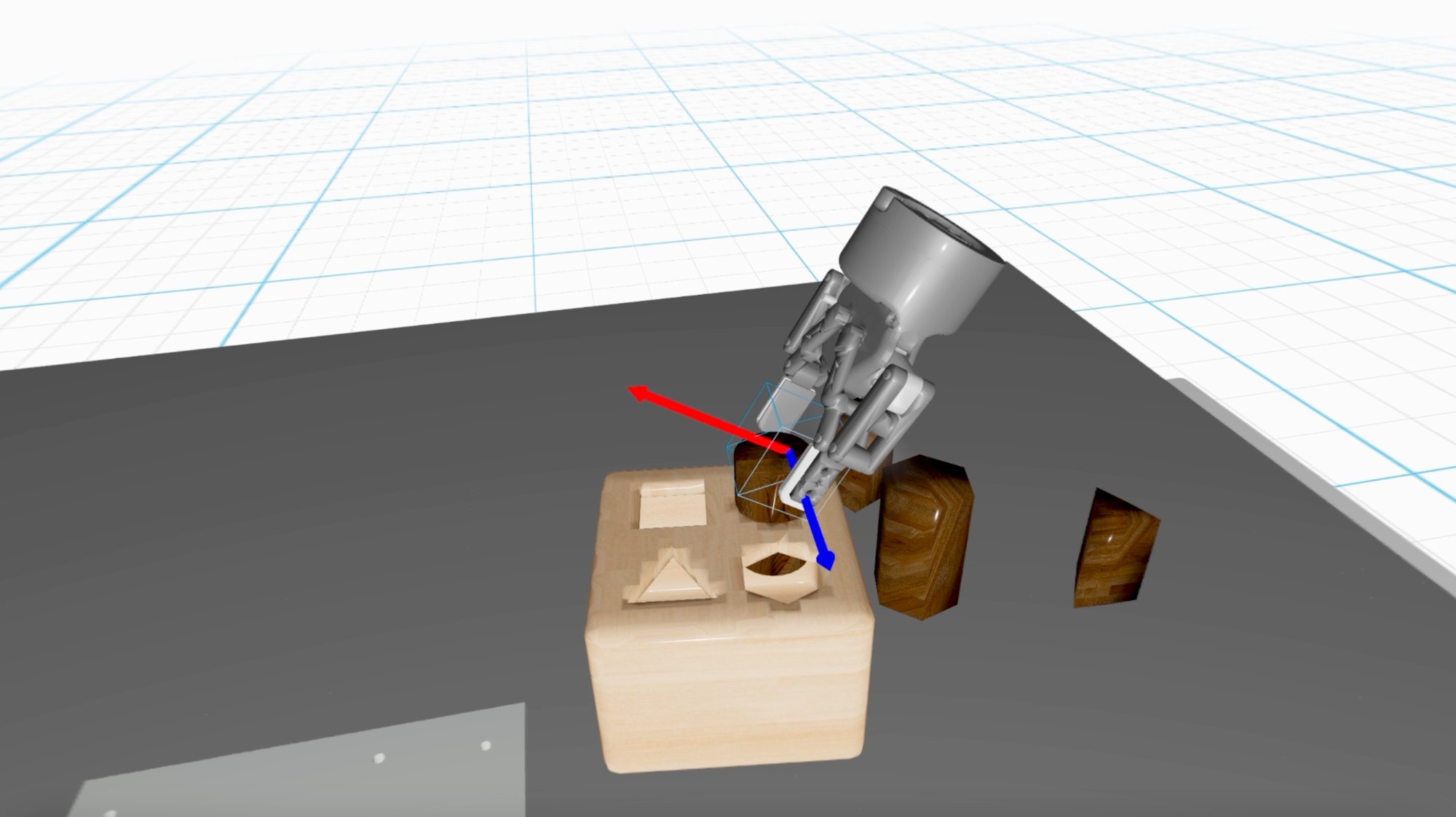}
\caption{Tight insertion}
\label{fig:sort_shapes}
\end{subfigure}%
\hfill%
\begin{subfigure}[b]{0.198\textwidth}
\includegraphics[width=\textwidth]{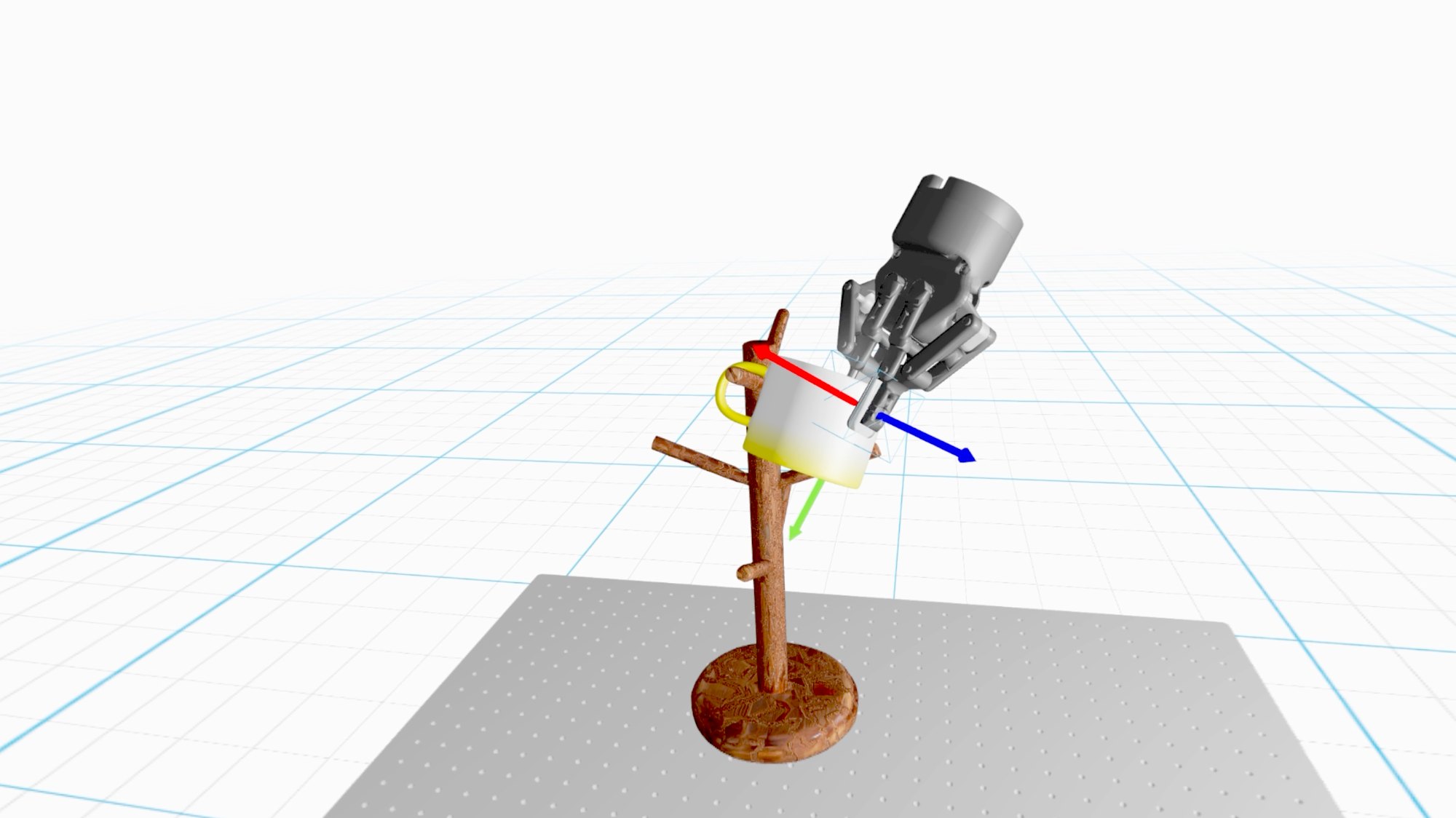}
\caption{Mug tree}
\label{fig:mug_tree}
\end{subfigure}%
\hfill%
\begin{subfigure}[b]{0.198\textwidth}
\includegraphics[width=\textwidth]{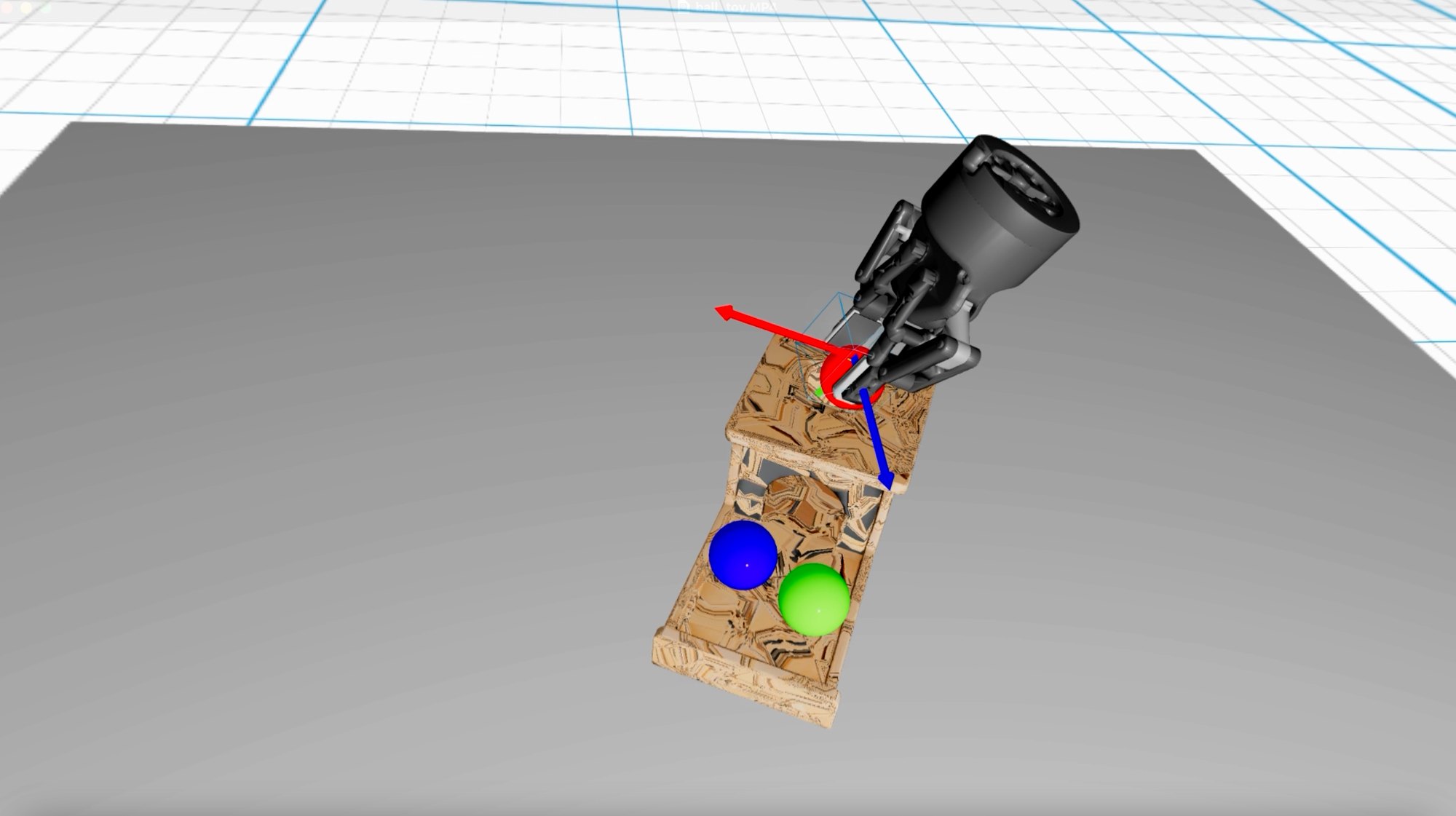}
\caption{Ball sorting toy}
\label{fig:ball_sorting_toy}
\end{subfigure}
\caption{
\textbf{Lucid-XR can simulate contact-rich manipulation of diverse physics.}
(a) granular materials to simulate pouring liquid, (b) deformable materials when tying a knot, (c) tight tolerances between objects in contact, in a shape insertion task. (d) placing a mug onto a drying rack (e) a ball sorting toy.
}
\label{fig:Diverse Tasks}
\end{figure}

We created contact-rich environments to evaluate various types of on-device physics simulation in Lucid-XR. Each environment tests a distinct type of physical interaction (see Fig.~\ref{fig:Diverse Tasks}). For each task the initial state is randomized within a small area.
\begin{itemize}
  \setlength\leftskip{-2.5em}       
  \item \textbf{Block Stacking:} This task requires creating a three block stack with a dextrous hand.
  \item \textbf{Pour Liquid:} This task requires picking up a cup with a dextrous hand, handing it off to another hand, and pouring into a sink -- involves a large scene, and granular particles flow.  
  \item \textbf{Ball Sorting:} This task requires sorting three balls by color into a toy -- Involves mixed rigid–particle collision in a toy sorter. 
  \item \textbf{Knot Tying:} This task requires tying a knot on a suspended rope with a single two-fingered gripper -- Involves deformation of a flexible string into a knot with self-contact
  \item \textbf{Kitchen-Sink:} This task requires picking up a cup, placing it onto a bowl, and then placing both into the kitchen sink  -- Involves a much larger scene, using SDFs, and is longer-horizon.
  \item \textbf{Mug Tree:} This task requires picking up a mug with a single two-fingered gripper and placing it on a tree rack  -- Involves modeling collisions with concave shapes using the SDF plugin,  
\end{itemize}



\subsection{Data Collection and Learning Setup}

\begin{wrapfigure}{r}{0.45\textwidth}
    \centering
    \vspace{-1em}
    \includegraphics[width=\linewidth]{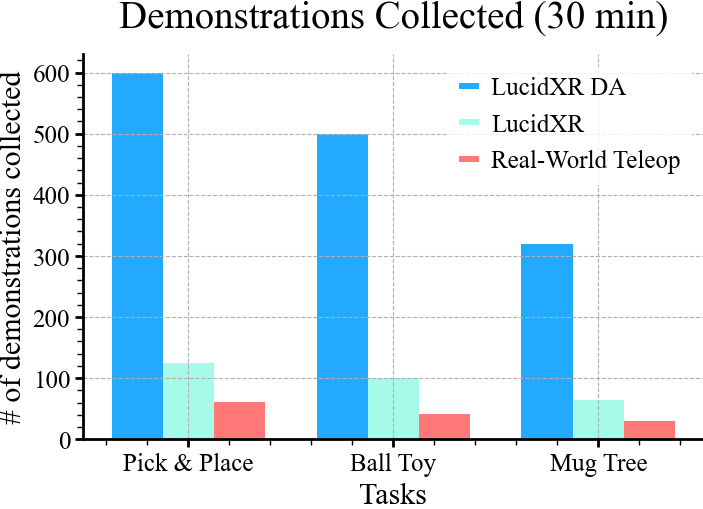}
    \caption{\textbf{Amount of data collected for LucidXR vs Real-World Teleop.}}
    \label{fig:data_collection_speed}
\end{wrapfigure}%
We record observations and actions as SE(3) mocap poses at 25\,Hz; rotations use the 6D representation of~\cite{Zhou2019rotation-rep}. Demonstrations are embodiment-agnostic, so behavior-cloned policies can deploy on any two-finger gripper. Policies take proprioception as well as either a wrist RGB view or three fixed RGB views of the workspace and output chunked absolute end-effector poses; we train with batch size 64.

\textbf{ACT.} We use standard ACT~\cite{zhao2023learning}: 15k updates, chunk size 25, learning rate $10^{-4}$, and a DETR-style VAE backbone~\cite{carion2020endtoendobjectdetectiontransformers} (4 encoders, 1 decoder, head dim 128, FFN 256). Images are color-jittered. At evaluation we apply temporal aggregation~\cite{zhao2023learning}.

\textbf{Diffusion.} We also train a score-based action denoiser~\cite{garipov2023compositionalsculptingiterativegenerative,chi2024diffusionpolicy} with a 1D U-Net~\cite{janner2022planningdiffusionflexiblebehavior} conditioned on image features via FiLM~\cite{perez2017filmvisualreasoninggeneral}. Optimization uses AdamW with an exponential LR schedule $10^{-3}\!\rightarrow\!10^{-5}$. Inference performs 1000 denoising steps and executes actions by chunks.


\subsection{Comparing Data Collection Speed in Virtual and the Real-world}
Resetting the scene in the real world involves manually replacing the object, moving the gripper, and safety checks. In vuer the entire physics simulation and reset logic runs on-device, so the user can reset the environment by pressing a button. This provides an uninterrupted data capture for the full thirty minutes. We collected \(30\) minutes of demonstrations for each of three tasks in both settings. As shown in~\ref{fig:data_collection_speed}, participants collected roughly \(2\times\) more demonstrations in Lucid-XR than with real-world teleoperation. When we further applied the augmentation pipeline, the effective dataset size increased to about \(5\times\) that of the real-world baseline.

\subsection{Real-to-Sim Evaluation}

\begin{figure}[t]
    \centering
    \begin{subfigure}[t]{0.65\linewidth}
        \centering
        \includegraphics[width=\linewidth]{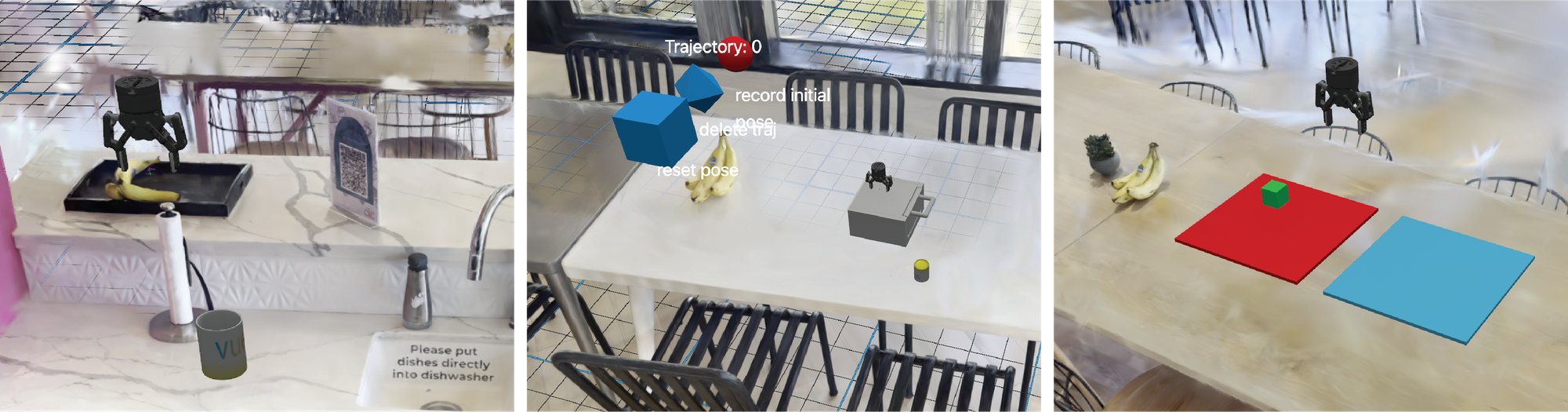}
        \caption{
            \textbf{Real-to-sim evaluation.}
            We made 3D Gaussian scans of real-world kitchen environments to stage the evaluation environments, as a rough estimate of the generalization capability of the trained policy. 
        }
        \label{fig:real-to-sim-subfigure}
    \end{subfigure}
    \hfill%
    \begin{subfigure}[t]{0.30\linewidth}
        \centering
        \includegraphics[width=\linewidth]{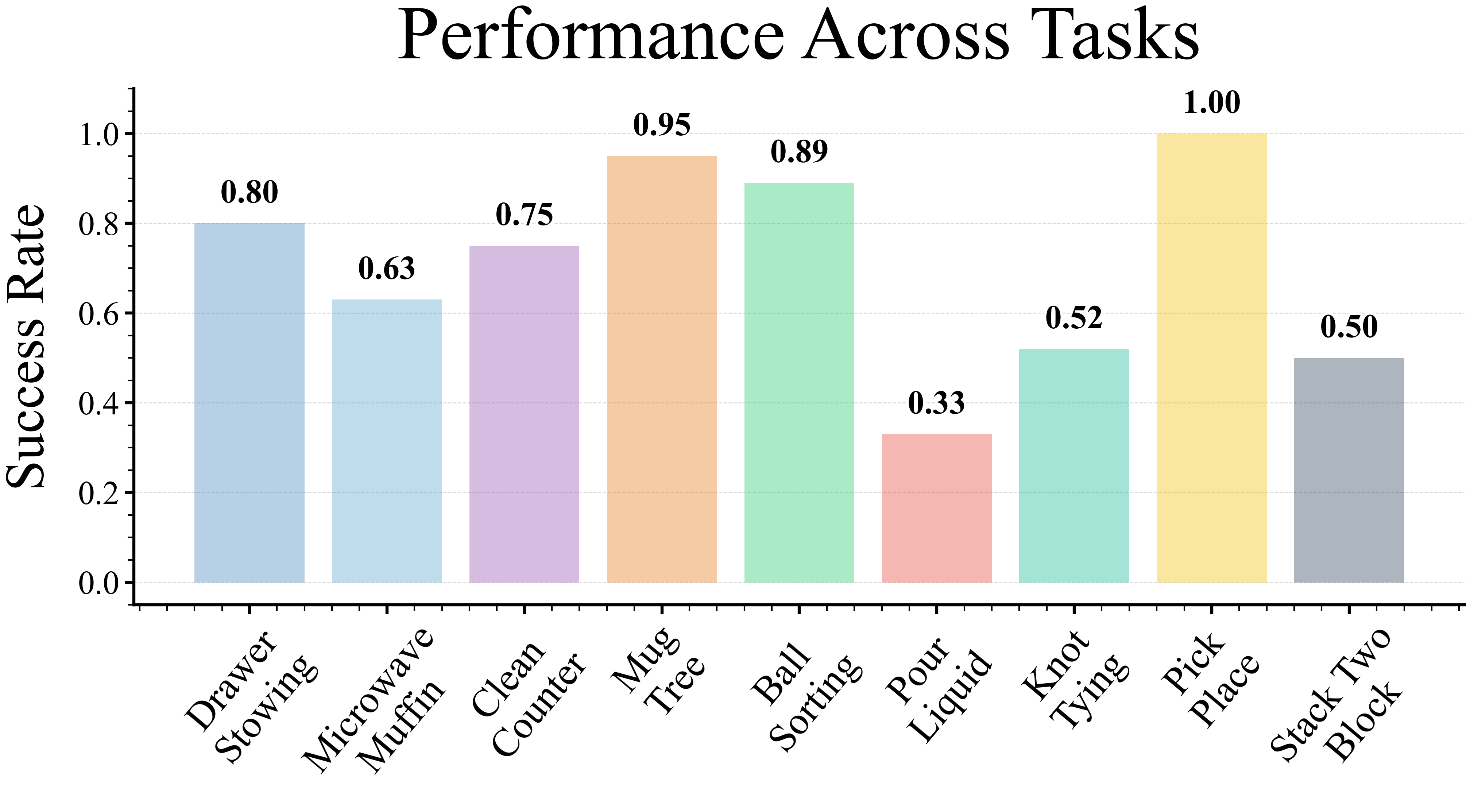}
        \caption{
            Success rate for each manipulation task, as measured in our extended‐reality data engine.
        }
        \label{fig:task-success-subfigure}
    \end{subfigure}
    \caption{Evaluation results from real-to-sim transfer (left) and per-task performance in simulation (right).}
    \label{fig:sim_to_sim_eval}
\end{figure}

We made a small number of 3D Gaussian models of kitchens with real-life clutter, and use them as staging environments for our simulated eval. Although the Lucid-XR policy has never seen data from these environments, it out-performed the same policy trained on simulated data (see ~Fig.\ref{fig:sim_to_sim_eval}).


\subsection{Sim-To-Real Evaluation}

\begin{figure}[t]
    \centering
    \begin{subfigure}[t]{0.4\linewidth}
         \centering
        \includegraphics[width=\linewidth]{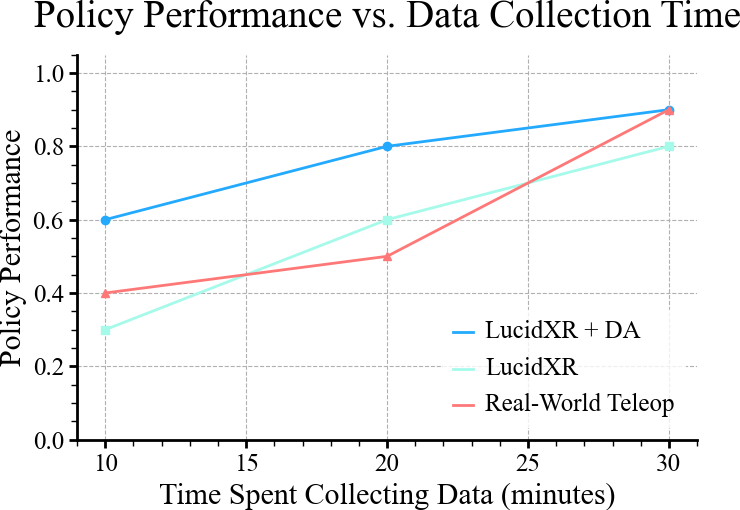}
        \caption{
        Success rates for sim-to-real and real-to-real policies}
        
        \label{fig:sim-vs-real}
    \end{subfigure}
    \hfill%
    \begin{subfigure}[t]{0.4\linewidth}
        \centering
        \includegraphics[width=\linewidth]{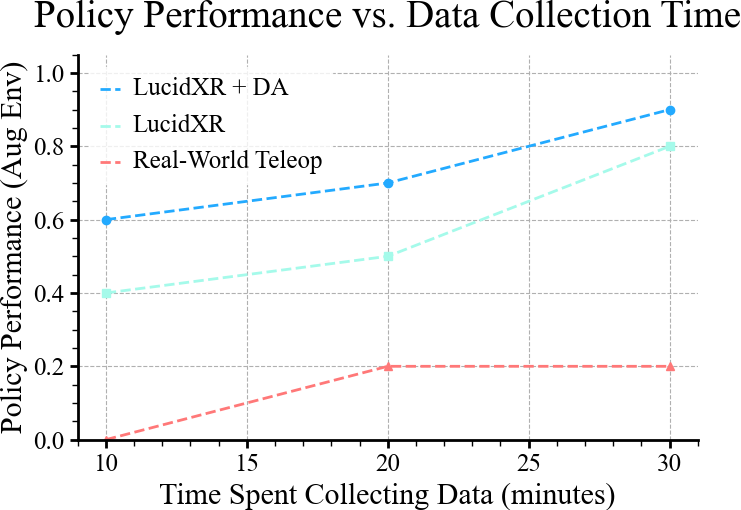}
        \caption{
            Success rates for sim-to-real and real-to-real policies on Augmented environment
        }
        \label{fig:sim-vs-real-augmented}
    \end{subfigure}
    \caption{}
    \label{fig:side_by_side_eval}
\end{figure}

We trained three policies to accomplish pick and place using datasets collected over 10, 20, and 30 minutes. For each duration, one dataset was collected in LucidXR and rendered with generative imagery and another via real robot teleoperation  where an Oculus headset was used to control the end-effector position. We further augmented the LucidXR dataset with methods from \ref{sec:augmentation}. We then evaluated all policies on a real robot in the same environment that the real dataset was collected on. Results showed that policies trained on purely synthetic data performed comparably to those trained on real data (see Fig.~\ref{fig:side_by_side_eval}).

To further test robustness, we repeated the evaluation under modified visual conditions. Specifically, we varied the lighting, color, and replaced the original wooden tabletop with either a textured tablecloth or a black tablecloth. Under these changes, the policy trained only on real-world demonstrations failed to generalize, while the LucidXR-trained policies maintained high success rates and outperformed the real-world baseline(see Fig.~\ref{fig:side_by_side_eval}).

\begin{table}[t]
    \centering
    \renewcommand{\arraystretch}{1.2}  
    \begin{tabular}{@{} l l l l @{}}
        \toprule
        \textbf{Kitchen Clearing} & \textbf{Base Env.} & \textbf{Low Clutter} & \textbf{High Clutter + Noise} \\
        \midrule
        ACT Policy         & 100\% & 0\%          & 0\%          \\
        ACT + LucidSim     & 100\%           & 90\%          & 25\%     \\
        \bottomrule
    \end{tabular}%
    \caption{Evaluation scores for the kitchen clearing task with unseen real-life meshes overlaid on the original environment. We score the Kitchen Clearing task based on two pick-and-places, allowing for 4 points per run.}
    \label{tab:task_results}
\end{table}

\section{Related Works}

\textbf{Extended reality for robot teleoperation and data collection.}

XR environments have been used to scale robot learning via teleoperation, yet each faces bandwidth or realism limits. \cite{ban2023hitchhiking} introduced “hitchhiking hands” for remote interaction, letting users control virtual avatars by gaze in VR. Our hitchhiking controller builds on this idea, enabling distant robot gripper control through XR hands. Hand-overlay methods~\cite{chen2024arcap, nechyporenko2024armadaaugmentedrealityrobot, 10661067, yang2024arcadescalabledemonstrationcollection} and mobile-AR approaches~\cite{duan2023ar2, wang2024eveenablingtrainrobots} overlay virtual robots on physical objects, but lack rich contact. Other AR/VR frameworks~\cite{cheng2024tv, vanhaastregt2024puppeteerrobotaugmentedreality, iyer2024openteachversatileteleoperation, 10.1007/s10846-021-01311-7} enable seamless teleoperation of physical robots, but throughput and environment diversity remain limited. Most closely related to our work are \cite{jiang2025irisimmersiverobotinteraction, DART}, toolkits for VR demo collection. A key difference is that LucidXR runs simulations on-device, avoiding cloud latency that hinders dexterous, dynamic control in systems like DART and IRIS. We envision LucidXR as a platform for scaling imitation learning with generative models. Works such as \cite{zhao2023learning} provide foundations for architectures like the Action Chunking Transformer.

\paragraph{Generative AI for Synthetic Data Augmentation.}
Recent works use generative AI across text, vision, and 3D assets to expand robot training data. Our image-generation pipeline builds on LucidSim~\cite{lucidsim}, which showed that locomotion policies trained entirely from generated images can deploy zero-shot in the real world. LucidSim uses text prompts and diffusion models to produce diverse scenes; we extend this with more semantic classes and randomized object geometry sampled from 3D assets. Other works~\cite{katara2023gen2simscalingrobotlearning, wang2024robogenunleashinginfinitedata, ma2024dreurekalanguagemodelguided, chen2024urdformer} generate diverse tasks but still lack realistic textures. These can be integrated into our pipeline to further expand data realism. In parallel, \cite{yu2023scalingrobotlearningsemantically, mandi2022cacti, jang2025dreamgenunlockinggeneralizationrobot, chen2023genaug} apply text-to-image generation to hallucinate new scenarios directly on real data, maintaining semantic consistency with robot objectives.

\textbf{Large-scale imitation learning and data aggregation.}
Multi-institution efforts aggregate demonstrations but remain embodiment-specific. Open X-Embodiment collects 50k+ demos for RT-X policies~\cite{open_x_embodiment_rt_x_2023}; RH20T~\cite{fang2023rh20t} and DROID~\cite{khazatsky2024droid} add 100k+ crowdsourced trajectories. Despite scale, diversity across robots, environments, and tasks remains limited, falling short of what is needed for truly generalist policies analogous to vision or NLP foundation models.

\section{Conclusion}

In this work, we present Lucid-XR, a generative-AI-powered learning pipeline for producing generalizable visual policies for manipulation. We demonstrate the potential for virtual demonstrations in a simulated world to produce real-world robot policies that generalize across object instances, appearance, and lighting conditions. We also present new designs for interacting with virtual robots through VR controllers and hand gestures. 
We believe virtual demonstration data has the potential to scale across the internet and close the data gap for training a generally capable robot foundation model.

\textbf{Deploying Across Embodiments}. Our observation is that as long as the data is collected on the full embodiment, the learned controller will work, and in many cases data collected with embodiment-free (a floating gripper) transfers readily. We note that cross-embodiment transfer in Lucid-XR is limited only by inverse-kinematics and mobility. This is an avenue that future work can explore.

\acknowledgments{This work was supported by Amazon.com Services LLC, Award \#2D-06310236, and by the Defence Science Technology Agency, Singapore, and by the Toyota Research Institute. We would like to thank the reviewers for their constructive criticism and for helping make this paper more comprehensive.}

\section{Limitations}

This work leaves a few rocks untouched. For instance, we rely on the controlling power of the same text prompt to generate consistent views. A key benefit of generating visual data is that it comes with paired text labels. Future iterations can benefit from the additional supervision that this pairing provides.


\bibliography{main}  

\begin{thebibliography}{41}
\providecommand{\natexlab}[1]{#1}
\providecommand{\url}[1]{\texttt{#1}}
\expandafter\ifx\csname urlstyle\endcsname\relax
  \providecommand{\doi}[1]{doi: #1}\else
  \providecommand{\doi}{doi: \begingroup \urlstyle{rm}\Url}\fi

\bibitem[Ryu(2007)]{Ryu2007vfx-reality}
J.-H. Ryu.
\newblock {Reality \& effect: A cultural history of visual effects}.
\newblock \emph{Communication Dissertations}, 2007.

\bibitem[Turnock(2009)]{Turnock2009vfx-ILM}
J.~Turnock.
\newblock {Before Industrial Light and Magic: the independent Hollywood special
  effects business, 1968--75: Research Article}.
\newblock \emph{New Rev. Film Telev. Stud.}, 7\penalty0 (2):\penalty0 133--156,
  June 2009.

\bibitem[Das(2023)]{Das2023vfx-evolution}
S.~Das.
\newblock {The evolution of visual effects in cinema: A journey from practical
  effects to {CGI}}.
\newblock \emph{Journal of Emerging Technologies and Innovative Research},
  10\penalty0 (11):\penalty0 303--309, 2023.

\bibitem[Murodillayev(2024)]{Murodillayev2024vfx-impact}
B.~Murodillayev.
\newblock {The impact of visual effects on the cinema experience: A
  comprehensive analysis}.
\newblock \emph{Art Des. Rev.}, 2024.

\bibitem[Todorov et~al.(2012)Todorov, Erez, and Tassa]{todorov2012mujoco}
E.~Todorov, T.~Erez, and Y.~Tassa.
\newblock Mujoco: A physics engine for model-based control.
\newblock In \emph{2012 IEEE/RSJ International Conference on Intelligent Robots
  and Systems}, pages 5026--5033. IEEE, 2012.
\newblock \doi{10.1109/IROS.2012.6386109}.

\bibitem[Park et~al.(2024)Park, Bhatia, Ankile, and Agrawal]{Park2024DexHubAD}
Y.~Park, J.~S. Bhatia, L.~L. Ankile, and P.~Agrawal.
\newblock Dexhub and dart: Towards internet scale robot data collection.
\newblock \emph{ArXiv}, abs/2411.02214, 2024.
\newblock URL \url{https://api.semanticscholar.org/CorpusID:273821640}.

\bibitem[{MuJoCo Documentation}(2025)]{mujoco_fluid_forces_2025}
{MuJoCo Documentation}.
\newblock Fluid forces.
\newblock \url{https://mujoco.readthedocs.io/en/latest/computation/fluid.html},
  2025.
\newblock Accessed: 2025-August-29.

\bibitem[Ban et~al.(2023)Ban, Matsumoto, and Narumi]{ban2023hitchhiking}
R.~Ban, K.~Matsumoto, and T.~Narumi.
\newblock Hitchhiking hands: Remote interaction by switching multiple hand
  avatars with gaze.
\newblock In \emph{SIGGRAPH Asia 2023 Emerging Technologies}, pages 1--2. 2023.

\bibitem[Cheng et~al.(2024{\natexlab{a}})Cheng, Ji, Chen, Yang, Yang, and
  Wang]{cheng2024exbody}
X.~Cheng, Y.~Ji, J.~Chen, R.~Yang, G.~Yang, and X.~Wang.
\newblock Expressive whole-body control for humanoid robots,
  2024{\natexlab{a}}.
\newblock URL \url{https://arxiv.org/abs/2402.16796}.

\bibitem[Cheng et~al.(2024{\natexlab{b}})Cheng, Li, Yang, Yang, and
  Wang]{cheng2024tv}
X.~Cheng, J.~Li, S.~Yang, G.~Yang, and X.~Wang.
\newblock Open-television: Teleoperation with immersive active visual feedback.
\newblock \emph{arXiv preprint arXiv:2407.01512}, 2024{\natexlab{b}}.

\bibitem[Yu et~al.(2024)Yu, Yang, Choi, Ravan, Leonard, and Isola]{lucidsim}
A.~Yu, G.~Yang, R.~Choi, Y.~Ravan, J.~Leonard, and P.~Isola.
\newblock Learning visual parkour from generated images.
\newblock In \emph{8th Annual Conference on Robot Learning}, 2024.

\bibitem[Mandlekar et~al.(2023)Mandlekar, Nasiriany, Wen, Akinola, Narang, Fan,
  Zhu, and Fox]{mandlekar2023mimicgendatagenerationscalable}
A.~Mandlekar, S.~Nasiriany, B.~Wen, I.~Akinola, Y.~Narang, L.~Fan, Y.~Zhu, and
  D.~Fox.
\newblock Mimicgen: A data generation system for scalable robot learning using
  human demonstrations, 2023.
\newblock URL \url{https://arxiv.org/abs/2310.17596}.

\bibitem[Zhou et~al.(2019)Zhou, Barnes, Lu, Yang, and Li]{Zhou2019rotation-rep}
Y.~Zhou, C.~Barnes, J.~Lu, J.~Yang, and H.~Li.
\newblock {On the continuity of rotation representations in neural networks}.
\newblock In \emph{{2019 IEEE/CVF Conference on Computer Vision and Pattern
  Recognition (CVPR)}}. IEEE, June 2019.

\bibitem[Zhao et~al.(2023)Zhao, Kumar, Levine, and Finn]{zhao2023learning}
T.~Z. Zhao, V.~Kumar, S.~Levine, and C.~Finn.
\newblock Learning fine-grained bimanual manipulation with low-cost hardware.
\newblock \emph{arXiv preprint arXiv:2304.13705}, 2023.

\bibitem[Carion et~al.(2020)Carion, Massa, Synnaeve, Usunier, Kirillov, and
  Zagoruyko]{carion2020endtoendobjectdetectiontransformers}
N.~Carion, F.~Massa, G.~Synnaeve, N.~Usunier, A.~Kirillov, and S.~Zagoruyko.
\newblock End-to-end object detection with transformers, 2020.
\newblock URL \url{https://arxiv.org/abs/2005.12872}.

\bibitem[Garipov et~al.(2023)Garipov, Peuter, Yang, Garg, Kaski, and
  Jaakkola]{garipov2023compositionalsculptingiterativegenerative}
T.~Garipov, S.~D. Peuter, G.~Yang, V.~Garg, S.~Kaski, and T.~Jaakkola.
\newblock Compositional sculpting of iterative generative processes, 2023.
\newblock URL \url{https://arxiv.org/abs/2309.16115}.

\bibitem[Chi et~al.(2024)Chi, Xu, Feng, Cousineau, Du, Burchfiel, Tedrake, and
  Song]{chi2024diffusionpolicy}
C.~Chi, Z.~Xu, S.~Feng, E.~Cousineau, Y.~Du, B.~Burchfiel, R.~Tedrake, and
  S.~Song.
\newblock Diffusion policy: Visuomotor policy learning via action diffusion.
\newblock \emph{The International Journal of Robotics Research}, 2024.

\bibitem[Janner et~al.(2022)Janner, Du, Tenenbaum, and
  Levine]{janner2022planningdiffusionflexiblebehavior}
M.~Janner, Y.~Du, J.~B. Tenenbaum, and S.~Levine.
\newblock Planning with diffusion for flexible behavior synthesis, 2022.
\newblock URL \url{https://arxiv.org/abs/2205.09991}.

\bibitem[Perez et~al.(2017)Perez, Strub, de~Vries, Dumoulin, and
  Courville]{perez2017filmvisualreasoninggeneral}
E.~Perez, F.~Strub, H.~de~Vries, V.~Dumoulin, and A.~Courville.
\newblock Film: Visual reasoning with a general conditioning layer, 2017.
\newblock URL \url{https://arxiv.org/abs/1709.07871}.

\bibitem[Chen et~al.(2024)Chen, Wang, Nguyen, Fei-Fei, and Liu]{chen2024arcap}
S.~Chen, C.~Wang, K.~Nguyen, L.~Fei-Fei, and C.~K. Liu.
\newblock Arcap: Collecting high-quality human demonstrations for robot
  learning with augmented reality feedback.
\newblock \emph{arXiv preprint arXiv:2410.08464}, 2024.

\bibitem[Nechyporenko et~al.(2024)Nechyporenko, Hoque, Webb, Sivapurapu, and
  Zhang]{nechyporenko2024armadaaugmentedrealityrobot}
N.~Nechyporenko, R.~Hoque, C.~Webb, M.~Sivapurapu, and J.~Zhang.
\newblock Armada: Augmented reality for robot manipulation and robot-free data
  acquisition, 2024.
\newblock URL \url{https://arxiv.org/abs/2412.10631}.

\bibitem[Jiang et~al.(2024)Jiang, Mattes, Jia, Schreiber, Neumann, and
  Lioutikov]{10661067}
X.~Jiang, P.~Mattes, X.~Jia, N.~Schreiber, G.~Neumann, and R.~Lioutikov.
\newblock A comprehensive user study on augmented reality-based data collection
  interfaces for robot learning.
\newblock In \emph{2024 19th ACM/IEEE International Conference on Human-Robot
  Interaction (HRI)}, pages 333--342, 2024.

\bibitem[Yang et~al.(2024)Yang, Ikeda, Bertasius, and
  Szafir]{yang2024arcadescalabledemonstrationcollection}
Y.~Yang, B.~Ikeda, G.~Bertasius, and D.~Szafir.
\newblock Arcade: Scalable demonstration collection and generation via
  augmented reality for imitation learning, 2024.
\newblock URL \url{https://arxiv.org/abs/2410.15994}.

\bibitem[Duan et~al.(2023)Duan, Wang, Shridhar, Fox, and Krishna]{duan2023ar2}
J.~Duan, Y.~R. Wang, M.~Shridhar, D.~Fox, and R.~Krishna.
\newblock Ar2-d2: Training a robot without a robot.
\newblock 2023.

\bibitem[Wang et~al.(2024)Wang, Chang, Duan, Fox, and
  Krishna]{wang2024eveenablingtrainrobots}
J.~Wang, C.-C. Chang, J.~Duan, D.~Fox, and R.~Krishna.
\newblock Eve: Enabling anyone to train robots using augmented reality, 2024.
\newblock URL \url{https://arxiv.org/abs/2404.06089}.

\bibitem[van Haastregt et~al.(2024)van Haastregt, Welle, Zhang, and
  Kragic]{vanhaastregt2024puppeteerrobotaugmentedreality}
J.~van Haastregt, M.~C. Welle, Y.~Zhang, and D.~Kragic.
\newblock Puppeteer your robot: Augmented reality leader-follower
  teleoperation, 2024.
\newblock URL \url{https://arxiv.org/abs/2407.11741}.

\bibitem[Iyer et~al.(2024)Iyer, Peng, Dai, Guzey, Haldar, Chintala, and
  Pinto]{iyer2024openteachversatileteleoperation}
A.~Iyer, Z.~Peng, Y.~Dai, I.~Guzey, S.~Haldar, S.~Chintala, and L.~Pinto.
\newblock Open teach: A versatile teleoperation system for robotic
  manipulation, 2024.
\newblock URL \url{https://arxiv.org/abs/2403.07870}.

\bibitem[Naceri et~al.(2021)Naceri, Mazzanti, Bimbo, Tefera, Prattichizzo,
  Caldwell, Mattos, and Deshpande]{10.1007/s10846-021-01311-7}
A.~Naceri, D.~Mazzanti, J.~Bimbo, Y.~T. Tefera, D.~Prattichizzo, D.~G.
  Caldwell, L.~S. Mattos, and N.~Deshpande.
\newblock The vicarios virtual reality interface for remote robotic
  teleoperation: Teleporting for intuitive tele-manipulation.
\newblock \emph{J. Intell. Robotics Syst.}, 101\penalty0 (4), Apr. 2021.
\newblock ISSN 0921-0296.
\newblock \doi{10.1007/s10846-021-01311-7}.
\newblock URL \url{https://doi.org/10.1007/s10846-021-01311-7}.

\bibitem[Jiang et~al.(2025)Jiang, Yuan, Dincer, Zhou, Li, Li, Haag, Schreiber,
  Li, Neumann, and Lioutikov]{jiang2025irisimmersiverobotinteraction}
X.~Jiang, Q.~Yuan, E.~U. Dincer, H.~Zhou, G.~Li, X.~Li, J.~Haag, N.~Schreiber,
  K.~Li, G.~Neumann, and R.~Lioutikov.
\newblock Iris: An immersive robot interaction system, 2025.
\newblock URL \url{https://arxiv.org/abs/2502.03297}.

\bibitem[Park et~al.(2024)Park, Bhatia, Ankile, and Agrawal]{DART}
Y.~Park, J.~S. Bhatia, L.~Ankile, and P.~Agrawal.
\newblock Dexhub and dart: Towards internet scale robot data collection, 2024.
\newblock URL \url{https://arxiv.org/abs/2411.02214}.

\bibitem[Katara et~al.(2023)Katara, Xian, and
  Fragkiadaki]{katara2023gen2simscalingrobotlearning}
P.~Katara, Z.~Xian, and K.~Fragkiadaki.
\newblock Gen2sim: Scaling up robot learning in simulation with generative
  models, 2023.
\newblock URL \url{https://arxiv.org/abs/2310.18308}.

\bibitem[Wang et~al.(2024)Wang, Xian, Chen, Wang, Wang, Fragkiadaki, Erickson,
  Held, and Gan]{wang2024robogenunleashinginfinitedata}
Y.~Wang, Z.~Xian, F.~Chen, T.-H. Wang, Y.~Wang, K.~Fragkiadaki, Z.~Erickson,
  D.~Held, and C.~Gan.
\newblock Robogen: Towards unleashing infinite data for automated robot
  learning via generative simulation, 2024.
\newblock URL \url{https://arxiv.org/abs/2311.01455}.

\bibitem[Ma et~al.(2024)Ma, Liang, Wang, Wang, Zhu, Fan, Bastani, and
  Jayaraman]{ma2024dreurekalanguagemodelguided}
Y.~J. Ma, W.~Liang, H.-J. Wang, S.~Wang, Y.~Zhu, L.~Fan, O.~Bastani, and
  D.~Jayaraman.
\newblock Dreureka: Language model guided sim-to-real transfer, 2024.
\newblock URL \url{https://arxiv.org/abs/2406.01967}.

\bibitem[Chen et~al.(2024)Chen, Walsman, Memmel, Mo, Fang, Vemuri, Wu, Fox, and
  Gupta]{chen2024urdformer}
Z.~Chen, A.~Walsman, M.~Memmel, K.~Mo, A.~Fang, K.~Vemuri, A.~Wu, D.~Fox, and
  A.~Gupta.
\newblock Urdformer: A pipeline for constructing articulated simulation
  environments from real-world images.
\newblock \emph{arXiv preprint arXiv:2405.11656}, 2024.

\bibitem[Yu et~al.(2023)Yu, Xiao, Stone, Tompson, Brohan, Wang, Singh, Tan, M,
  Peralta, Ichter, Hausman, and Xia]{yu2023scalingrobotlearningsemantically}
T.~Yu, T.~Xiao, A.~Stone, J.~Tompson, A.~Brohan, S.~Wang, J.~Singh, C.~Tan,
  D.~M, J.~Peralta, B.~Ichter, K.~Hausman, and F.~Xia.
\newblock Scaling robot learning with semantically imagined experience, 2023.
\newblock URL \url{https://arxiv.org/abs/2302.11550}.

\bibitem[Mandi et~al.(2022)Mandi, Bharadhwaj, Moens, Song, Rajeswaran, and
  Kumar]{mandi2022cacti}
Z.~Mandi, H.~Bharadhwaj, V.~Moens, S.~Song, A.~Rajeswaran, and V.~Kumar.
\newblock Cacti: A framework for scalable multi-task multi-scene visual
  imitation learning.
\newblock \emph{arXiv preprint arXiv:2212.05711}, 2022.

\bibitem[Jang et~al.(2025)Jang, Ye, Lin, Xiang, Bjorck, Fang, Hu, Huang,
  Kundalia, Lin, Magne, Mandlekar, Narayan, Tan, Wang, Wang, Wang, Xu, Zeng,
  Zheng, Zheng, Liu, Zettlemoyer, Fox, Kautz, Reed, Zhu, and
  Fan]{jang2025dreamgenunlockinggeneralizationrobot}
J.~Jang, S.~Ye, Z.~Lin, J.~Xiang, J.~Bjorck, Y.~Fang, F.~Hu, S.~Huang,
  K.~Kundalia, Y.-C. Lin, L.~Magne, A.~Mandlekar, A.~Narayan, Y.~L. Tan,
  G.~Wang, J.~Wang, Q.~Wang, Y.~Xu, X.~Zeng, K.~Zheng, R.~Zheng, M.-Y. Liu,
  L.~Zettlemoyer, D.~Fox, J.~Kautz, S.~Reed, Y.~Zhu, and L.~Fan.
\newblock Dreamgen: Unlocking generalization in robot learning through video
  world models, 2025.
\newblock URL \url{https://arxiv.org/abs/2505.12705}.

\bibitem[Chen et~al.(2023)Chen, Kiami, Gupta, and Kumar]{chen2023genaug}
Z.~Chen, S.~Kiami, A.~Gupta, and V.~Kumar.
\newblock Genaug: Retargeting behaviors to unseen situations via generative
  augmentation.
\newblock \emph{arXiv preprint arXiv:2302.06671}, 2023.

\bibitem[Collaboration et~al.(2023)Collaboration, O'Neill, Rehman, Gupta,
  Maddukuri, Gupta, Padalkar, Lee, Pooley, Gupta, Mandlekar, Jain, Tung,
  Bewley, Herzog, Irpan, Khazatsky, Rai, Gupta, Wang, Kolobov, Singh, Garg,
  Kembhavi, Xie, Brohan, Raffin, Sharma, Yavary, Jain, Balakrishna, Wahid,
  Burgess-Limerick, Kim, Schölkopf, Wulfe, Ichter, Lu, Xu, Le, Finn, Wang, Xu,
  Chi, Huang, Chan, Agia, Pan, Fu, Devin, Xu, Morton, Driess, Chen, Pathak,
  Shah, Büchler, Jayaraman, Kalashnikov, Sadigh, Johns, Foster, Liu, Ceola,
  Xia, Zhao, Frujeri, Stulp, Zhou, Sukhatme, Salhotra, Yan, Feng, Schiavi,
  Berseth, Kahn, Yang, Wang, Su, Fang, Shi, Bao, Amor, Christensen, Furuta,
  Bharadhwaj, Walke, Fang, Ha, Mordatch, Radosavovic, Leal, Liang, Abou-Chakra,
  Kim, Drake, Peters, Schneider, Hsu, Vakil, Bohg, Bingham, Wu, Gao, Hu, Wu,
  Wu, Sun, Luo, Gu, Tan, Oh, Wu, Lu, Yang, Malik, Silvério, Hejna, Booher,
  Tompson, Yang, Salvador, Lim, Han, Wang, Rao, Pertsch, Hausman, Go,
  Gopalakrishnan, Goldberg, Byrne, Oslund, Kawaharazuka, Black, Lin, Zhang,
  Ehsani, Lekkala, Ellis, Rana, Srinivasan, Fang, Singh, Zeng, Hatch, Hsu,
  Itti, Chen, Pinto, Fei-Fei, Tan, Fan, Ott, Lee, Weihs, Chen, Lepert, Memmel,
  Tomizuka, Itkina, Castro, Spero, Du, Ahn, Yip, Zhang, Ding, Heo, Srirama,
  Sharma, Kim, Irshad, Kanazawa, Hansen, Heess, Joshi, Suenderhauf, Liu, Palo,
  Shafiullah, Mees, Kroemer, Bastani, Sanketi, Miller, Yin, Wohlhart, Xu,
  Fagan, Mitrano, Sermanet, Abbeel, Sundaresan, Chen, Vuong, Rafailov, Tian,
  Doshi, Mart{'i}n-Mart{'i}n, Baijal, Scalise, Hendrix, Lin, Qian, Zhang,
  Mendonca, Shah, Hoque, Julian, Bustamante, Kirmani, Levine, Lin, Moore, Bahl,
  Dass, Sonawani, Tulsiani, Song, Xu, Haldar, Karamcheti, Adebola, Guist,
  Nasiriany, Schaal, Welker, Tian, Ramamoorthy, Dasari, Belkhale, Park, Nair,
  Mirchandani, Osa, Gupta, Harada, Matsushima, Xiao, Kollar, Yu, Ding, Davchev,
  Zhao, Armstrong, Darrell, Chung, Jain, Kumar, Vanhoucke, Guizilini, Zhan,
  Zhou, Burgard, Chen, Chen, Wang, Zhu, Geng, Liu, Liangwei, Li, Pang, Lu, Ma,
  Kim, Chebotar, Zhou, Zhu, Wu, Xu, Wang, Bisk, Dou, Cho, Lee, Cui, Cao, Wu,
  Tang, Zhu, Zhang, Jiang, Li, Li, Iwasawa, Matsuo, Ma, Xu, Cui, Zhang, Fu, and
  Lin]{open_x_embodiment_rt_x_2023}
O.~X.-E. Collaboration, A.~O'Neill, A.~Rehman, A.~Gupta, A.~Maddukuri,
  A.~Gupta, A.~Padalkar, A.~Lee, A.~Pooley, A.~Gupta, A.~Mandlekar, A.~Jain,
  A.~Tung, A.~Bewley, A.~Herzog, A.~Irpan, A.~Khazatsky, A.~Rai, A.~Gupta,
  A.~Wang, A.~Kolobov, A.~Singh, A.~Garg, A.~Kembhavi, A.~Xie, A.~Brohan,
  A.~Raffin, A.~Sharma, A.~Yavary, A.~Jain, A.~Balakrishna, A.~Wahid,
  B.~Burgess-Limerick, B.~Kim, B.~Schölkopf, B.~Wulfe, B.~Ichter, C.~Lu,
  C.~Xu, C.~Le, C.~Finn, C.~Wang, C.~Xu, C.~Chi, C.~Huang, C.~Chan, C.~Agia,
  C.~Pan, C.~Fu, C.~Devin, D.~Xu, D.~Morton, D.~Driess, D.~Chen, D.~Pathak,
  D.~Shah, D.~Büchler, D.~Jayaraman, D.~Kalashnikov, D.~Sadigh, E.~Johns,
  E.~Foster, F.~Liu, F.~Ceola, F.~Xia, F.~Zhao, F.~V. Frujeri, F.~Stulp,
  G.~Zhou, G.~S. Sukhatme, G.~Salhotra, G.~Yan, G.~Feng, G.~Schiavi,
  G.~Berseth, G.~Kahn, G.~Yang, G.~Wang, H.~Su, H.-S. Fang, H.~Shi, H.~Bao,
  H.~B. Amor, H.~I. Christensen, H.~Furuta, H.~Bharadhwaj, H.~Walke, H.~Fang,
  H.~Ha, I.~Mordatch, I.~Radosavovic, I.~Leal, J.~Liang, J.~Abou-Chakra,
  J.~Kim, J.~Drake, J.~Peters, J.~Schneider, J.~Hsu, J.~Vakil, J.~Bohg,
  J.~Bingham, J.~Wu, J.~Gao, J.~Hu, J.~Wu, J.~Wu, J.~Sun, J.~Luo, J.~Gu,
  J.~Tan, J.~Oh, J.~Wu, J.~Lu, J.~Yang, J.~Malik, J.~Silvério, J.~Hejna,
  J.~Booher, J.~Tompson, J.~Yang, J.~Salvador, J.~J. Lim, J.~Han, K.~Wang,
  K.~Rao, K.~Pertsch, K.~Hausman, K.~Go, K.~Gopalakrishnan, K.~Goldberg,
  K.~Byrne, K.~Oslund, K.~Kawaharazuka, K.~Black, K.~Lin, K.~Zhang, K.~Ehsani,
  K.~Lekkala, K.~Ellis, K.~Rana, K.~Srinivasan, K.~Fang, K.~P. Singh, K.-H.
  Zeng, K.~Hatch, K.~Hsu, L.~Itti, L.~Y. Chen, L.~Pinto, L.~Fei-Fei, L.~Tan,
  L.~J. Fan, L.~Ott, L.~Lee, L.~Weihs, M.~Chen, M.~Lepert, M.~Memmel,
  M.~Tomizuka, M.~Itkina, M.~G. Castro, M.~Spero, M.~Du, M.~Ahn, M.~C. Yip,
  M.~Zhang, M.~Ding, M.~Heo, M.~K. Srirama, M.~Sharma, M.~J. Kim, M.~Z. Irshad,
  N.~Kanazawa, N.~Hansen, N.~Heess, N.~J. Joshi, N.~Suenderhauf, N.~Liu, N.~D.
  Palo, N.~M.~M. Shafiullah, O.~Mees, O.~Kroemer, O.~Bastani, P.~R. Sanketi,
  P.~T. Miller, P.~Yin, P.~Wohlhart, P.~Xu, P.~D. Fagan, P.~Mitrano,
  P.~Sermanet, P.~Abbeel, P.~Sundaresan, Q.~Chen, Q.~Vuong, R.~Rafailov,
  R.~Tian, R.~Doshi, R.~Mart{'i}n-Mart{'i}n, R.~Baijal, R.~Scalise, R.~Hendrix,
  R.~Lin, R.~Qian, R.~Zhang, R.~Mendonca, R.~Shah, R.~Hoque, R.~Julian,
  S.~Bustamante, S.~Kirmani, S.~Levine, S.~Lin, S.~Moore, S.~Bahl, S.~Dass,
  S.~Sonawani, S.~Tulsiani, S.~Song, S.~Xu, S.~Haldar, S.~Karamcheti,
  S.~Adebola, S.~Guist, S.~Nasiriany, S.~Schaal, S.~Welker, S.~Tian,
  S.~Ramamoorthy, S.~Dasari, S.~Belkhale, S.~Park, S.~Nair, S.~Mirchandani,
  T.~Osa, T.~Gupta, T.~Harada, T.~Matsushima, T.~Xiao, T.~Kollar, T.~Yu,
  T.~Ding, T.~Davchev, T.~Z. Zhao, T.~Armstrong, T.~Darrell, T.~Chung, V.~Jain,
  V.~Kumar, V.~Vanhoucke, V.~Guizilini, W.~Zhan, W.~Zhou, W.~Burgard, X.~Chen,
  X.~Chen, X.~Wang, X.~Zhu, X.~Geng, X.~Liu, X.~Liangwei, X.~Li, Y.~Pang,
  Y.~Lu, Y.~J. Ma, Y.~Kim, Y.~Chebotar, Y.~Zhou, Y.~Zhu, Y.~Wu, Y.~Xu, Y.~Wang,
  Y.~Bisk, Y.~Dou, Y.~Cho, Y.~Lee, Y.~Cui, Y.~Cao, Y.-H. Wu, Y.~Tang, Y.~Zhu,
  Y.~Zhang, Y.~Jiang, Y.~Li, Y.~Li, Y.~Iwasawa, Y.~Matsuo, Z.~Ma, Z.~Xu, Z.~J.
  Cui, Z.~Zhang, Z.~Fu, and Z.~Lin.
\newblock Open {X-E}mbodiment: Robotic learning datasets and {RT-X} models.
\newblock \url{https://arxiv.org/abs/2310.08864}, 2023.

\bibitem[Fang et~al.(2023)Fang, Fang, Tang, Liu, Wang, Zhu, and
  Lu]{fang2023rh20t}
H.-S. Fang, H.~Fang, Z.~Tang, J.~Liu, J.~Wang, H.~Zhu, and C.~Lu.
\newblock Rh20t: A robotic dataset for learning diverse skills in one-shot.
\newblock In \emph{RSS 2023 Workshop on Learning for Task and Motion Planning},
  2023.

\bibitem[Khazatsky et~al.(2024)Khazatsky, Pertsch, Nair, Balakrishna, Dasari,
  Karamcheti, Nasiriany, Srirama, Chen, Ellis, Fagan, Hejna, Itkina, Lepert,
  Ma, Miller, Wu, Belkhale, Dass, Ha, Jain, Lee, Lee, Memmel, Park,
  Radosavovic, Wang, Zhan, Black, Chi, Hatch, Lin, Lu, Mercat, Rehman, Sanketi,
  Sharma, Simpson, Vuong, Walke, Wulfe, Xiao, Yang, Yavary, Zhao, Agia, Baijal,
  Castro, Chen, Chen, Chung, Drake, Foster, Gao, Herrera, Heo, Hsu, Hu,
  Jackson, Le, Li, Lin, Lin, Ma, Maddukuri, Mirchandani, Morton, Nguyen,
  O'Neill, Scalise, Seale, Son, Tian, Tran, Wang, Wu, Xie, Yang, Yin, Zhang,
  Bastani, Berseth, Bohg, Goldberg, Gupta, Gupta, Jayaraman, Lim, Malik,
  Martín-Martín, Ramamoorthy, Sadigh, Song, Wu, Yip, Zhu, Kollar, Levine, and
  Finn]{khazatsky2024droid}
A.~Khazatsky, K.~Pertsch, S.~Nair, A.~Balakrishna, S.~Dasari, S.~Karamcheti,
  S.~Nasiriany, M.~K. Srirama, L.~Y. Chen, K.~Ellis, P.~D. Fagan, J.~Hejna,
  M.~Itkina, M.~Lepert, Y.~J. Ma, P.~T. Miller, J.~Wu, S.~Belkhale, S.~Dass,
  H.~Ha, A.~Jain, A.~Lee, Y.~Lee, M.~Memmel, S.~Park, I.~Radosavovic, K.~Wang,
  A.~Zhan, K.~Black, C.~Chi, K.~B. Hatch, S.~Lin, J.~Lu, J.~Mercat, A.~Rehman,
  P.~R. Sanketi, A.~Sharma, C.~Simpson, Q.~Vuong, H.~R. Walke, B.~Wulfe,
  T.~Xiao, J.~H. Yang, A.~Yavary, T.~Z. Zhao, C.~Agia, R.~Baijal, M.~G. Castro,
  D.~Chen, Q.~Chen, T.~Chung, J.~Drake, E.~P. Foster, J.~Gao, D.~A. Herrera,
  M.~Heo, K.~Hsu, J.~Hu, D.~Jackson, C.~Le, Y.~Li, K.~Lin, R.~Lin, Z.~Ma,
  A.~Maddukuri, S.~Mirchandani, D.~Morton, T.~Nguyen, A.~O'Neill, R.~Scalise,
  D.~Seale, V.~Son, S.~Tian, E.~Tran, A.~E. Wang, Y.~Wu, A.~Xie, J.~Yang,
  P.~Yin, Y.~Zhang, O.~Bastani, G.~Berseth, J.~Bohg, K.~Goldberg, A.~Gupta,
  A.~Gupta, D.~Jayaraman, J.~J. Lim, J.~Malik, R.~Martín-Martín,
  S.~Ramamoorthy, D.~Sadigh, S.~Song, J.~Wu, M.~C. Yip, Y.~Zhu, T.~Kollar,
  S.~Levine, and C.~Finn.
\newblock Droid: A large-scale in-the-wild robot manipulation dataset.
\newblock 2024.

\end{thebibliography}

\clearpage
\section*{Appendix}


We include additional simulated environments; details on how we engineer the physics involved; in addition to details on the generative workflow in this appendix.
We also include examples of the virtual demonstrations that we collected, and examples of the synthetic images used for training. A live version of the presentation is included in
The accompanying video.


\subsection{Examples of Dexterous Virtual Demonstrations}

\begin{figure}[h]
\includegraphics[width=0.333\linewidth]{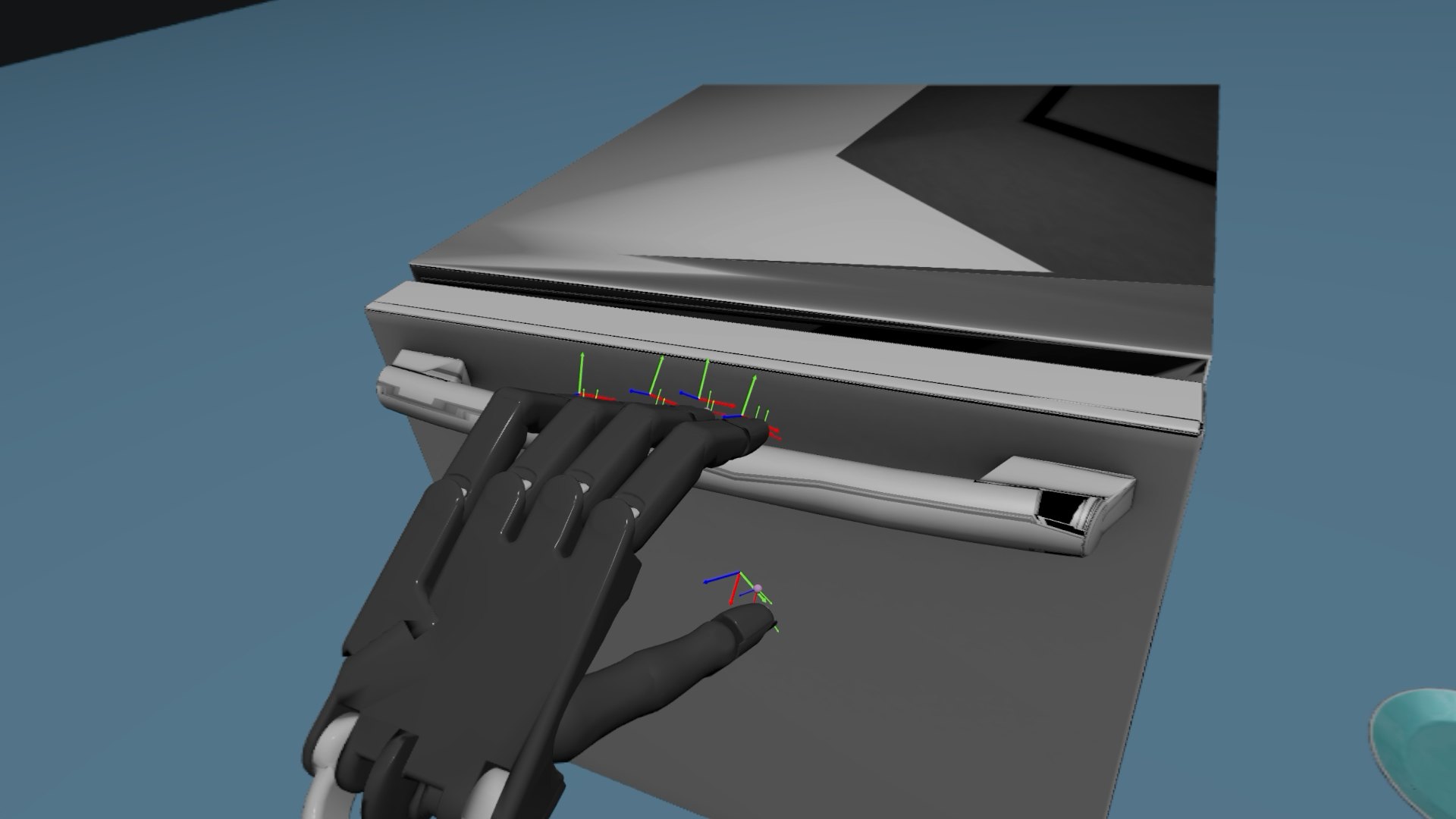}%
\includegraphics[width=0.333\linewidth]{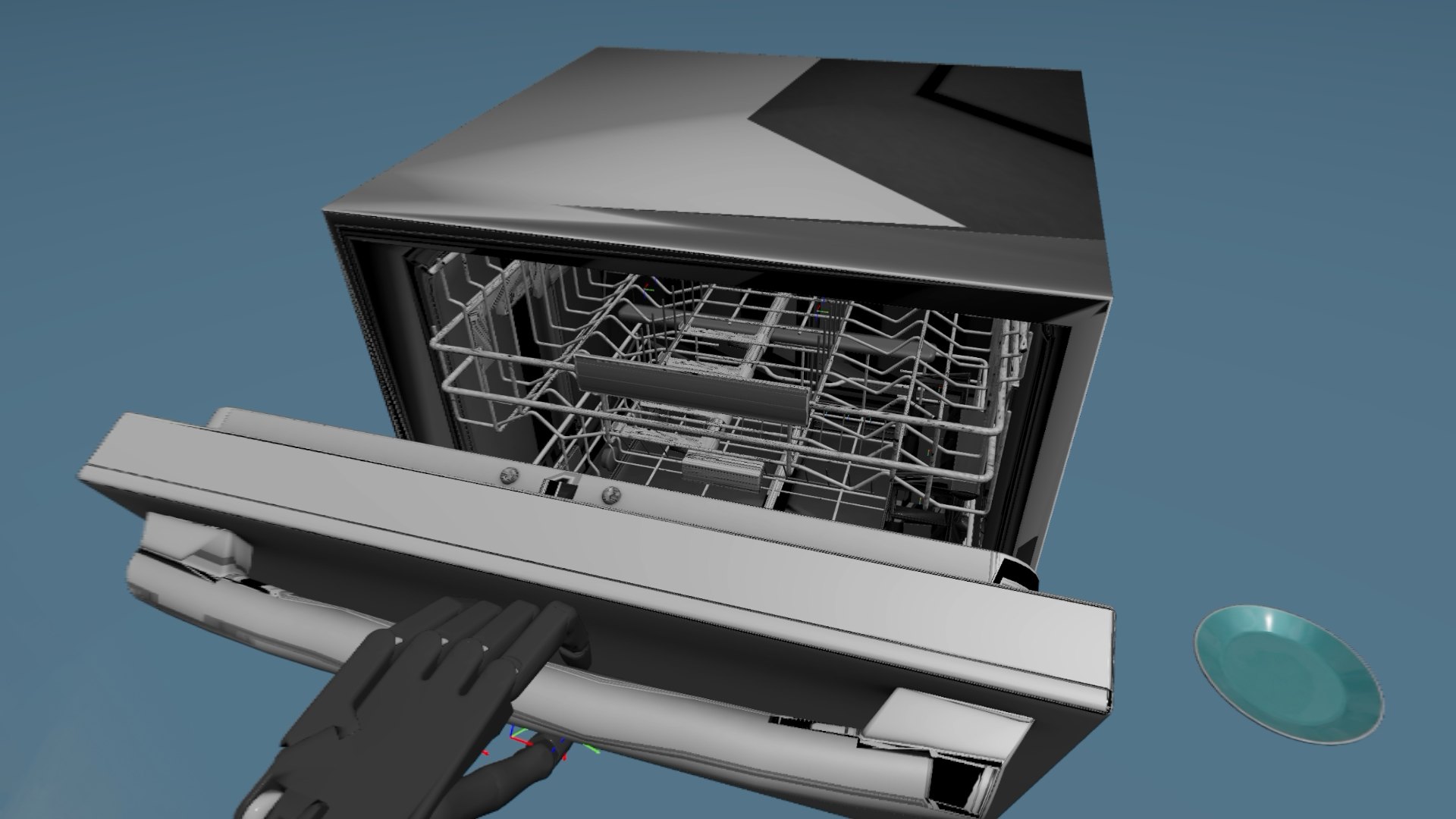}%
\includegraphics[width=0.333\linewidth]{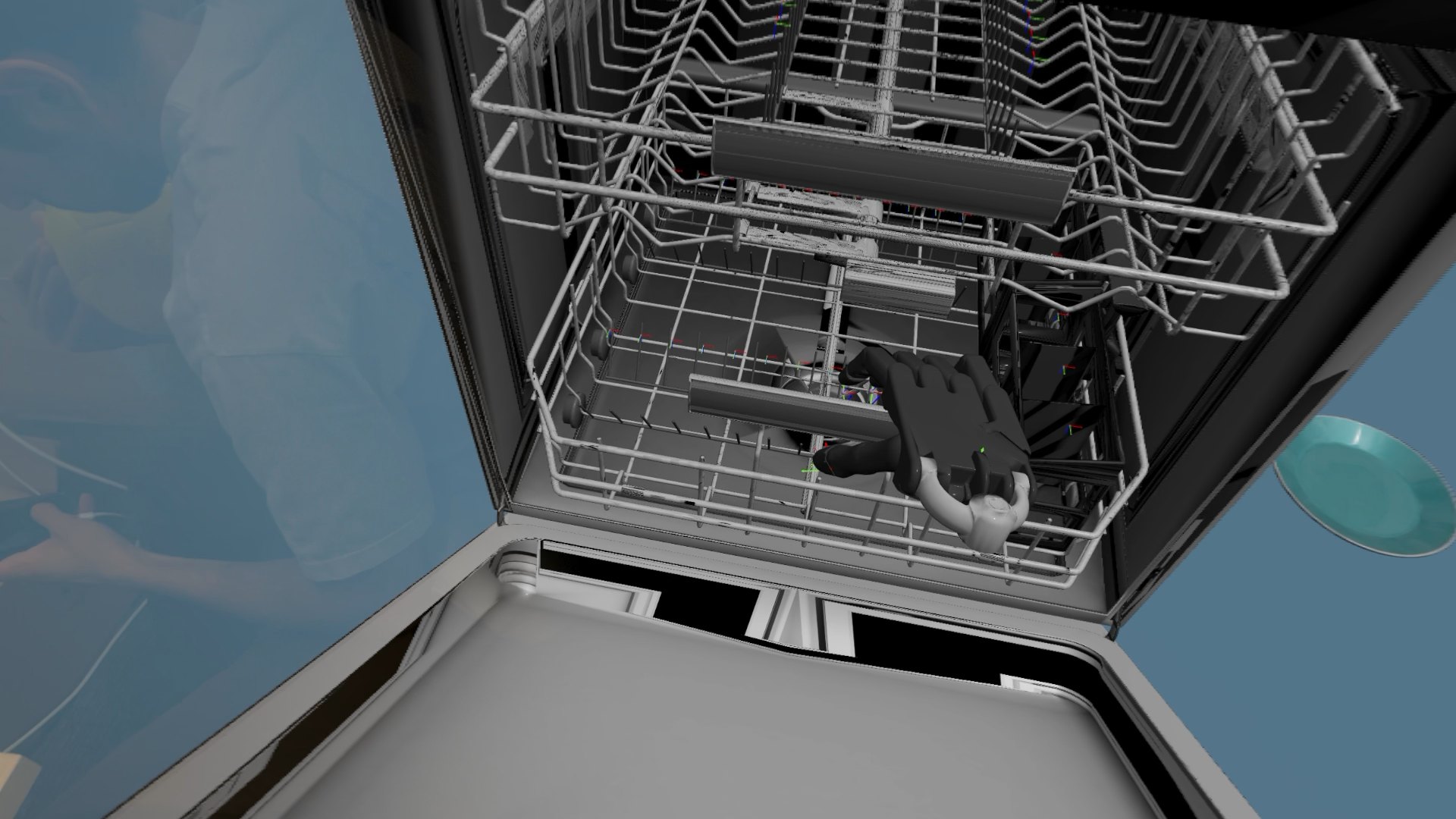}\newline%
\includegraphics[width=0.333\linewidth]{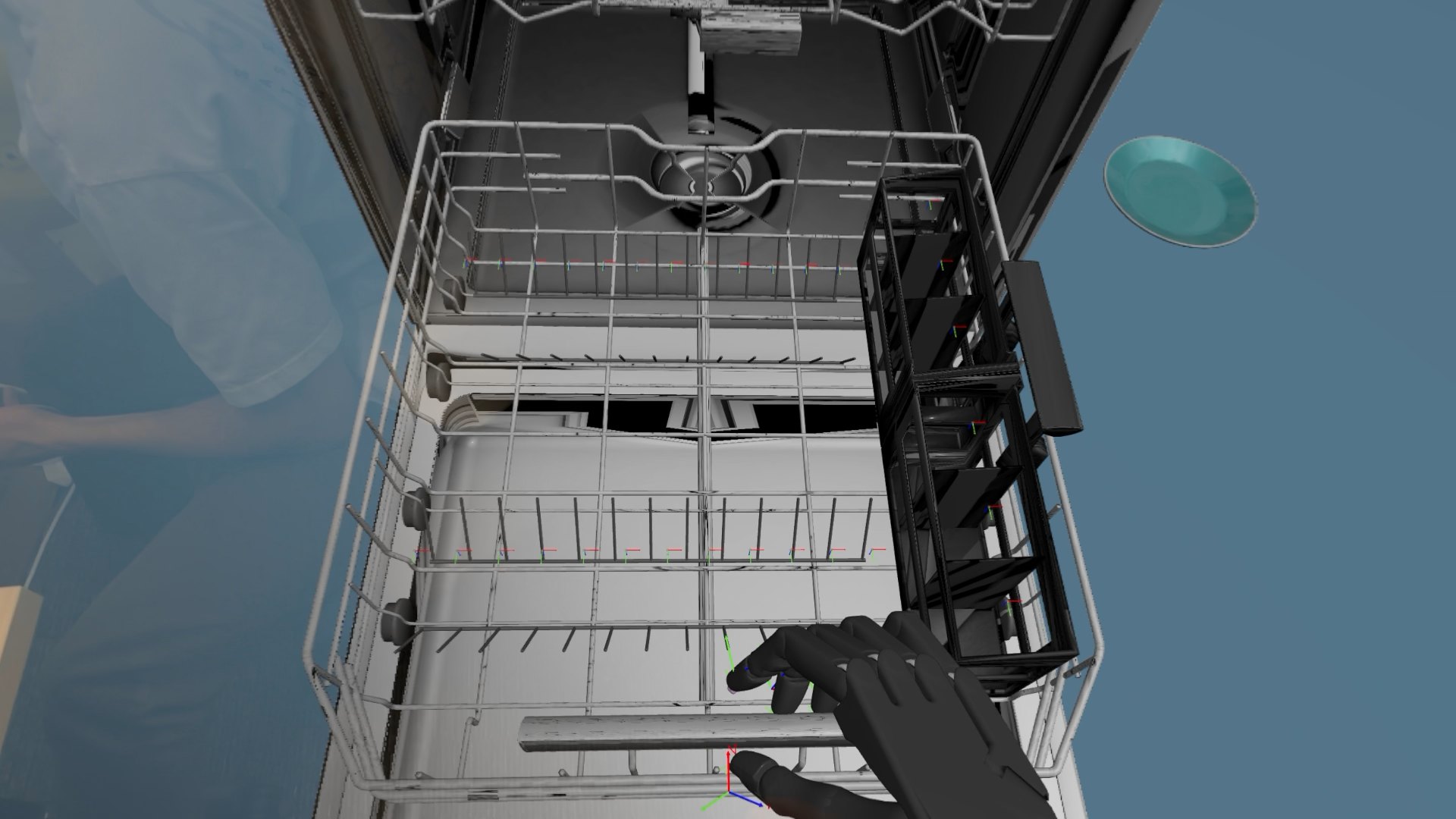}%
\includegraphics[width=0.333\linewidth]{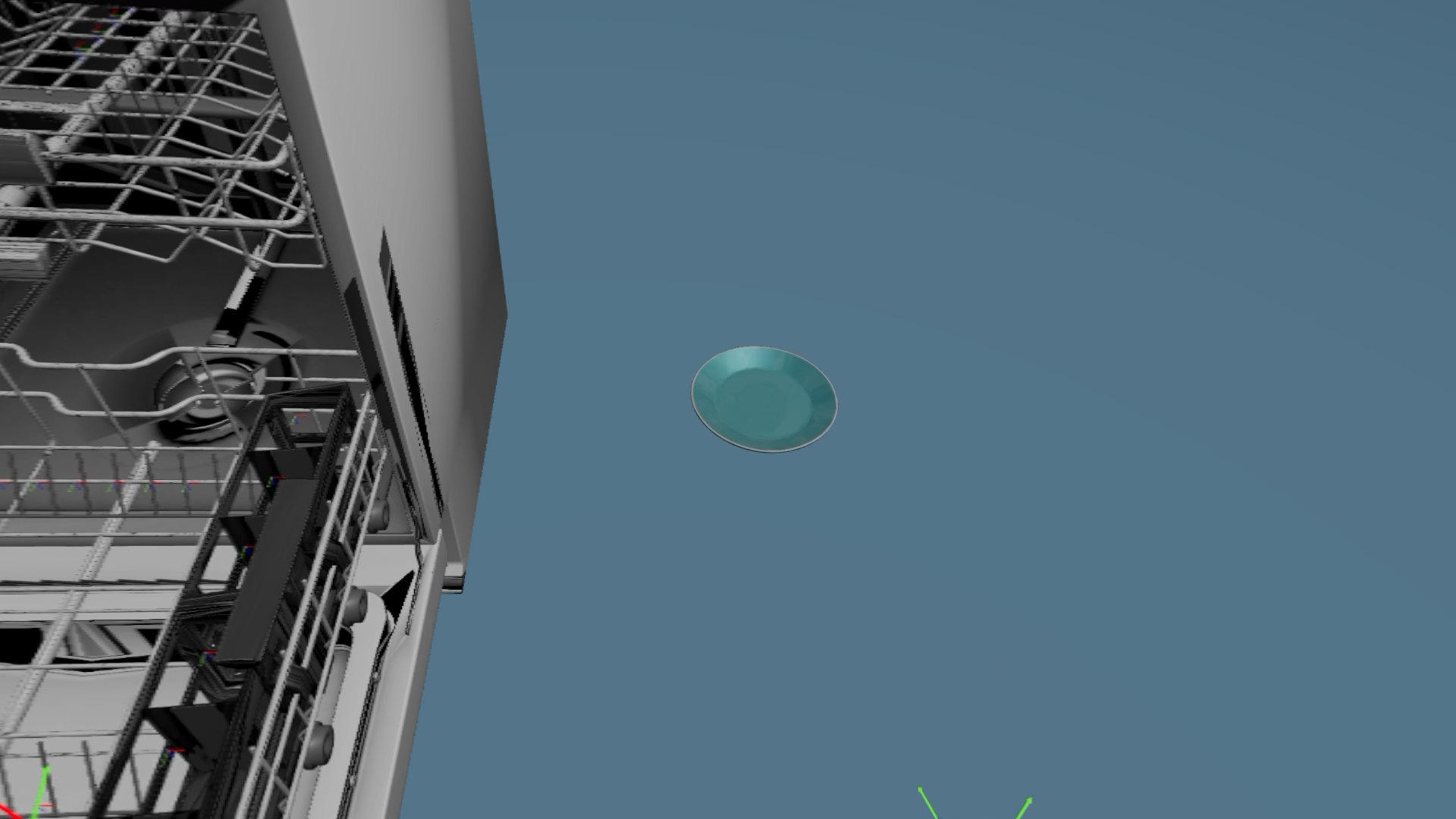}%
\includegraphics[width=0.333\linewidth]{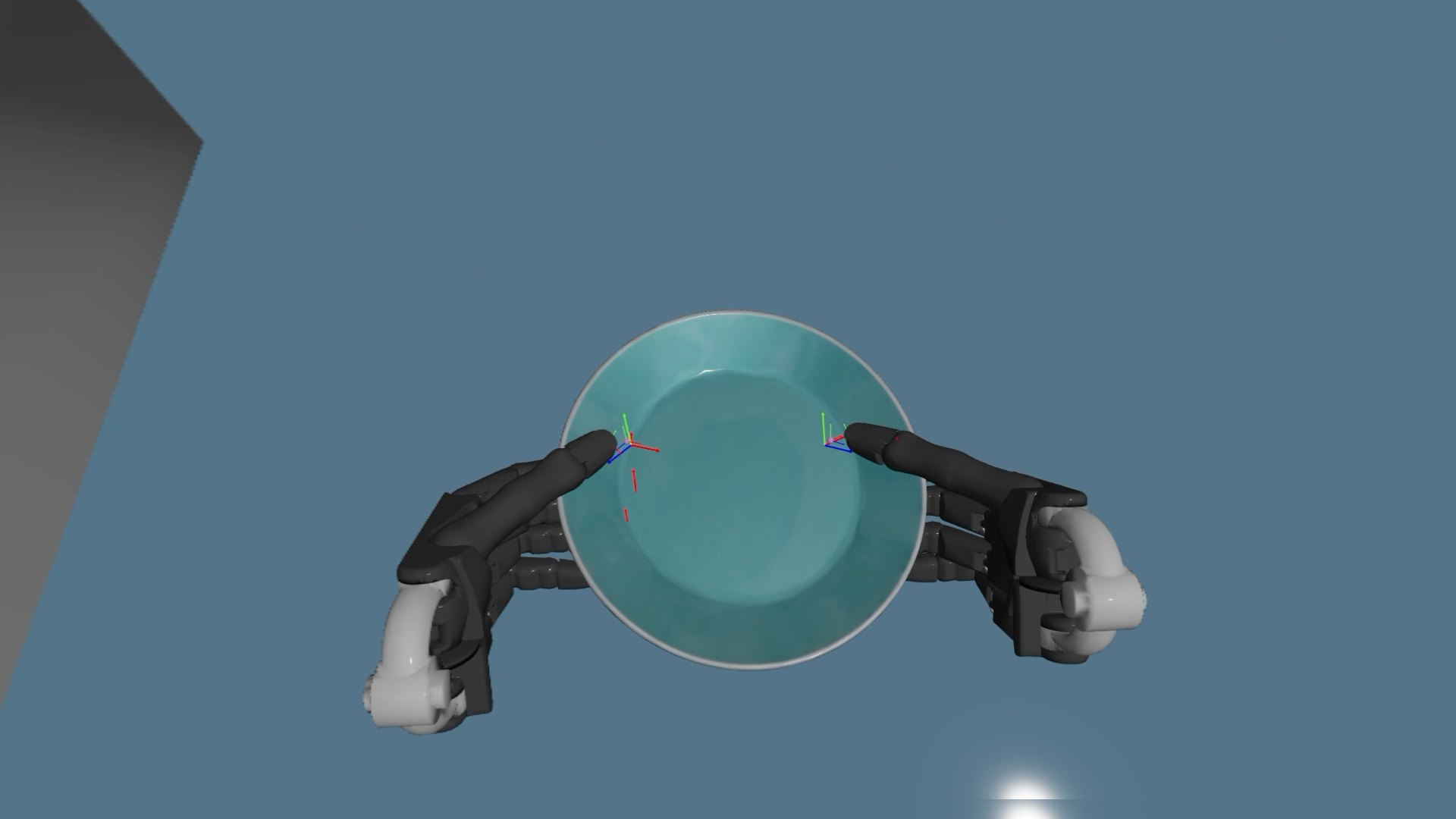}\newline%
\includegraphics[width=0.333\linewidth]{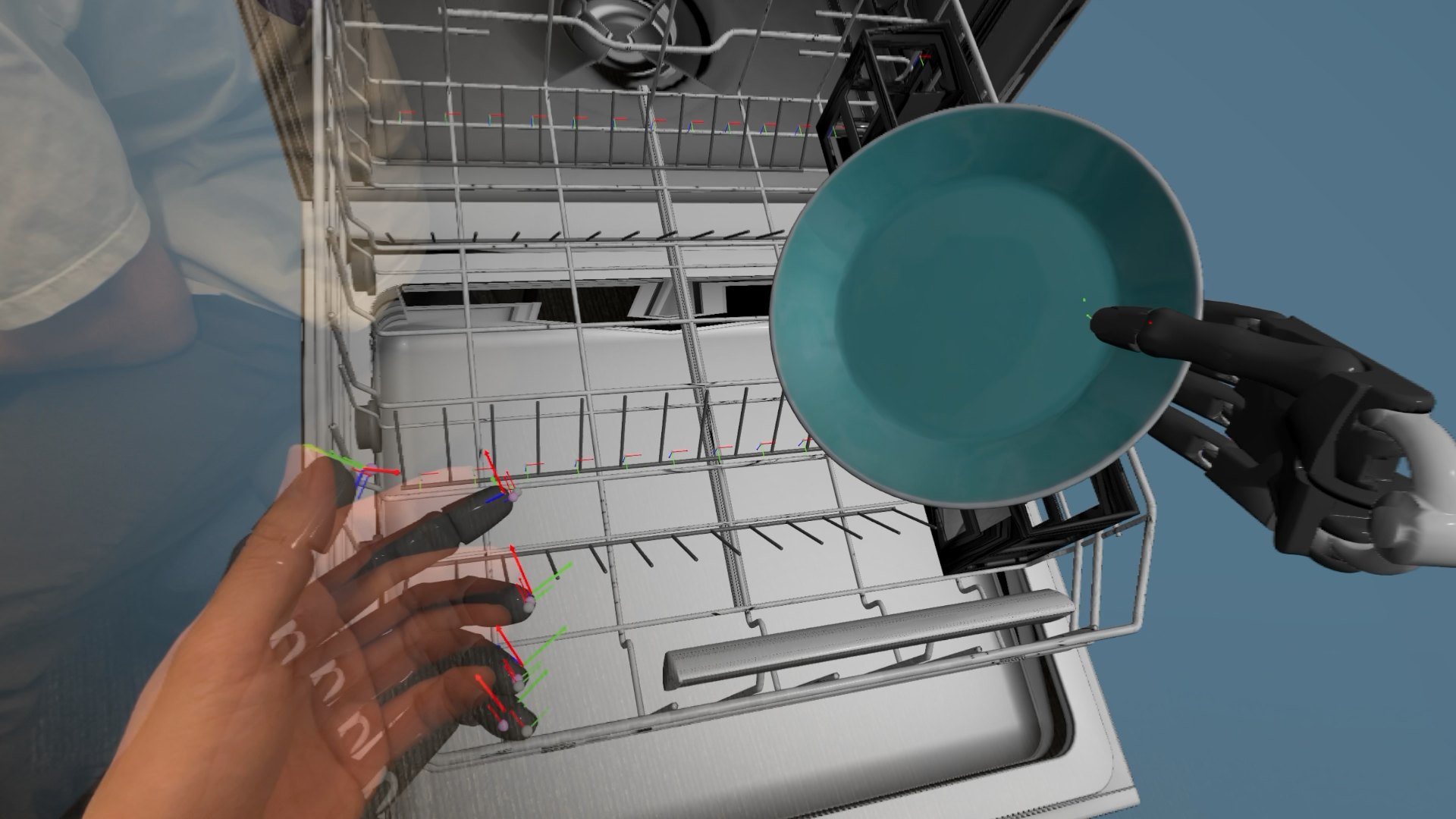}%
\includegraphics[width=0.333\linewidth]{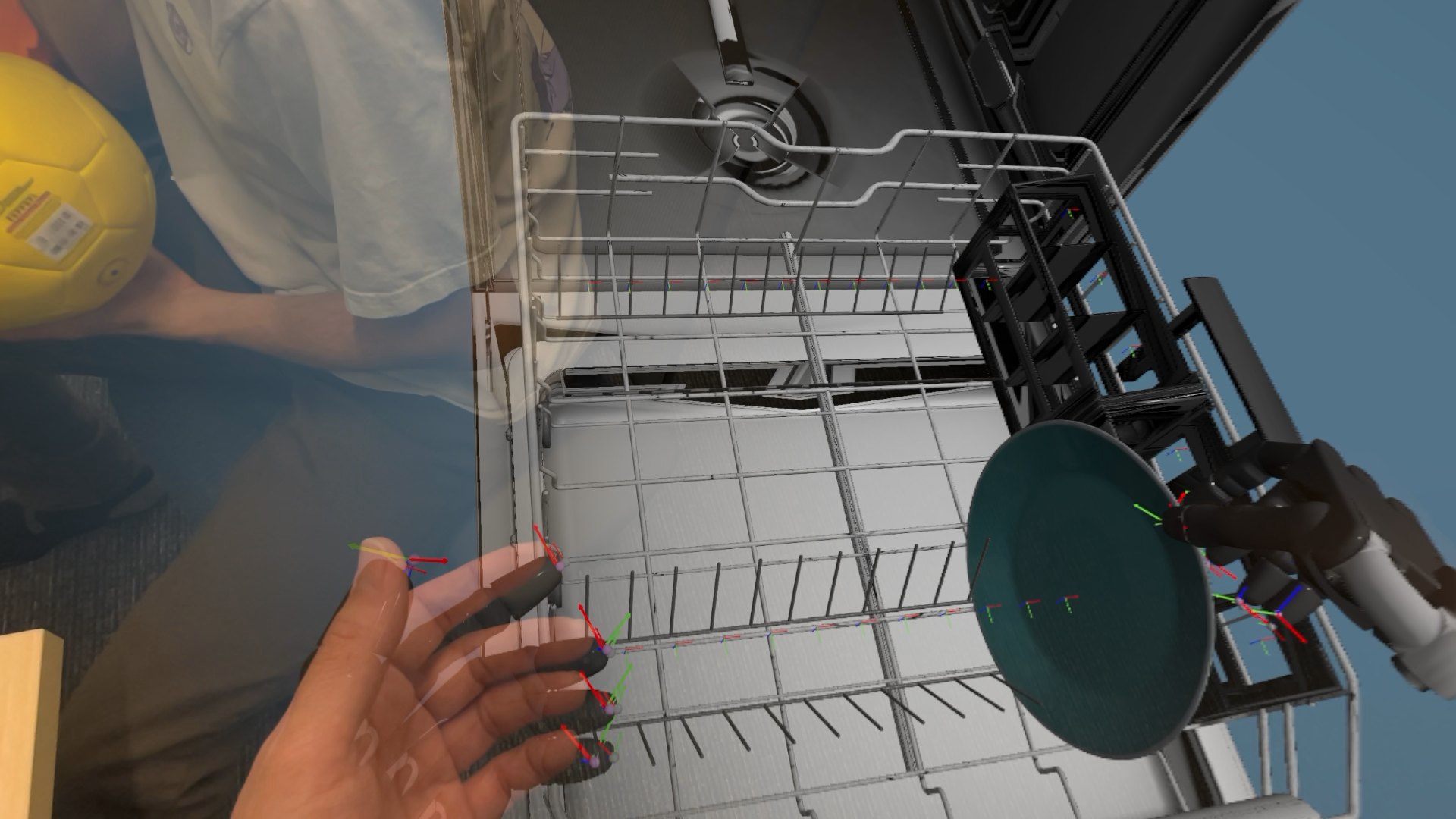}%
\includegraphics[width=0.333\linewidth]{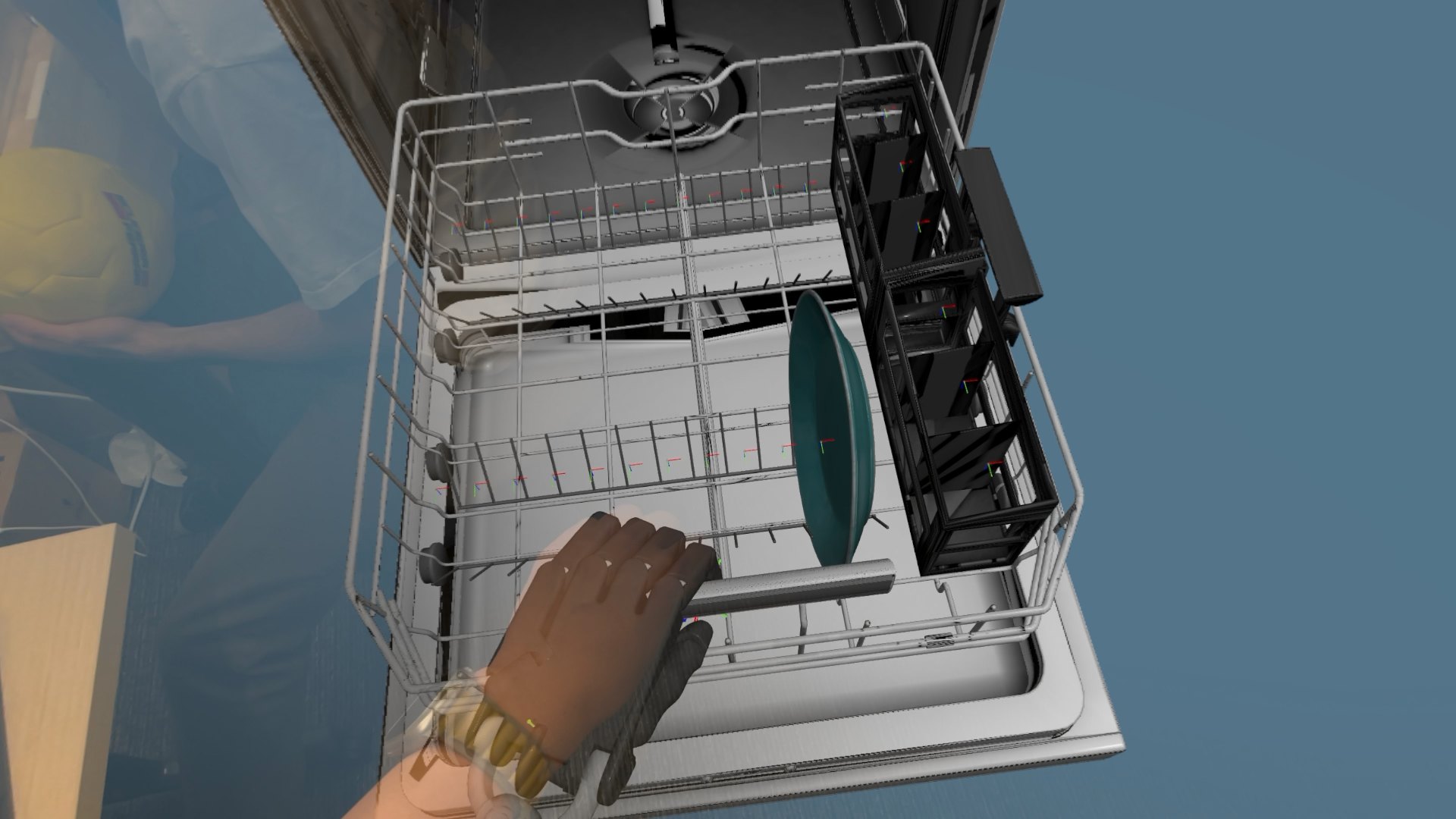}\newline%
\includegraphics[width=0.333\linewidth]{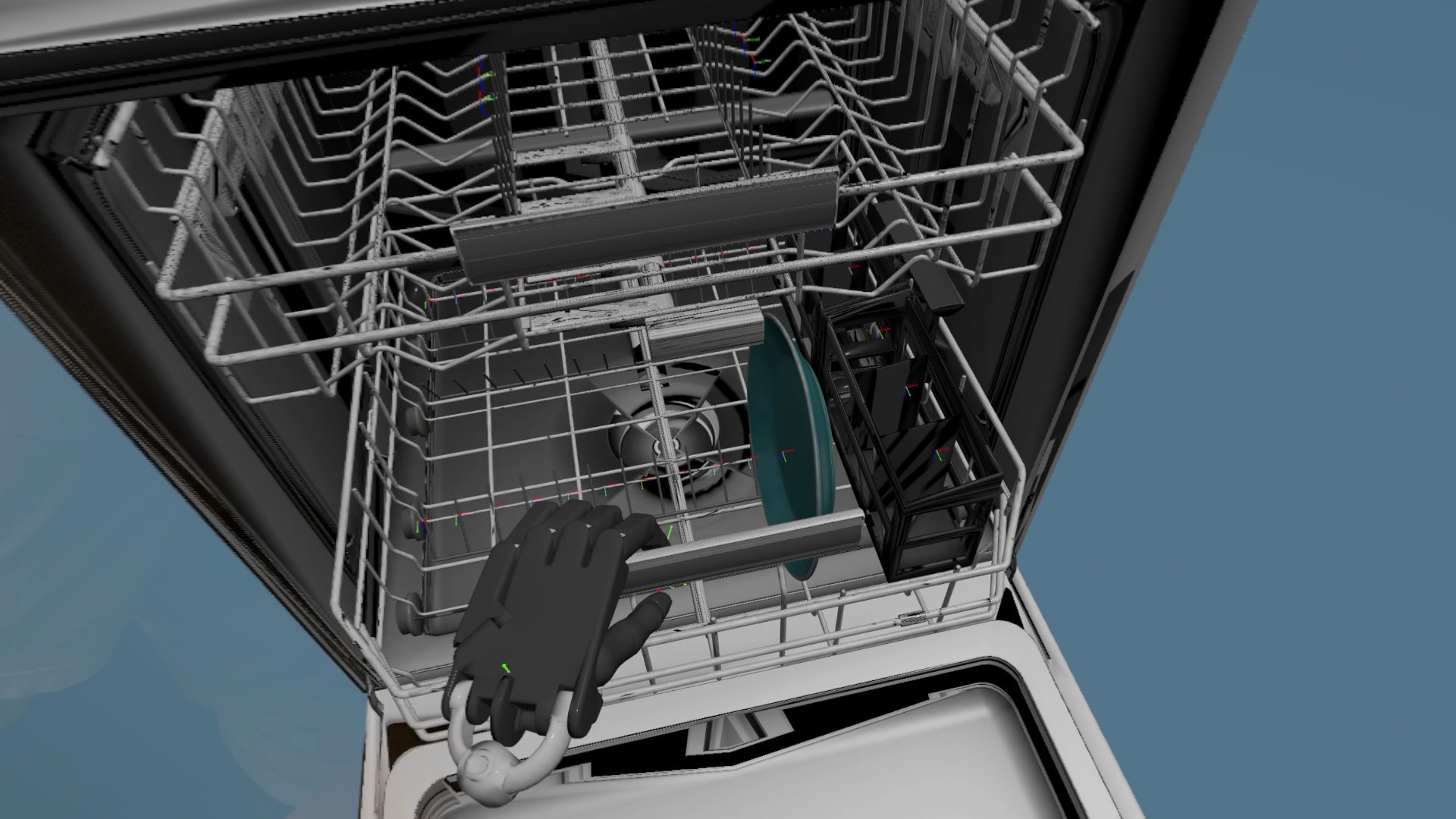}%
\includegraphics[width=0.333\linewidth]{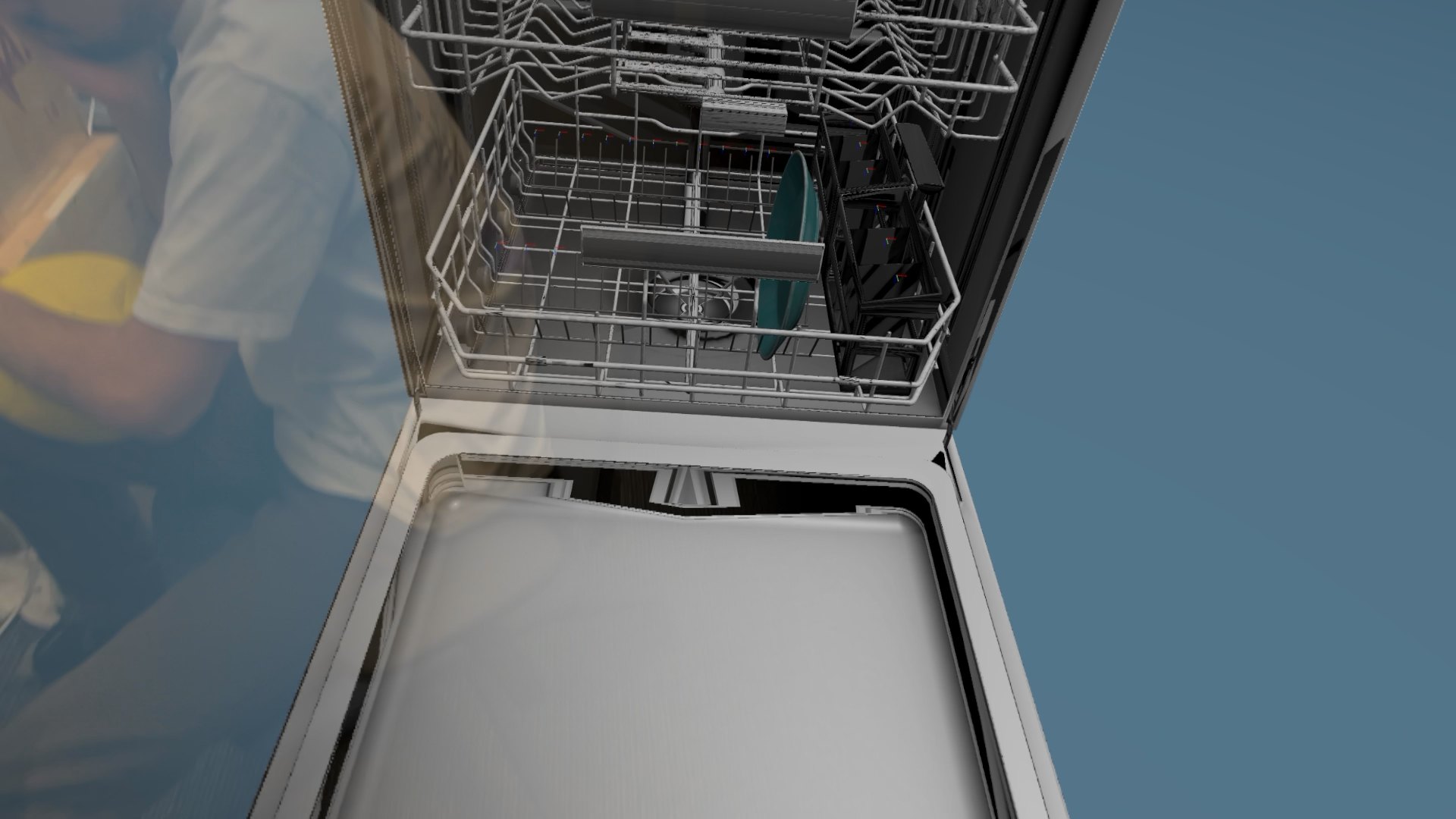}%
\includegraphics[width=0.333\linewidth]{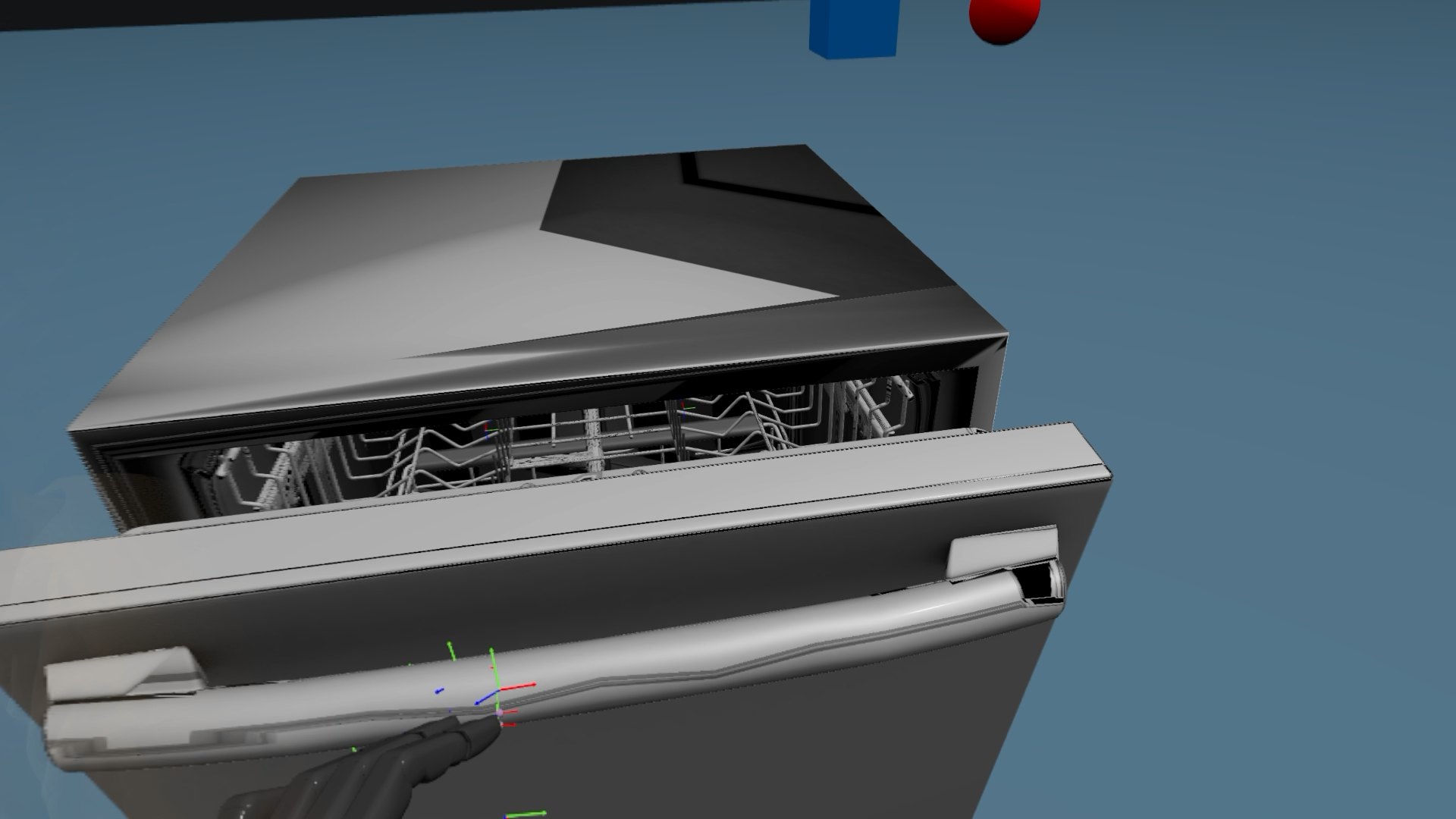}%
\caption{\textbf{Loading a Dishwasher.} Showing dexterous manipulation of a dishwasher with articulated doors, a plate, and its interaction with the racks inside the dishwasher.}
\label{fig:loading-dishwasher}
\end{figure}

\subsection{Examples of Policy Unroll in Simulation}

\begin{figure}[h]
\includegraphics[width=0.333\linewidth]{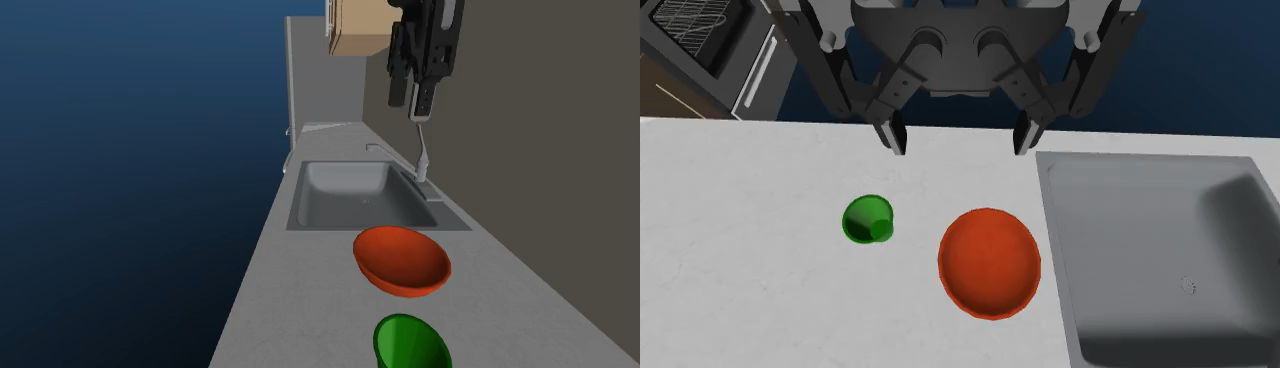}%
\includegraphics[width=0.333\linewidth]{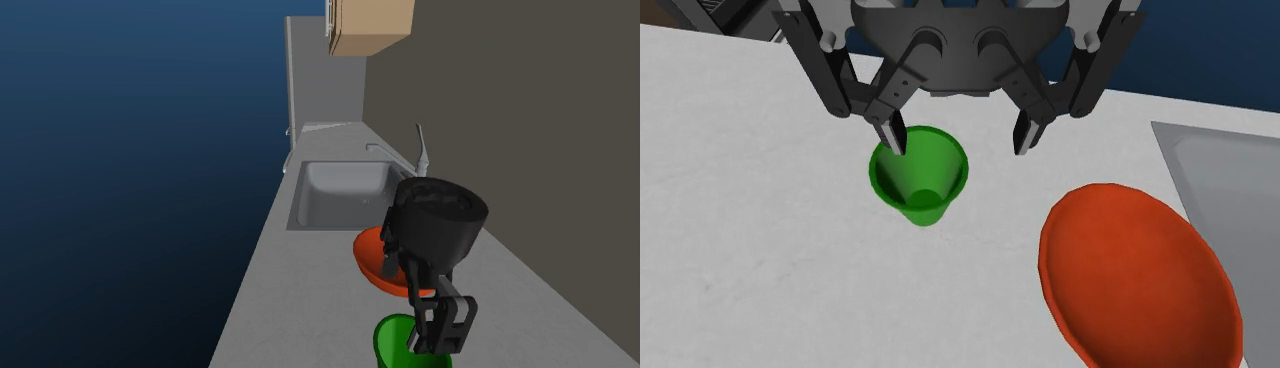}%
\includegraphics[width=0.333\linewidth]{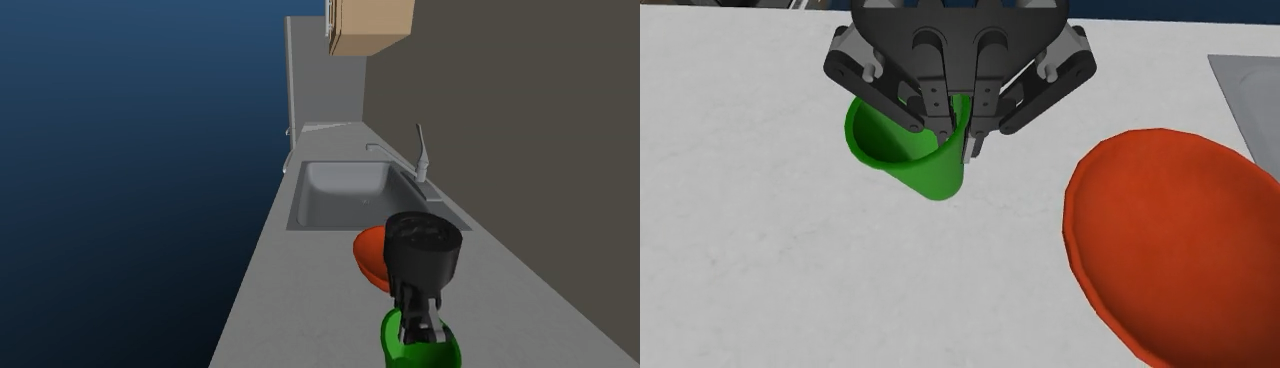}\newline%
\includegraphics[width=0.333\linewidth]{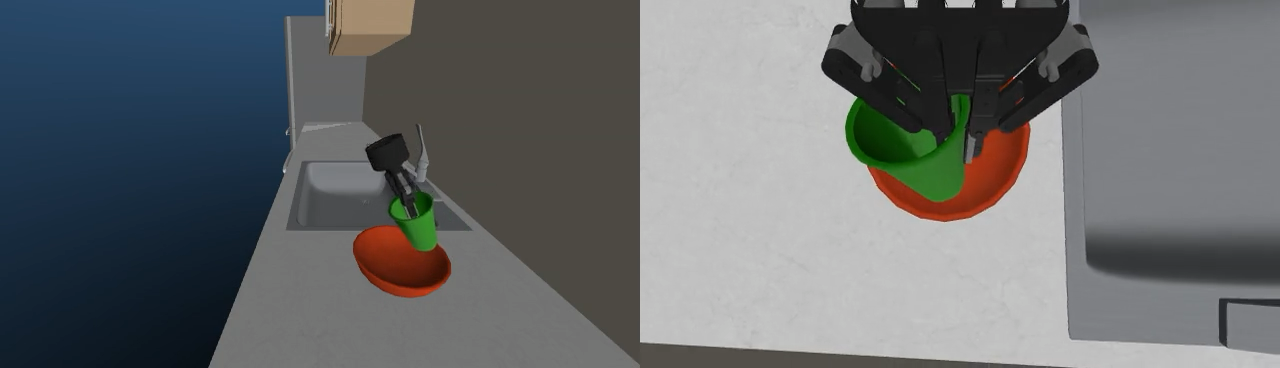}%
\includegraphics[width=0.333\linewidth]{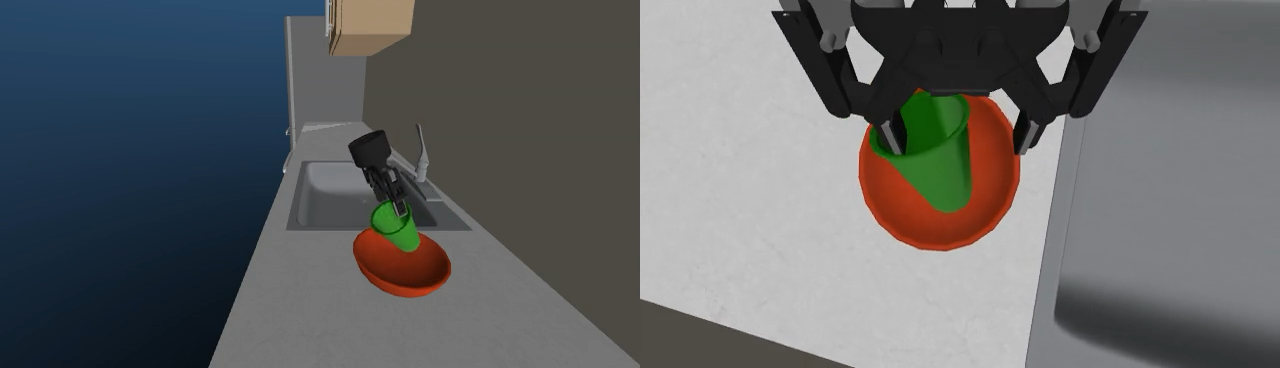}%
\includegraphics[width=0.333\linewidth]{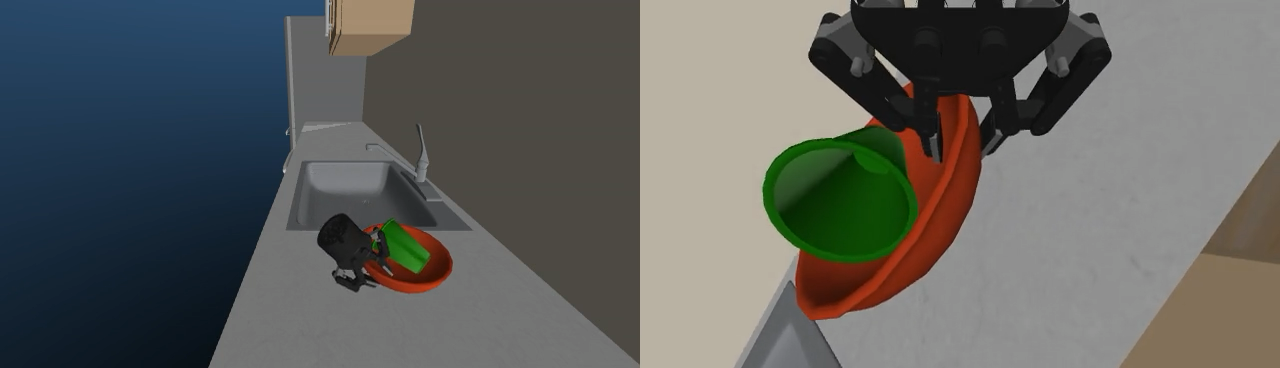}\newline%
\includegraphics[width=0.333\linewidth]{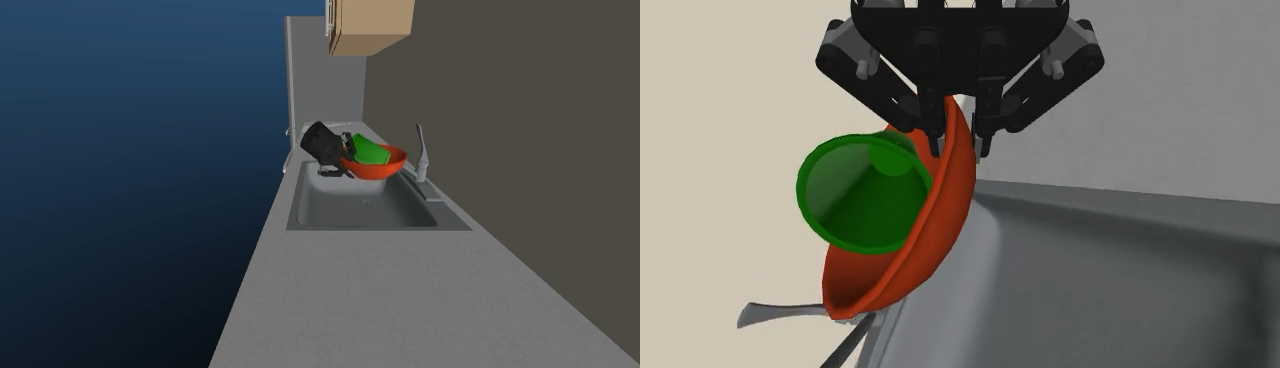}%
\includegraphics[width=0.333\linewidth]{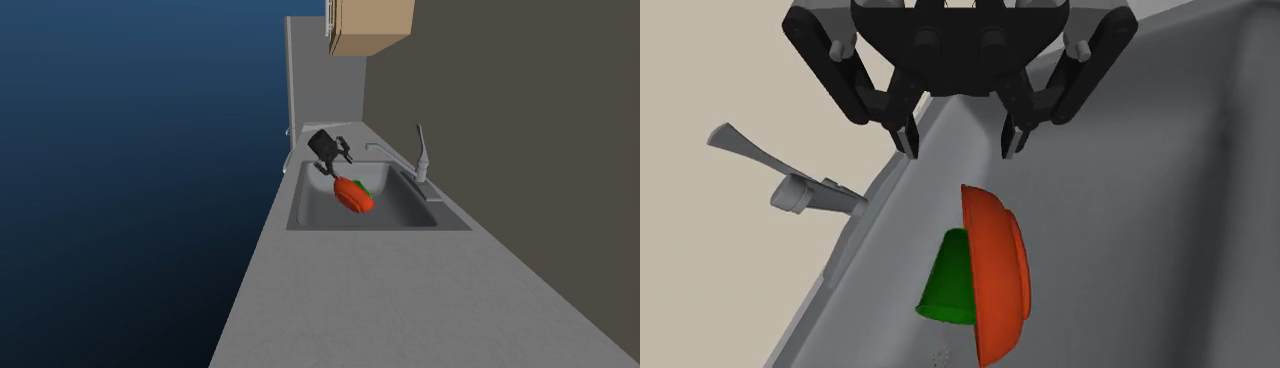}%
\includegraphics[width=0.333\linewidth]{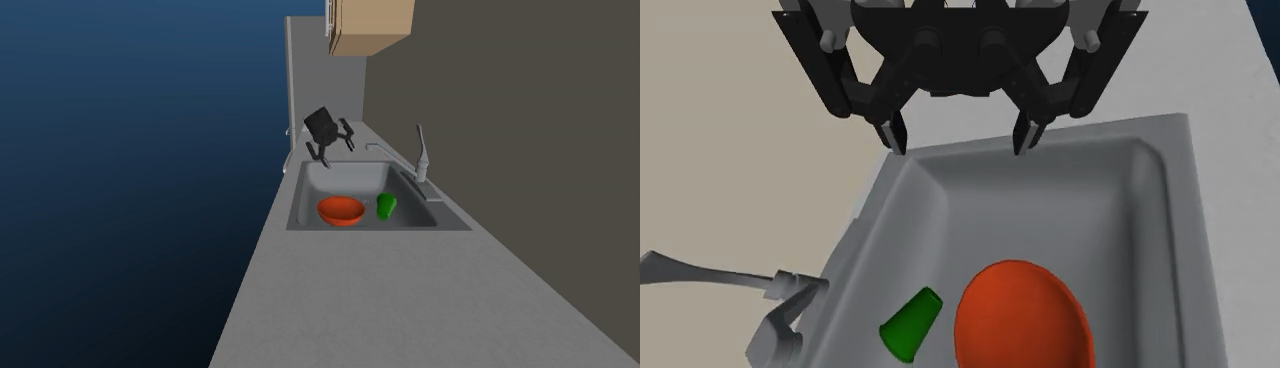}%
\caption{
\textbf{Cleaning The Kitchen.}  
The robot learns to pick up a cup and stack it on top of the bowl, followed by placing both objects into the kitchen sink.}
\label{fig:cleaning-kitchen}
\end{figure}

\subsection{Examples of Learned Re-try Behavior}

\begin{figure}[h]
\centering
\includegraphics[width=0.8\linewidth]{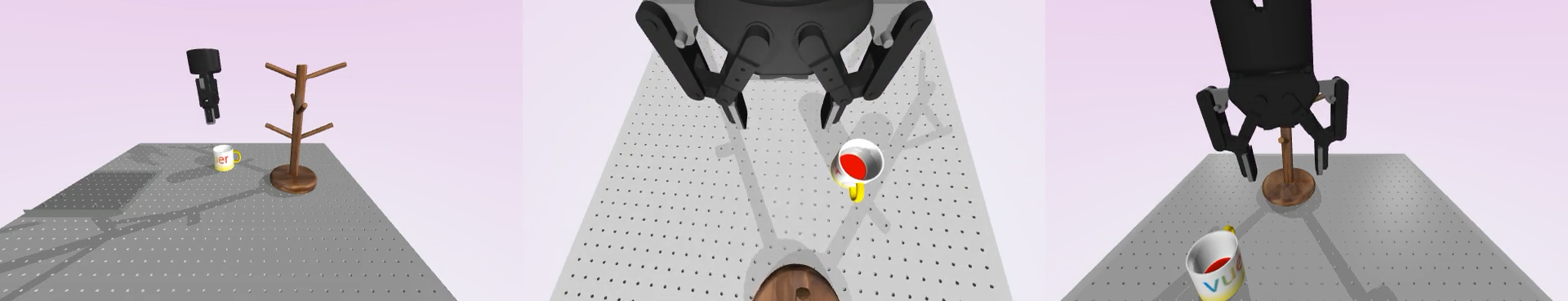}\newline%
\includegraphics[width=0.8\linewidth]{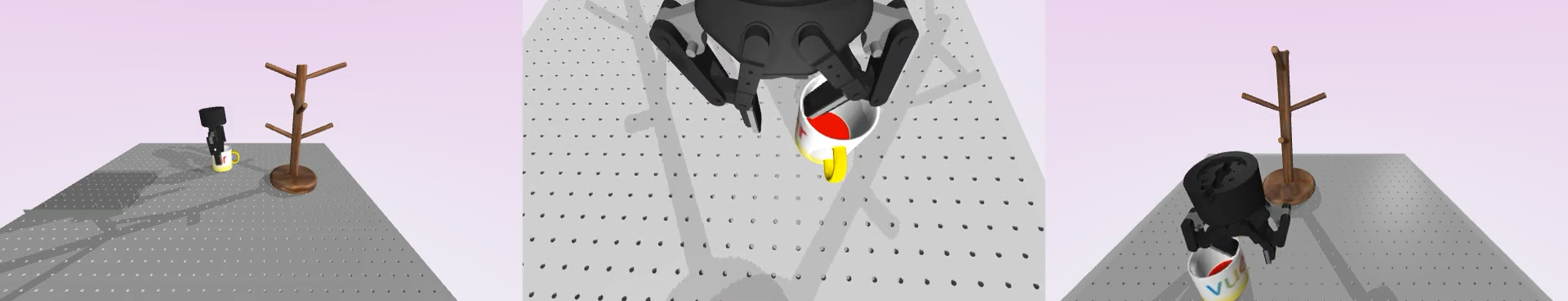}\newline%
\includegraphics[width=0.8\linewidth]{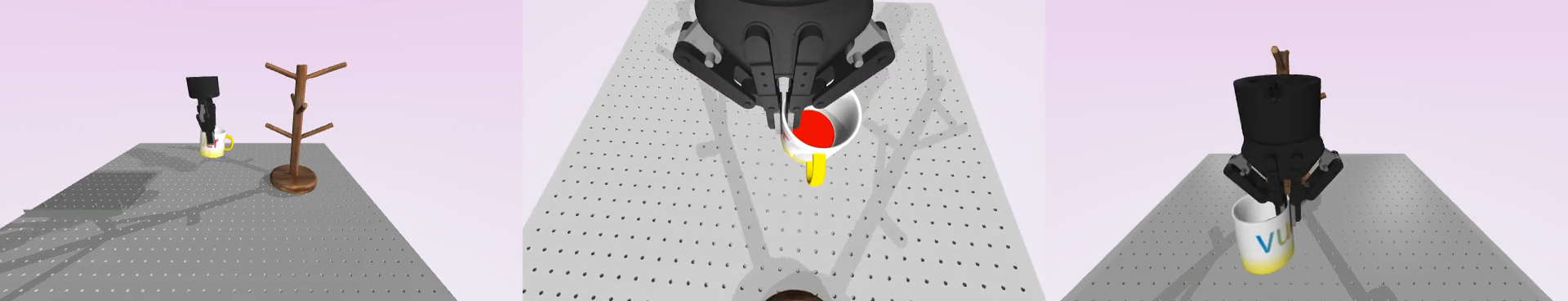}\newline%
\includegraphics[width=0.8\linewidth]{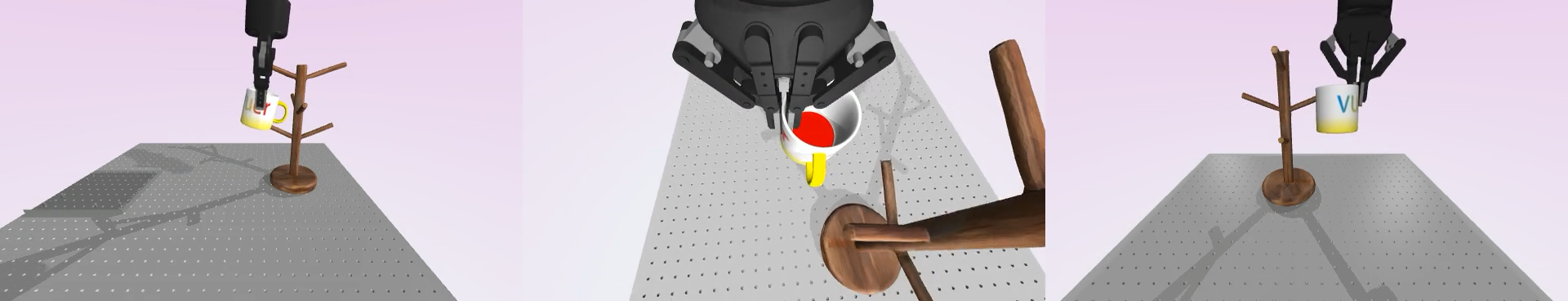}\newline%
\includegraphics[width=0.8\linewidth]{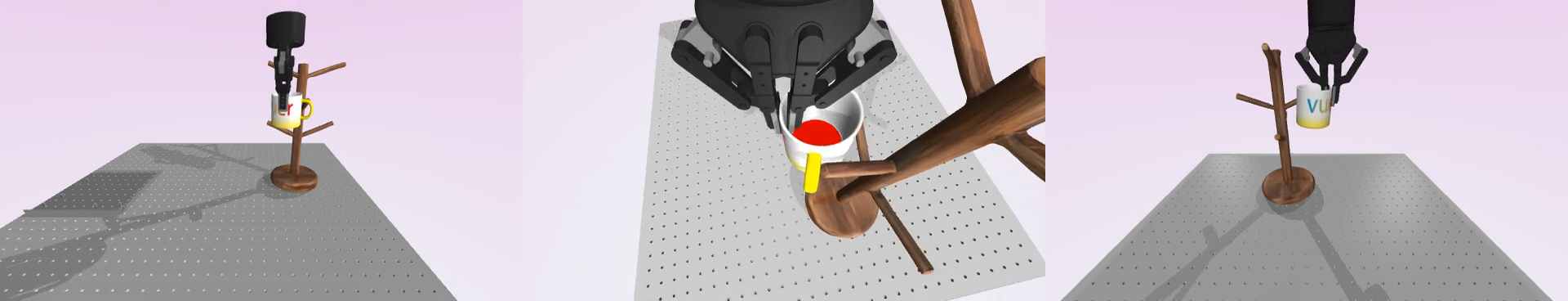}\newline%
\includegraphics[width=0.8\linewidth]{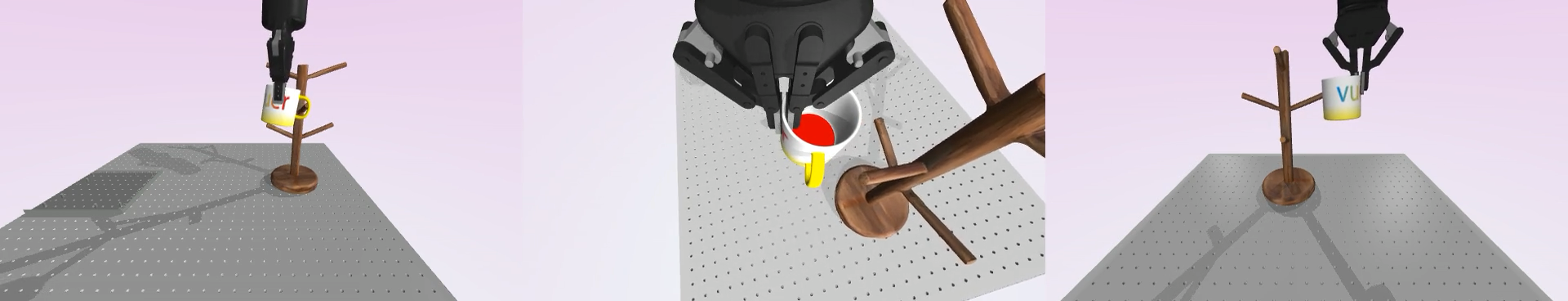}\newline%
\includegraphics[width=0.8\linewidth]{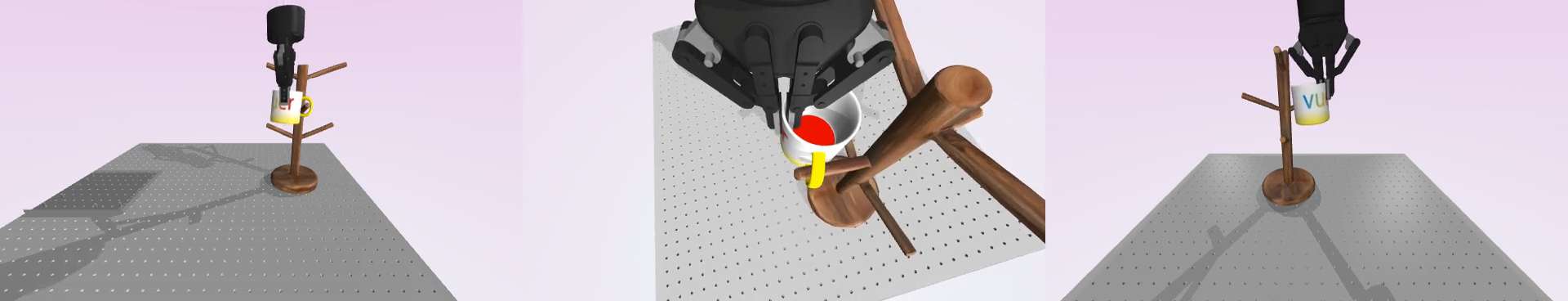}\newline%
\includegraphics[width=0.8\linewidth]{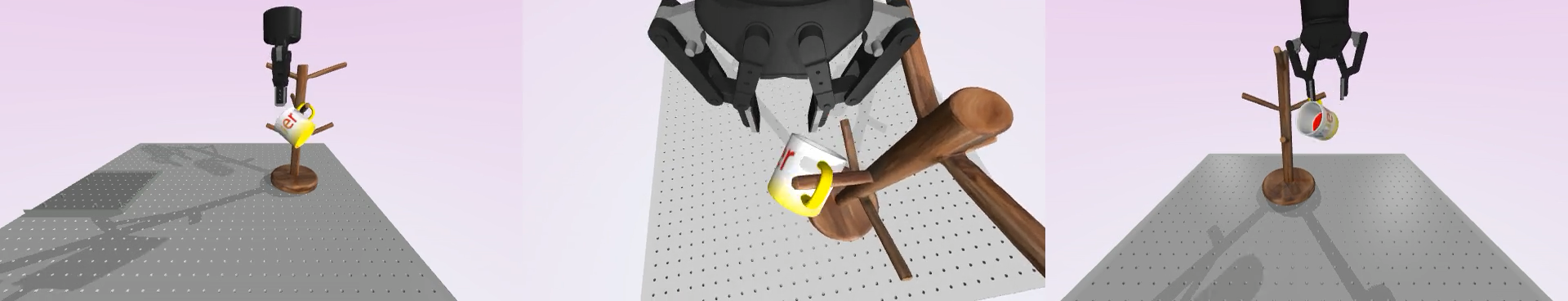}\newline%
\caption{\textbf{Re-try Behavior in the Mug Tree Environment.} 
The policy came into contact twice: First during the picking up of the mug; the second time is when trying to hang the mug onto the arm of the mug tree drying rack.
}
\label{fig:retry-behavior}
\end{figure}

We noticed robust, re-try behavior from the learned policy. We present the image sequence in Figure.~\ref{fig:retry-behavior}.%

\subsection{Additional Examples of Synthetic Data}

\begin{figure}[h]%
\includegraphics[width=0.25\linewidth]{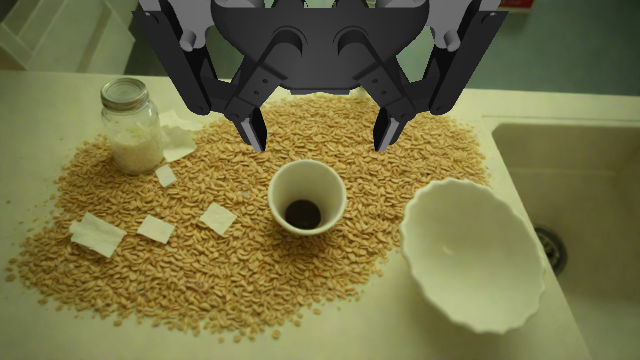}\hfill%
\includegraphics[width=0.25\linewidth]{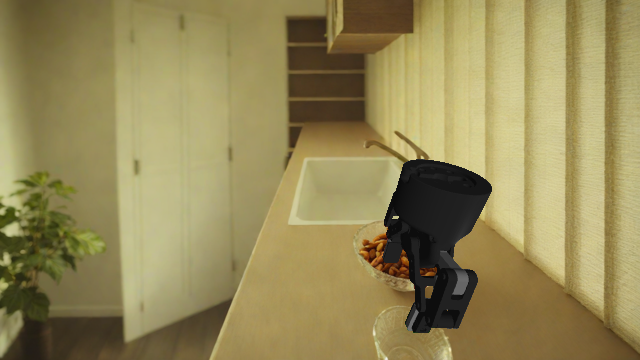}\hfill%
\includegraphics[width=0.25\linewidth]{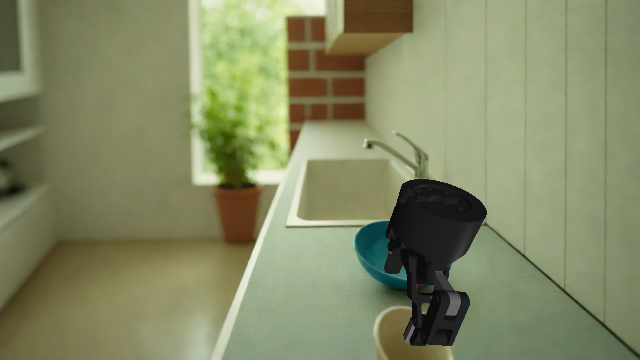}\hfill%
\includegraphics[width=0.25\linewidth]{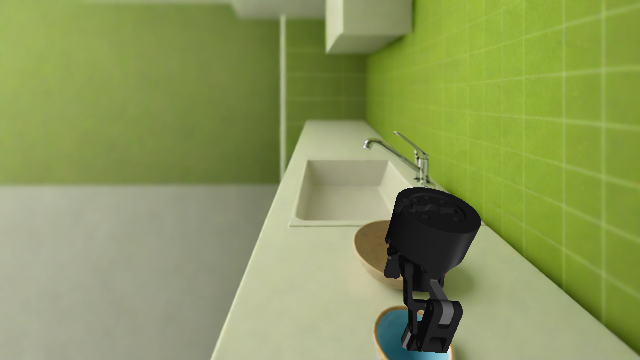}\hfill%
\includegraphics[width=0.25\linewidth]{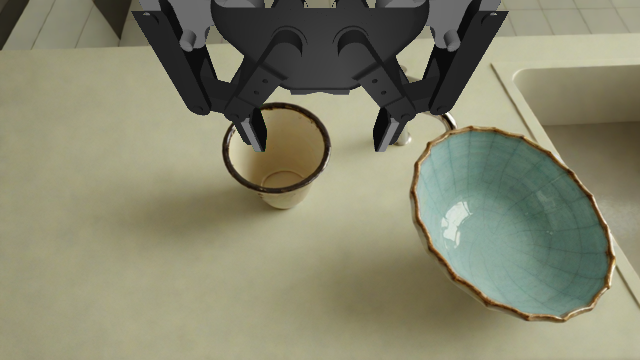}\hfill%
\includegraphics[width=0.25\linewidth]{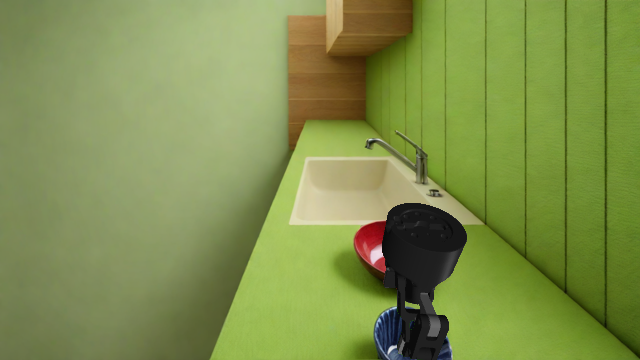}\hfill%
\includegraphics[width=0.25\linewidth]{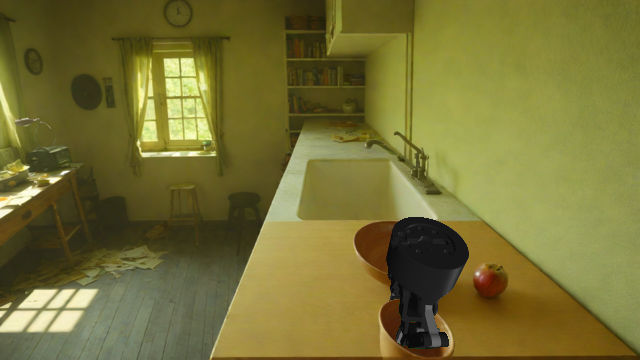}\hfill%
\includegraphics[width=0.25\linewidth]{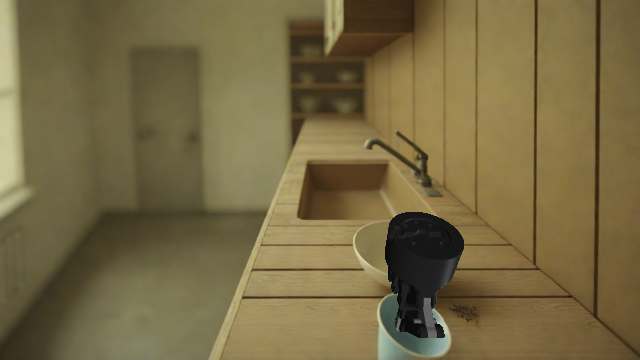}\hfill%
\includegraphics[width=0.25\linewidth]{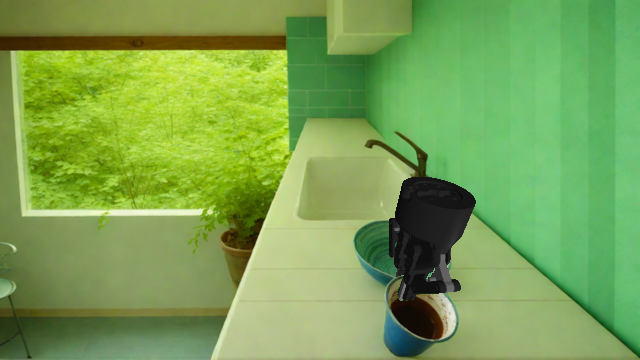}\hfill%
\includegraphics[width=0.25\linewidth]{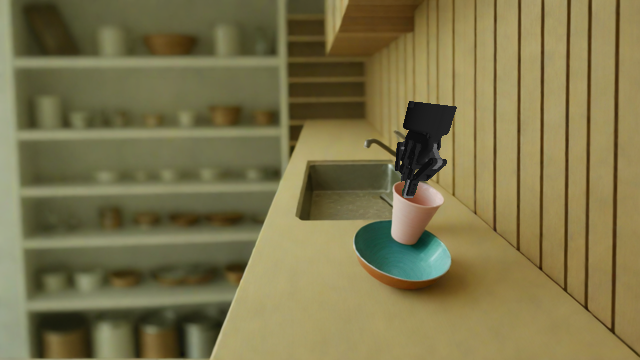}\hfill%
\includegraphics[width=0.25\linewidth]{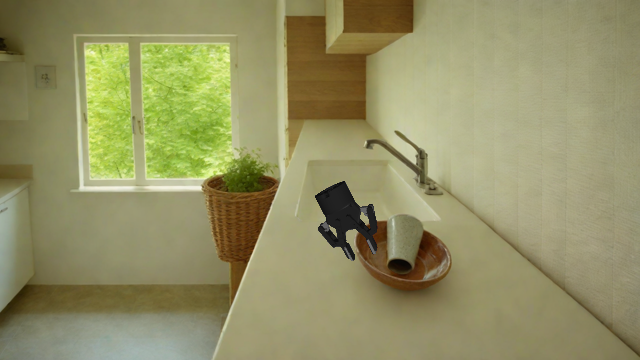}\hfill%
\includegraphics[width=0.25\linewidth]{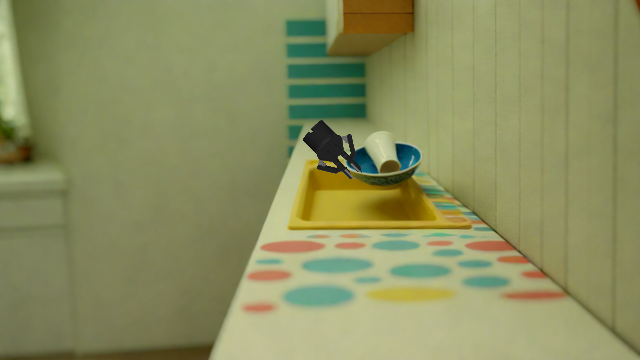}\hfill%
\includegraphics[width=0.25\linewidth]{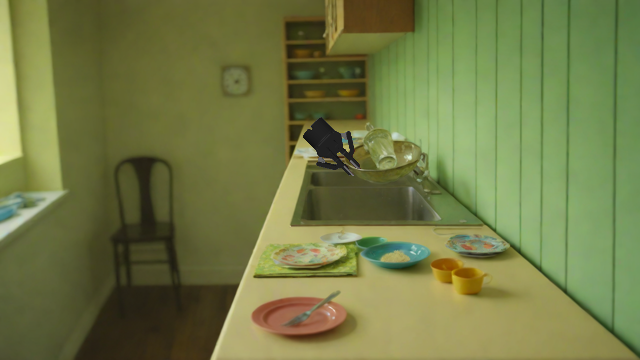}\hfill%
\includegraphics[width=0.25\linewidth]{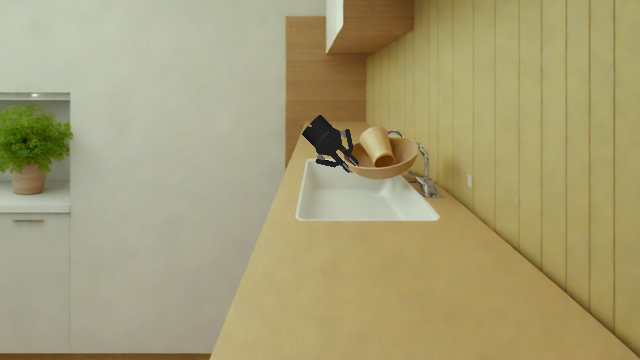}\hfill%
\includegraphics[width=0.25\linewidth]{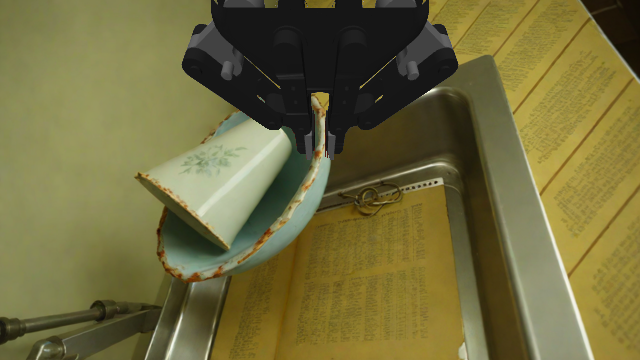}\hfill%
\includegraphics[width=0.25\linewidth]{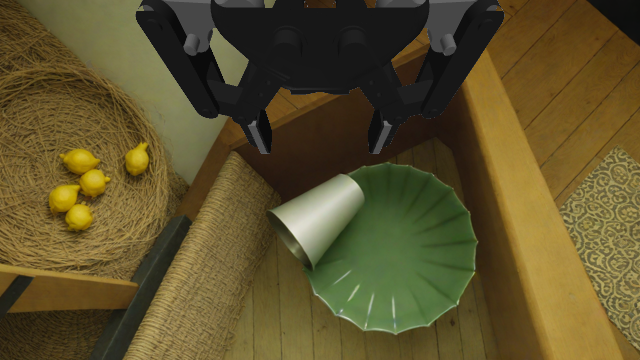}%
\caption{\textbf{Additional Examples of Generated Images from Lucid-XR.} 
Notice the control over lighting, geometry, and content diversity.}
\label{fig:additiona-synthetic-data-examples}
\end{figure}

We include additional examples of synthetic images generated by the LucidXR pipeline in Figure.~\ref{fig:additiona-synthetic-data-examples}.%


\subsection{The Vuer Scene Description Language}

Usage of the MuJoCo engine API in Python tends to follow an imperative pattern,
where objects, material texture, and lighting are changed by mutating MuJoCo's
physics and modeling buffers. To improve  the readability and reusability of the
MuJoCo simulation, we developed a declarative scene description language that
treats the scene as a nested set of scene components that form a tree. This module, \texttt{\color{blue}vuer-mujoco}, enables the user to sketch out a scene via 
the following:

\begin{lstlisting}
from vuer_mujoco import Box, DefaultScene, SimpleTable
from your_code import make_camera_rig

camera_rig = make_camera_rig()
lighting_rig = make_lighting_rig()

table = SimpleTable(pos=[0.0, 0.0, 0.0])

# a 10cm cube, initialized slightly above a table.
cube = Box(size=[0.01, 0.01, 0.01], pos=[0., 0.1, 0.02] + table.surface_origin)

scene = DefaultScene(
    *camera_rig.get_all_cameras(),
    lighting_rig.key, lighting_rig.fill, lighting_rig.back,
    table, cube, robot, robot.mocap_points,
)
\end{lstlisting}

\subsection{Tips on Engineering Synthetic Environments}

\paragraph{Adhesion.} 
Adhesive interactions are involved in many physical processes, including 
surface tension, magnetic interaction with ferro- and para-magnetic materials,
interaction with sticky surfaces, such as an object against the silicone rubber material in a gripper. 

We provide a fishing toy environment that simulates a toy fishing pole with
a magnetic hook that the robot can use to pick up a wooden fish. This simultaneously
involves interaction with flexible material -- a rope. We set the weight of the fish
pieces so that the adhesion is strong enough to lift them off the table, but still 
weak enough to detach with a slight shake of the fishing rod. 

\paragraph{Generating 3D Assets.}
All of the 3D assets shown above are generated either from a text description of the object or stock images of the physical item found on Amazon.com. We use a free version
of the meshy.ai service. The resulting 3D meshes usually contain a large number of faces,
with physical dimensions that are off by two orders of magnitude. We post-process 
these 3D assets by first simplifying them in MeshLab, and then centering and rescaling
using a custom script.

\begin{figure}
\begin{subfigure}[b]{0.5\textwidth}
\includegraphics[width=\linewidth]{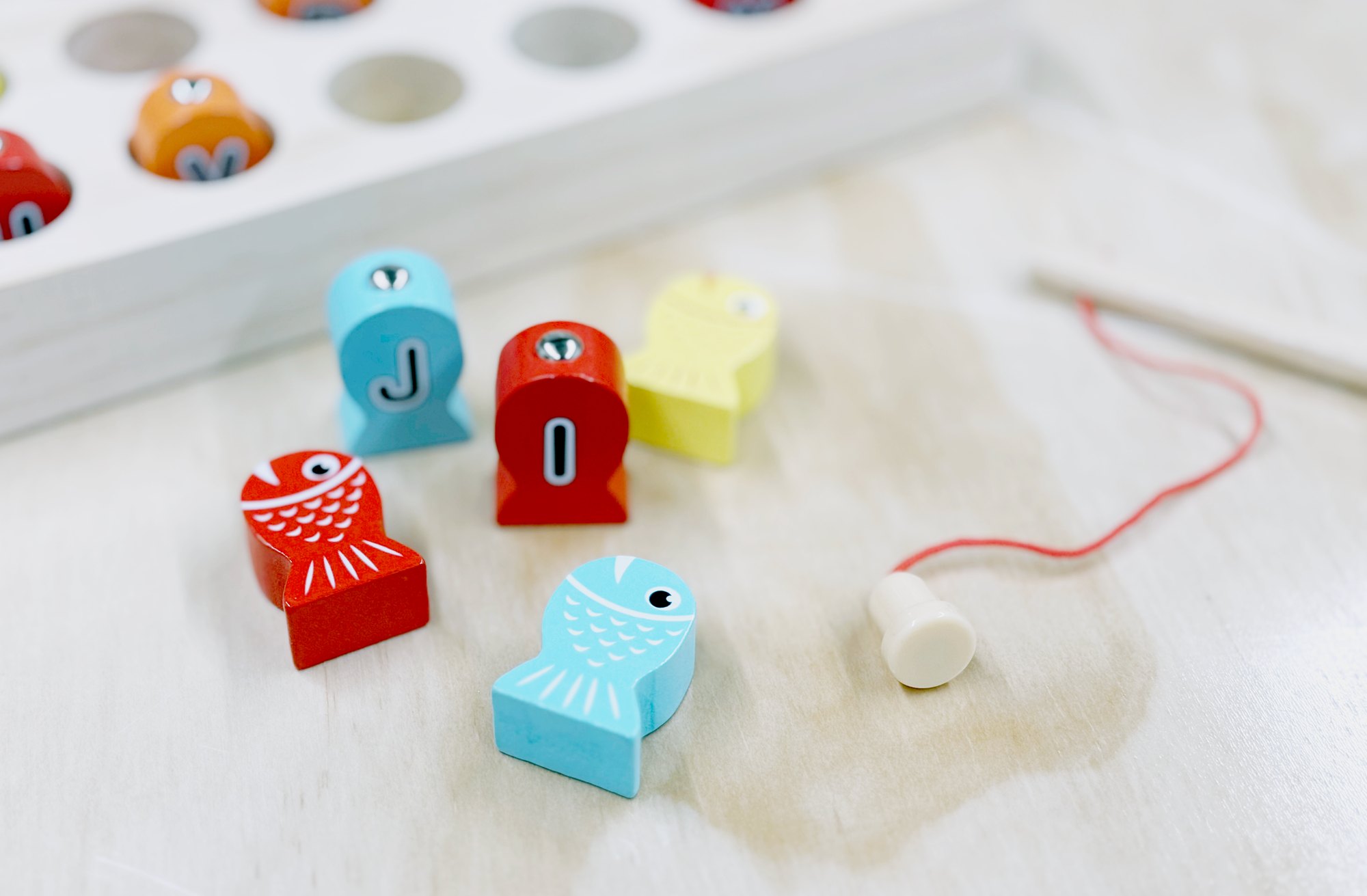}
\caption{Real-world photo of the toy.}
\end{subfigure}\hfill
\begin{subfigure}[b]{0.47\textwidth}
\includegraphics[width=\linewidth]{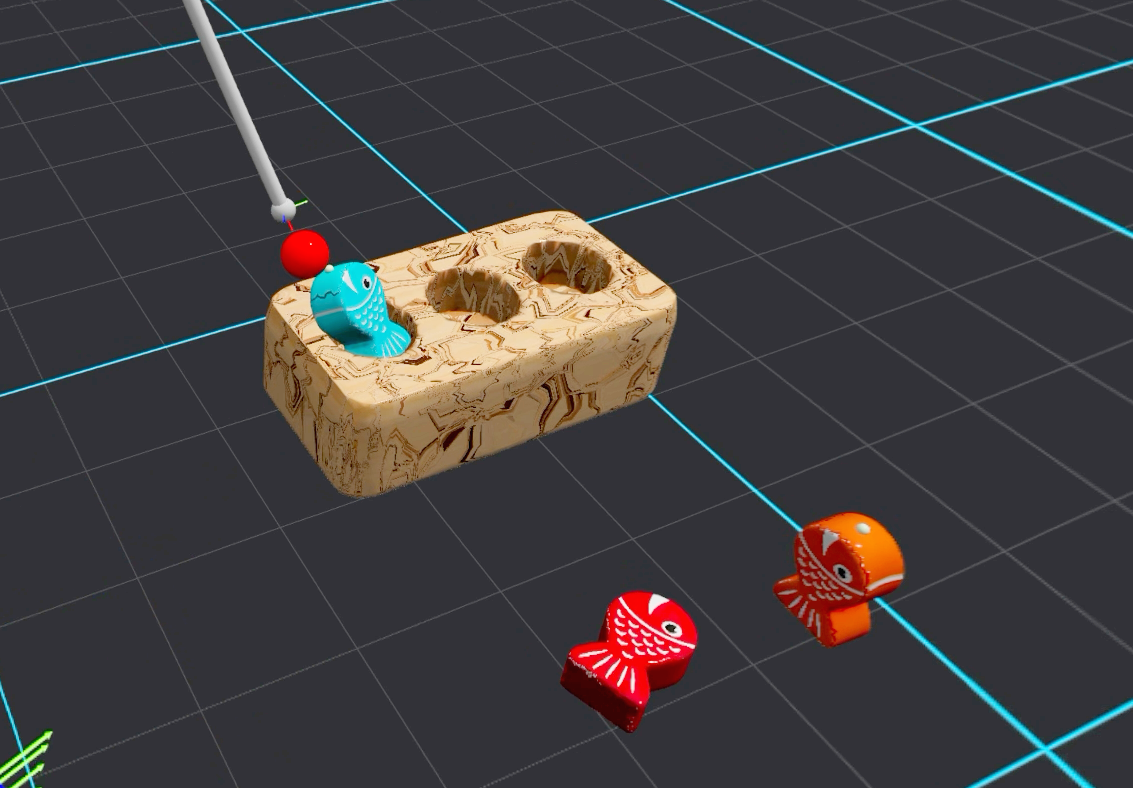}
\caption{Details of simulated learning environment.}
\end{subfigure}
\caption{The real-world fishing toy set (left), versus the simulated learning environment used to collect data (right). The fishing rod has a magnetic hook attached to the end of the line that interacts with a ferromagnetic nail head embedded in the fish. We simulate this by defining adhesion actuators on a small sphere body attached to the fish geometry.}\label{fig:fishing-toy}
\end{figure}

\begin{lstlisting}
from vuer_mujoco import MuJoCoRope

rope = MuJoCoRope( 
  postamble="""
    <actuator>
      <adhesion body="f1-magnet" ctrlrange="0.15 0.16" gain="1"/>
      <adhesion body="f2-magnet" .../>
      <adhesion body="f3-magnet" .../>
    </actuator>
  """,
)
\end{lstlisting}




\subsection{Real-to-sim Evaluation Setup}

We include two realistic testing environments: a) Clean Kitchen
and b) Messy Kitchen. These environments serve as a more controllable proxy of the real-world experiments, to enable faster iteration on the data and learning pipeline.



\begin{figure}[h]
\begin{subfigure}[b]{0.48\linewidth}
\includegraphics[width=\linewidth]{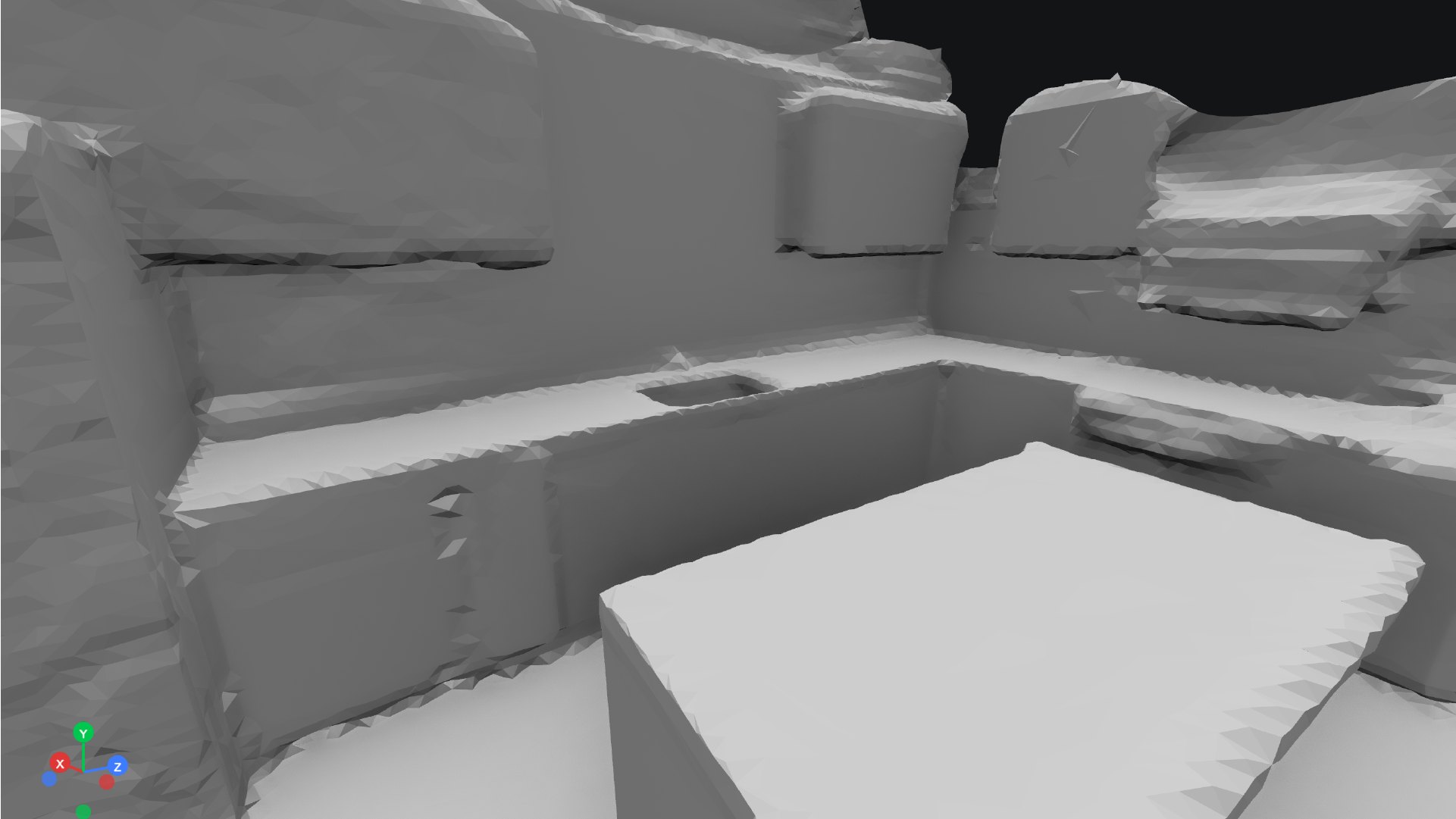}
\caption{Collision Geometry}
\end{subfigure}\hfill%
\begin{subfigure}[b]{0.48\linewidth}
\includegraphics[width=\linewidth]{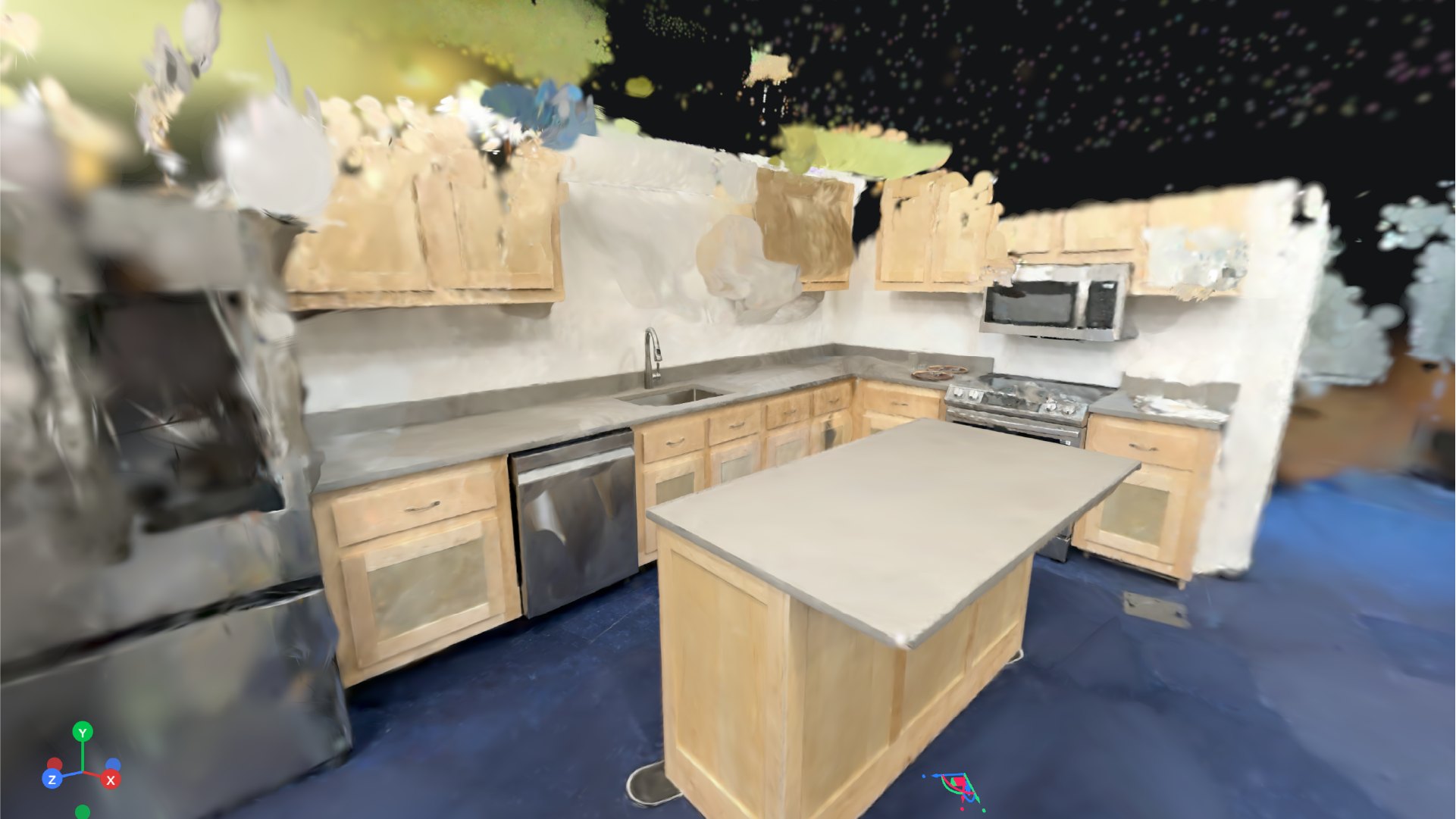}
\caption{Render}
\end{subfigure}
    \caption{\textbf{Clean Kitchen.} Contains no clutter. We place 3D object assets programmatically.}
    \label{fig:align-polycam-mesh}
\end{figure}

\begin{figure}[t]
\begin{subfigure}[b]{0.48\linewidth}
\includegraphics[width=\linewidth]{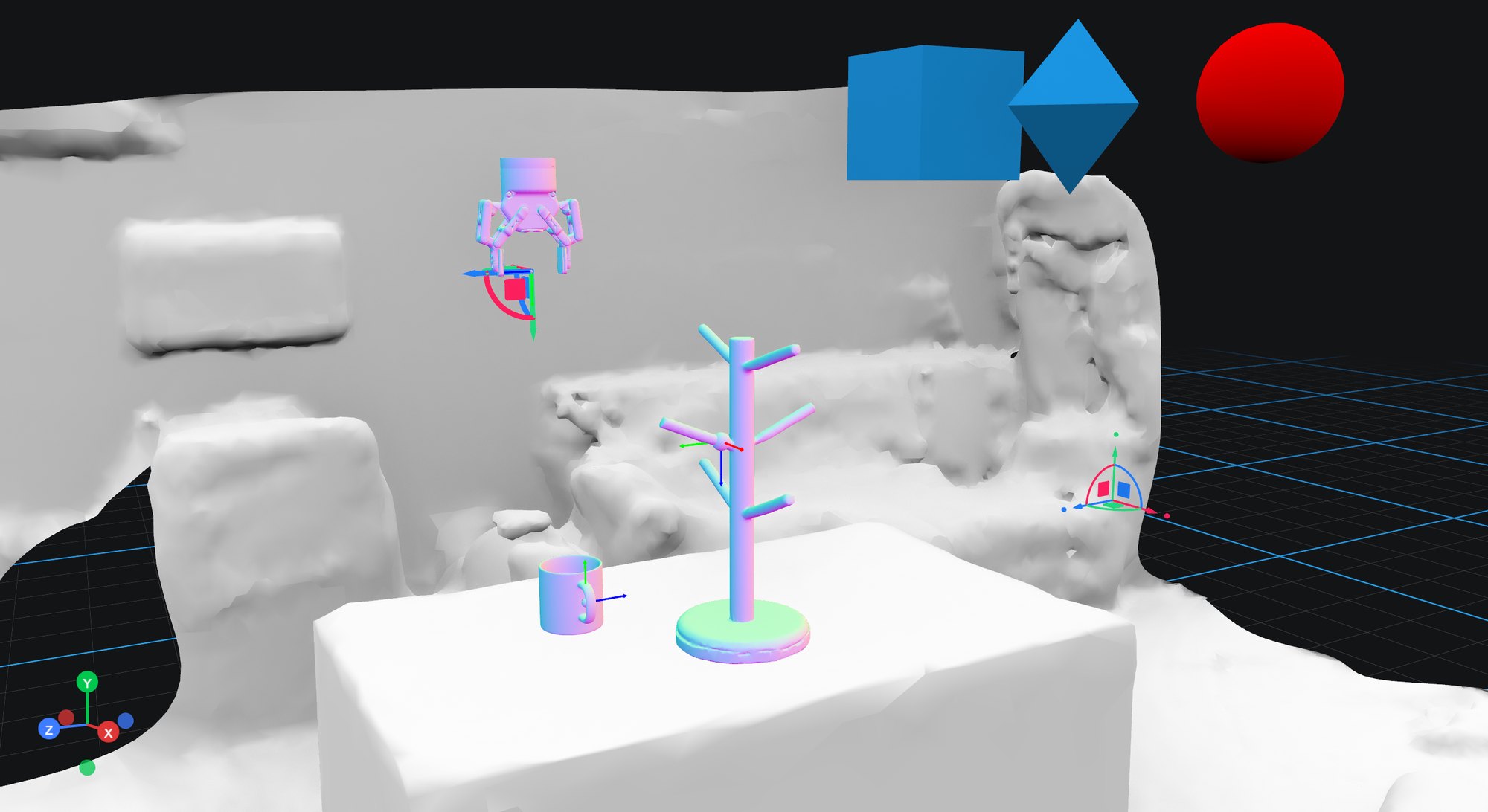}
\caption{Aligning the 3D mesh with the MuJoCo scene.}
\end{subfigure}\hfill%
\begin{subfigure}[b]{0.48\linewidth}
\includegraphics[width=\linewidth]{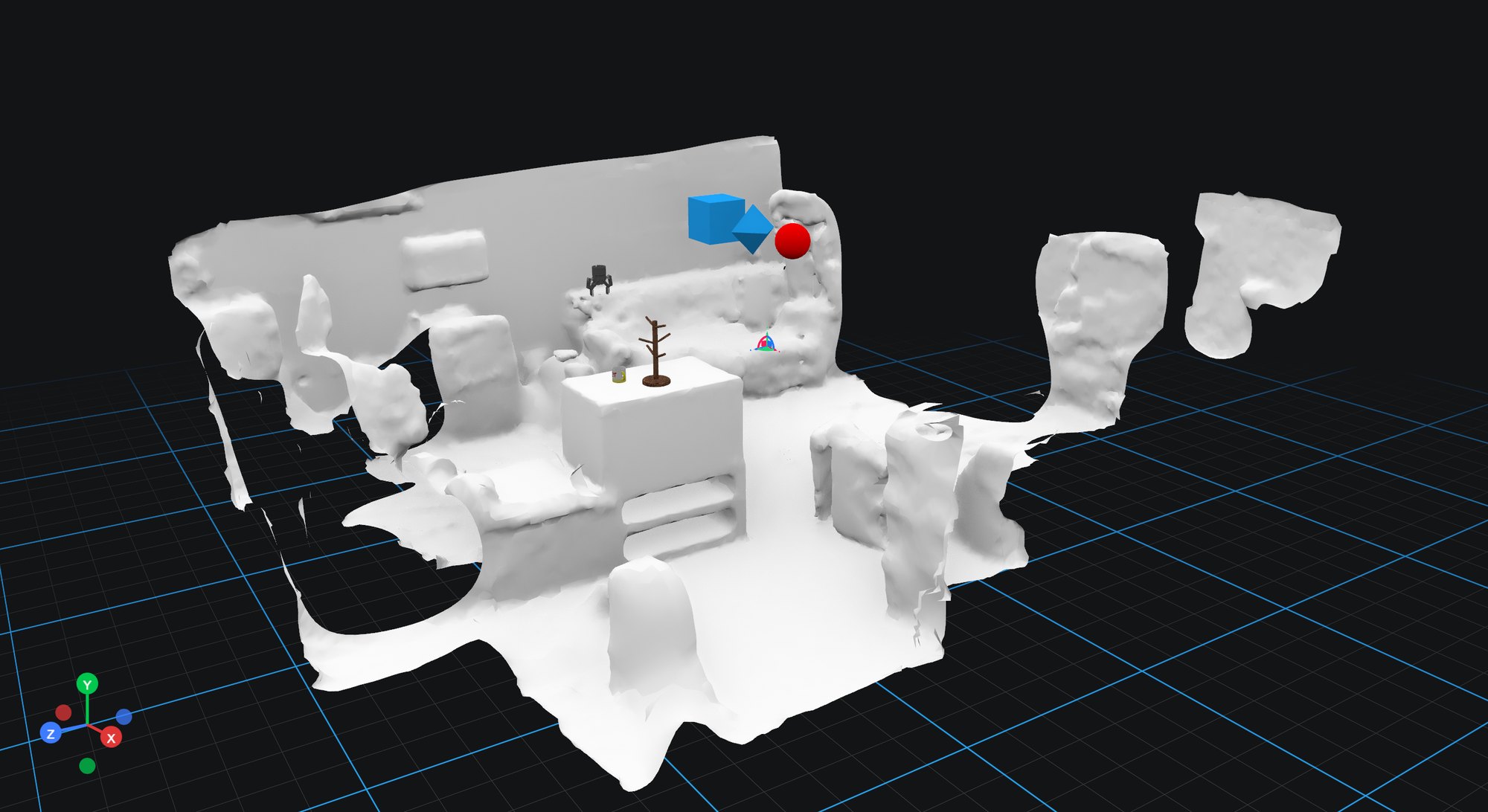}
\caption{The scene after alignment.}
\end{subfigure}
\caption{\textbf{Aligning scan with the physics environment.}}
\label{fig:eval-align-mesh}
\vspace{-1em}
\end{figure}

We align the 3D mesh from PolyCam to the MuJoCo environment (see Figure.~\ref{fig:eval-align-mesh}) manually, shown in Figure~\ref{fig:align-polycam-mesh}.

\subsection{Image Generation Workflow}
\begin{figure}[b]%
\centering%
\vspace{-35pt}%
\includegraphics[width=0.95\linewidth]{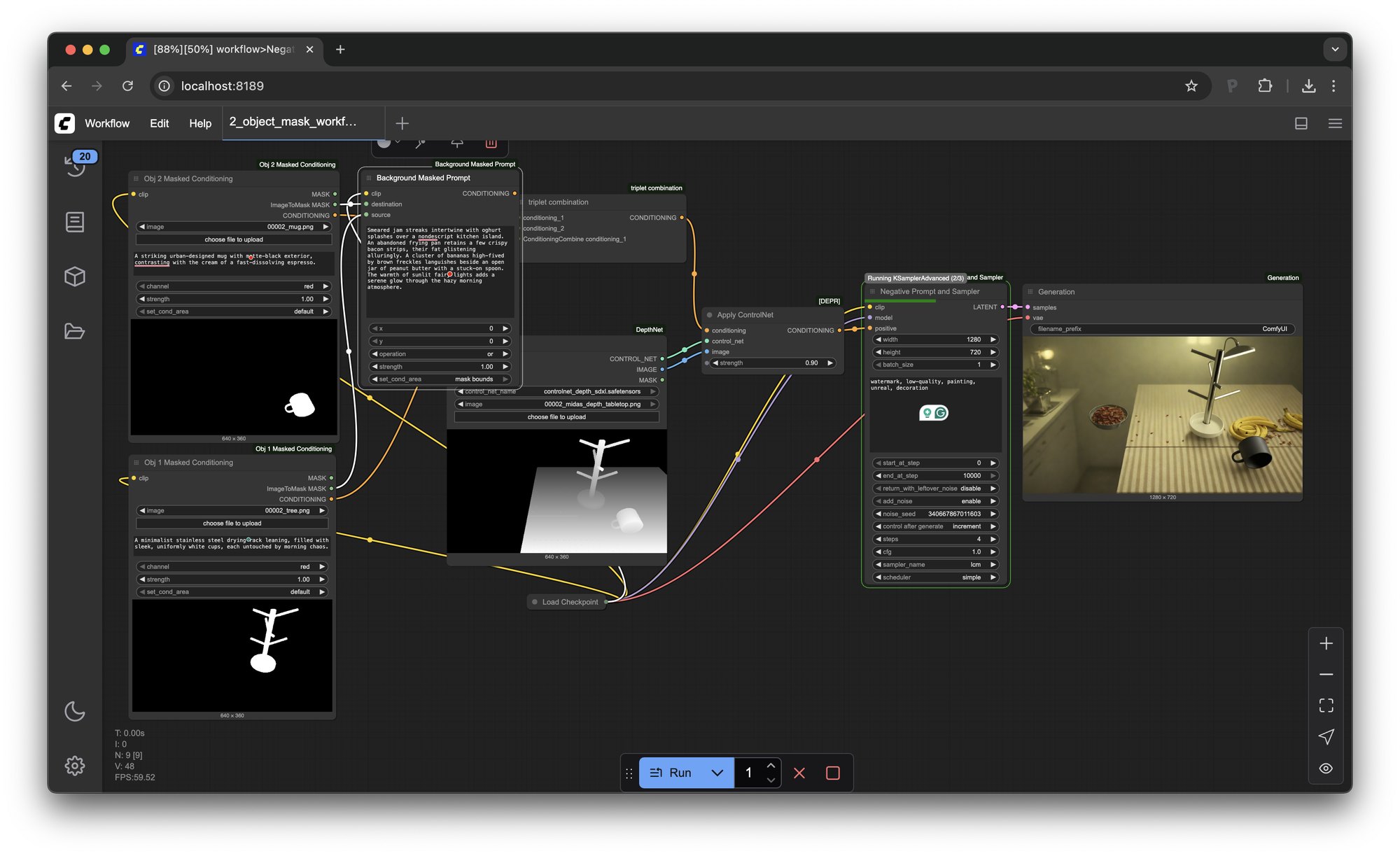}%
\caption{\textbf{Image Generation Workflow.} Using two object masks plus 
a normalized inverse depth image, we are able to control the geometry, 
lighting, and the composition of the generated images.}
\label{fig:comfyui}
\end{figure}
We provide the complete image generation workflow in JSON form in the supplementary material. This workflow can be loaded into ComfyUI as-is. A screenshot of this workflow can be found in Figure.~\ref{fig:comfyui}.

\clearpage
\subsection{Training Details}

We use the Action-Chunking transformer architecture from \cite{zhao2023learning}. Our policy uses multiple backbones (for each input camera feed). The hyperparameters we used are in Table.~\ref{tab:act-hyperparams}.

\begin{table}[h]
\centering
\caption{Training Hyperparameters.}
\footnotesize
\label{tab:act-hyperparams}
\begin{tabular}{lr}
\toprule
\small
\textbf{learning rate}          & $5\mathrm{e}{-5}$ \\
\textbf{batch size}             & 32                 \\
\textbf{number of encoder layers}      & 4                 \\
\textbf{number of decoder layers}      & 7                 \\
\textbf{feedforward dimension}  & 3200              \\
\textbf{hidden dimension}       & 512               \\
\textbf{number of heads}               & 8                 \\
\textbf{chunk size}             & 10               \\
\textbf{KL-weight}              & 10                \\
\textbf{dropout}                & 0.1               \\
\bottomrule
\end{tabular}
\end{table}

\subsection{Example Image Prompts}

We produce a small number of randomly selected text prompts below. 

\textbf{Prompts for the kitchen cup/bowl Scene:}

\begin{lstlisting}
bowl = "A rich green patinated copper bowl, its surface embellished with intricate embossing echoing the craftsmanship of generations past, each detail wrought with precision."

cup = "A minimalist acrylic cup, hard-edged and visually weightless, mimicking the ethos of contemporary transparency."

background = "A chaotic ensemble of baking supplies lie scattered around, flour dusts the edges of a marble countertop, while reflective surfaces from an overhead light palette interact sporadically creating patterns on the brushed steel faucet."
\end{lstlisting}%
\captionof{lstlisting}{
\textbf{The Crossroad of Time.}
Illustrating the intersection of aesthetic past and present through culinary objects.
(Minimalist cup, antique bowl)
}

\begin{lstlisting}
bowl = "A ceramic bowl with swirling patterns of cobalt blue, sunflower yellow, and rose pink glazes, with a high-gloss finish."

cup = "A teacup marked by abstract patches of emerald green and crimson red, seemingly creating a kaleidoscopic effect."

background = "A backdrop filled with chaotic elements \u2014 a countertop cluttered with spatulas, over-ripe fruit, and colorful cracked ceramic tiles leaning against the wall."
\end{lstlisting}%
\captionof{lstlisting}{
\textbf{Mosaic Glaze Fantasy.}
A close-up of vibrant ceramics with a dazzling faucet.
    (Colorful ceramics, shining faucet)
}

\textbf{Prompts for the mug tree environment:}

\begin{lstlisting}
mug = "An oversized mug with a faded \u2018World's Best Bartender\u2019 logo, filled with room-temperature coffee and a tiny lipstick mark on its rim."
mug_tree = "A sleek, modern steel drying rack holding a neat row of eclectic mugs, each hanging at a slightly different angle."
background = "The counter is scattered with an array of bar staples, including an upturned shaker, a jar of maraschino cherries left open with sticky syrup pooled at the base, tiny, colorful drink umbrellas laying flat, and cocktail recipe cards partially obscured by a dishcloth. The dim overhead lighting casts a cozy yet neglected ambiance, reflecting off of glass surfaces, highlighting water stains and the subtle sheen of unpolished wood. Seasonal drink posters curl at the corners on the walls behind, and the muted hum of conversation drifts from unseen patrons."

\end{lstlisting}%
\captionof{lstlisting}{
\textbf{Afternoon Clutter.}
An up-close view of a neglected bar counter amidst a busy afternoon, capturing the hustle and bustle pausing for just a moment. (bar counter)
}
\begin{lstlisting}
mug = "A tall novelty mug with light visible scuffs, its personality yet vibrant, with a traveler's emblem boldly emblazoned.",
mug_tree = "Innovatively designed transparent drying rack, its architecture raising daily-use mugs to a dignified height",
background = "The bar counter's imperfections speak of immediate use \u2013 glistening droplets from pint glasses left haphazardly, several salt grains shining under muted sunlight, and a stray beer tab nestled amongst dried orange zest. A sharp, storied scratch veers perilously close to a vase of gaudy red carnations. Decorative but decisively aged beer mats pattern the visible stained wood, while a distant chorus of cheerful banter tingles the air.",
\end{lstlisting}%
\captionof{lstlisting}{
\textbf{Unexpected Afternoon Clutter.}
A surprisingly busy scene of a bar counter overtaken by mid-day merriment. (bar counter)
}

\end{document}